\documentclass[twoside,11pt]{article}

\newcommand\chngd[1]{{\color{red} #1}}
\renewcommand\chngd[1]{#1}

\newcommand\chngdjair[1]{{\color{red} #1}}
\renewcommand\chngdjair[1]{#1}

\newcommand\chngdjaircrc[1]{{\color{blue} #1}}
\renewcommand\chngdjaircrc[1]{#1}

\newcommand\blue[1]{{\color{blue} #1}}
\newcommand\green[1]{{\color{green} #1}}
\newcommand\red[1]{{\color{red} #1}}

\usepackage[utf8]{inputenc} 
\usepackage[T1]{fontenc}    
\usepackage{hyperref}       
\usepackage{url}            
\usepackage{booktabs}       
\usepackage{amsfonts}       
\usepackage{nicefrac}       
\usepackage{microtype}      
\usepackage{xcolor}         
\usepackage{caption}        %
\usepackage{soul} 
\usepackage{amsmath}        
\usepackage{rotating}       

\usepackage[section]{placeins}

\usepackage{graphicx}
\usepackage{subcaption}
\usepackage{wrapfig}
\usepackage{verbatim}
\usepackage{listings} 
\newcommand{\denselist}{\itemsep -2pt\partopsep 0pt}

\usepackage{jair, theapa, rawfonts}

\usepackage[noend]{algpseudocode}
\usepackage{algorithm}


\jairheading{77}{2023}{821-890}{10/2022}{07/2023}
\ShortHeadings{MDP Playground}{Rajan, Borja, Guttikonda, Ferreira, Biedenkapp, von Hartz \& Hutter}
\firstpageno{821}

\begin{document}

\title{MDP Playground: An Analysis and Debug \\ Testbed for Reinforcement Learning}

\author{\name Raghu Rajan \email rajanr@cs.uni-freiburg.de\\
  \addr University of Freiburg \vspace{0.23cm}\\
\name Jessica Lizeth Borja Diaz \email jessicalizethborja@gmail.com\\
  \addr University of Freiburg \vspace{0.23cm}\\
\name Suresh Guttikonda \email suresh.guttikonda@students.uni-freiburg.de\\
  \addr University of Freiburg \vspace{0.23cm}\\
\name Fabio Ferreira \email ferreira@cs.uni-freiburg.de\\
  \addr University of Freiburg \vspace{0.23cm}\\
\name André Biedenkapp \email biedenka@cs.uni-freiburg.de\\
  \addr University of Freiburg \vspace{0.23cm}\\ 
\name Jan Ole von Hartz \email hartzj@cs.uni-freiburg.de\\
  \addr University of Freiburg \vspace{0.23cm}\\
\name Frank Hutter \email fh@cs.uni-freiburg.de\\
  \addr University of Freiburg \vspace{0.23cm}\\
}


\maketitle

\begin{abstract}
We present \textit{MDP Playground}, a testbed for Reinforcement Learning (RL) agents with \textit{dimensions of hardness} that can be controlled independently to challenge agents in different ways and obtain varying degrees of hardness in toy and complex RL environments. We consider and allow control over a wide variety of dimensions, including \textit{delayed rewards}, \textit{sequence lengths}, \textit{reward density}, \textit{stochasticity}, \textit{image representations}, \textit{irrelevant features}, \textit{time unit}, \textit{action range} and more. We define a parameterised collection of fast-to-run toy environments in \textit{OpenAI Gym} by varying these dimensions and propose to use these to understand agents better. We then show how to design experiments using \textit{MDP Playground} to gain insights on the toy environments. We also provide wrappers that can inject many of these dimensions into any \textit{Gym} environment. We experiment with these wrappers on \textit{Atari} and \textit{Mujoco} to allow for understanding the effects of these dimensions on environments that are more complex than the toy environments. We also compare the effect of the dimensions on the toy and complex environments. Finally, we show how to use \textit{MDP Playground} to debug agents, to study the interaction of multiple dimensions and describe further use-cases.

\end{abstract}


\section{Introduction}
We begin our discussion by motivating why we need dimensions of hardness and toy environments in Reinforcement Learning.

\subsection{Need for Dimensions of Hardness} Reinforcement Learning (RL) algorithms can solve many disparate tasks, such as helicopter aerobatics, game-playing and continuous control \shortcite{ng_kim2004autonomous,dqn_mnih2013playing,sac_haarnoja2018}. However, multiple recent papers (e.g., \shortciteR{deep_rl_that_matters,implementation_matters_engstrom_iclr20,matters_on_policy_methods_andrychowicz_iclr21}) have shown that RL agents tend to be brittle. It has been argued that current deep RL research has been increasing the complexity of the dynamics but has not paid much attention to the state distributions and reward distribution over which RL policies work and that this has made RL agents brittle \shortcite{andersson_benchmark_dimensions_icml_ws_2018}. We believe as a step towards understanding and designing better RL agents, research analysing the effects of common challenges faced by RL agents is of growing importance \shortcite{bsuite_osb2019behaviour,challenges_real_world_rl_arxiv_2003_11881}. To this end, we have designed a framework, \textit{MDP Playground}, which allows such challenges to be added, in a controlled manner, to RL environments. \chngdjair{We describe a set of such challenges in this paper which we term \textit{dimensions of hardness} as they can be controlled independently of each other in \textit{MDP Playground}.}

\subsection{Need for Toy Environments} Furthermore, a lot of the insights obtained on the \textit{standard} environments are on very complex and in many instances \textit{blackbox} environments. There are many different types of such environments, as many as there are different kinds of tasks in RL \shortcite<e.g.>{todorov2012mujoco,atari_bellemare_2013,cobbe_procgen_icml_20}. They are each tied to \textit{specific} kinds of tasks. The underlying assumptions in many of these environments are that of a Markov Decision Process (MDP) \shortcite{puterman_mdps_book_1994} or a Partially Observable MDP (POMDP) \shortcite<see e.g.,>{sutton2018reinforcement}. However, there is a lack of simple and accessible problems which capture common challenges seen in RL and let researchers experiment with them in a fine-grained manner. Many researchers design their own toy problems which capture key aspects of their problems and then try to use them to gain insights on \textit{whitebox} environments\footnote{\chngdjair{We use the term \textit{whitebox} in analogy with \textit{blackbox}, i.e., whitebox environments are those which are accessible and where we know all the internal details of how they work.}} because the standard \textit{complex} environments, such as \textit{Atari} and \textit{Mujoco}, are too expensive or too opaque.

To sum up the disadvantages of complex environments: \textbf{1)} They are very expensive to evaluate. For example, a DQN \shortcite{dqn_mnih2013playing} run on an \textit{Atari} \shortcite{atari_bellemare_2013} environment took us 4 CPU days and 64GB of memory to run. \textbf{2)} The environment structure itself is so complex that it can lead to ``lucky'' agents performing better (e.g., in \shortciteR{deep_rl_that_matters}). Furthermore, different implementations even using the same libraries can lead to very different results \shortcite{deep_rl_that_matters}.
\textbf{3)} Many dimensions of hardness are concurrently present in the environments and do not allow us to independently test their impact on agents' performances.

Motivated by the above discussion on dimensions of hardness and toy environments, we introduce \textit{MDP Playground}, a platform whose intent is to: \textbf{1)} help us to understand RL agents better on toy and complex environments. This is achieved by injecting dimensions of hardness into the environments in a fine-grained manner. This is similar in spirit to the work of \shortciteA{imagenet_c_hendrycks_iclr_2019}. \textbf{2)} ``unit test'' characteristics of agents on toy MDPs. The unit tests are aimed at overcoming the disadvantages of complex environments mentioned above - the evaluation is cheap, the structure of the environments is simple and accessible and one can \textit{independently} inject desired dimensions of hardness.

We would like to emphasise here that the toy environments in \textit{MDP Playground} are intended as a \textit{complement} to complex environments, and not to replace them, because they trade off compute, accessibility and control at the expense of being further removed from the real world. They may be useful to try out new ideas quickly but may not lead to immediate results on the complex environments which may be the long-term goal of research. As an example of how this may be useful, we mention here that Q-learning \shortcite{sutton2018reinforcement} and Double Q-learning \shortcite{double_q_Hasselt_2010_nips} were shown as proofs of concept on tabular or small environments several years before the advent of deep RL and complex environments. Neither could be shown transfer to complex environments at the time of publication. As such, toy environments in \textit{MDP Playground} are meant as accessible environments where the ground truth is known and where one might quickly test out whether a new idea helps against a dimension of hardness and not, e.g., to tune hyperparameters for complex environments.


The main contributions of this paper are:
\begin{itemize}
\denselist
\item We identify and discuss dimensions of hardness of MDPs that can have a significant effect on agent performance, both for discrete and continuous environments; 
\item We discuss experimental designs for toy and complex environments that: $1)$ show insights on toy environments alone, $2)$ compare the effect of dimensions on toy and complex environments;
\item The toy environments in \textit{MDP Playground} provide unit testing and debugging capabilities for RL agents with the ground truth being available; a single seed of an experiment can be run in as few as $30$ seconds on a single core of a laptop;
\item Many of the studied dimensions of hardness stem from partial observability and the related experiments provide an analysis of how partial observability can degrade performance; to the best of our knowledge, this is the first such systematic analysis.
\end{itemize}

\chngdjair{The paper is structured as follows. We define MDPs, NMRDPs \shortcite{bacchus1996rewarding} and POMDPs needed to understand the dimensions of hardness in \textit{MDP Playground} and then describe the motivations for the dimensions of hardness in \textit{MDP Playground} from a high-level and general point of view in Section \ref{sec:background_dimensions}. We describe \textit{MDP Playground} and its components in Section \ref{sec:mdpp}. We provide a low-level description with details of the specific implementation of the dimensions of hardness in \textit{MDP Playground} in Section \ref{sec:mdpp:detailed}. We discuss design decisions involved in implementing \textit{MDP Playground} in Section \ref{sec:design_decisions}. We perform experiments and analyse them in Section \ref{sec:experiments}. We discuss related work in Section \ref{sec:related_work}. We discuss limitations of our approach and its ethical and societal implications in Section \ref{sec:limitations_and_societal_impact}. Finally, we provide conclusions and discuss future work in Section \ref{sec:conclusion}.}

\section{Dimensions of Hardness in MDPs}
\label{sec:background_dimensions}
To identify dimensions of hardness and to distinguish between different kinds of \textit{state} that we will encounter, we need to begin with some definitions.

\subsection{MDPs, NMRDPs and POMDPs}
We define an MDP as a 7-tuple $(MS, A, P, R, \rho_o, \gamma, T)$, where $MS$ is the set of states, $A$ is the set of actions, \chngdjaircrc{$P: MS \times A \times MS \rightarrow \mathbb{R}^+$ describes the transition dynamics as a probability density of reaching a state given the current state and action\footnote{\chngdjaircrc{The default MDPs in \textit{MDP Playground} are deterministic. In such cases, $P$ can also be represented as $P: MS \times A \rightarrow MS$. When describing such deterministic transition functions, we will assume this to be the case and talk about $P$ as such.}}, $R: MS \times A \times MS \times \mathbb{R} \rightarrow \mathbb{R}^+$ describes the reward dynamics as a probability density of obtaining a reward given the current state, action and next state}, $\rho_o: MS \rightarrow \mathbb{R}^+$ is the initial state distribution, $\gamma$ is the discount factor and $T$ is the set of terminal states. \chngdjair{We will consider systems which also have non-Markovian rewards because in the case of some dimensions of hardness, we have rewards which depend on the past history of the system. So, we additionally define an NMRDP, a \textit{Non-Markovian Reward Decision Process} following \shortcite{bacchus1996rewarding} which corresponds to our MDP of the system but has a different state space, $NS$. It will be worthwhile to distinguish between the current state of the system which is non-Markovian and its corresponding Markovian state which contains the history, to be able to fully understand the details of \textit{MDP Playground}. A state $ns \in NS$ will represent the current state of the system and a state $ms \in MS$ will represent the Markovian state. As such, any state $ms \in MS$ would correspond to a tuple of states $(ns_0, ns_1, ...)$ from $NS$. We will refer to the states from $NS$ as $N$-states and the states from $MS$ as $M$-states. The NMRDP is defined using $NS$ instead of $MS$ in its definitions of $P$ and $R$\footnote{As a result, $P$ and $R$ would be overloaded symbols. Which $P$ or $R$ we are referring to should be clear from the context. If unclear, we will mention it in the text.}}

Finally, we define a POMDP with two additional components - $O$ represents the set of observations and $\Omega: MS \times A \times O \rightarrow \mathbb{R}^+$ describes the probability density function of an observation given an $M$-state and action. Following \shortciteA{pomdp_info_state_definitions}, we will use \textit{information state} to mean the state representation used by the agent and \textit{belief state} \chngdjaircrc{\shortcite{cassandra_aaai_1994_acting_optimally_pomdps}} as the posterior belief of the unobserved Markov state given the full observation history. POMDPs can be modelled as fully observable MDPs by considering the belief state (i.e., the posterior belief of the unobserved state given all the observations made by the agent) as an information state \shortcite{pomdp_info_state_definitions}. As a result, the theory and algorithms for exact and approximate planning for MDPs become applicable to POMDPs and using the belief state as the information state by an agent would be sufficient to compute an optimal policy. However, since the full observation history is not tractable to store for many environments, the last few observations are used as the information state which renders it only partially observable. This is an important point to keep in mind because some of the motivated dimensions of hardness are actually due to the information state being non-Markov.\footnote{\chngdjair{We have introduced MDPs, NMRDPs and POMDPs, because all the concepts - the historical state of the system (which is Markov), the current state of the system (which is non-Markov) and the observations are distinct and aid an unambiguous explanation of the details of the dimensions of hardness.}}

We also mention here the $Q^*$-value \shortcite{dqn_mnih2013playing} and use it as an example to argue how violations of assumptions may lead to degradation in performance. For a Markov state $ms \in MS$ and action $a$, a policy $\pi$ and $r_t$ the reward a timestep $t$, $Q^*$ is defined as: $Q^*(ms, a) = \max\limits_{{\pi}} \mathbb{E}\left[\sum_{t=0}^{\infty}\gamma^t r_t \| ms_t = ms, a_t = a, \pi\right]$.


\subsection{Motivations of Dimensions and Implementations}
\label{sec:motivations_dimensions}
We try to identify dimensions of hardness that can be injected independently of each other by going over the many components of an MDP and motivated by the literature. We try to motivate as many as possible, to allow researchers to systematically study and gain new insights. 
\chngdjair{To make the understanding of these dimensions of hardness more concrete, we briefly also describe the implementation of the dimensions of hardness in \textit{MDP Playground} in this section. To be able do so, we briefly also describe the toy discrete and continuous environments in MDP Playground. The text in this section is more of a \textit{top-down}/\textit{dimensions of hardness}-oriented presentation. For a more detailed and lower-level description of these environments and their dimensions of hardness, we refer the reader to Section \ref{sec:mdpp:detailed}, where additional dimensions which are mostly trivial to understand are described as well.

In the discrete case, the $N$-state and action spaces of the system, $NS$ and $A$, contain \textit{categorical} elements that are labelled from $0$ to $\|NS\| - 1$ and $0$ to $\|A\| - 1$ respectively, however, there is no ordering. As such, elements of both $NS$ and $A$ are 1-dimensional integers.} $P$ is formulated as a graph such that each $N$-state is a vertex and elements of $NS$ are connected to other elements of $NS$ through elements of $A$ which form the edges. All the vertices in the graph are set to have the same degree which lends $P$ a uniform structure: this avoids having ``lucky'' regions in the environment. In the continuous case, environments correspond to the simplest real world task we could find: moving a rigid body to a target point \shortcite{klink_self_paced_rl}. $P$ is formulated such that each action dimension affects the corresponding space dimension: for a given $N$-state, $ns$, the derivative of $N$-state, $\dot{ns}$, is set to be equal to the action applied for \textit{time unit} seconds on the body, in the simplest version of the environment. This is integrated over time to yield the next state.

\begin{figure}[t]
        \centering
        \begin{subfigure}[]{0.23\textwidth}
            \centering
            \includegraphics[width=\textwidth]{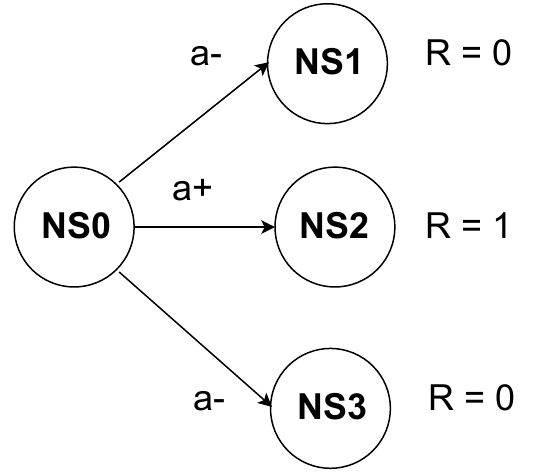}
            \caption[]%
            {{\small }}    
            \label{fig:r_density_diag}
        \end{subfigure}
        \begin{subfigure}[]{0.23\textwidth}
            \centering
            \includegraphics[width=\textwidth]{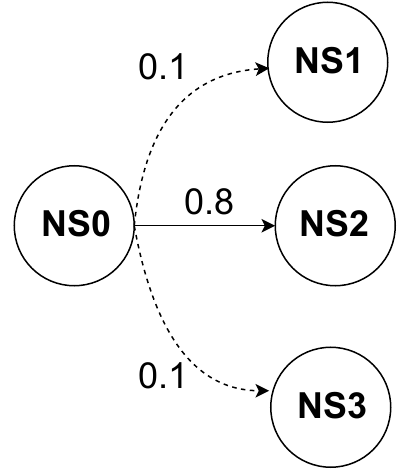}
            \caption[]%
            {{\small }}
            \label{fig:p_noise_diag}
        \end{subfigure}
        \begin{subfigure}[]{0.23\textwidth}
            \centering
            \includegraphics[width=\textwidth]{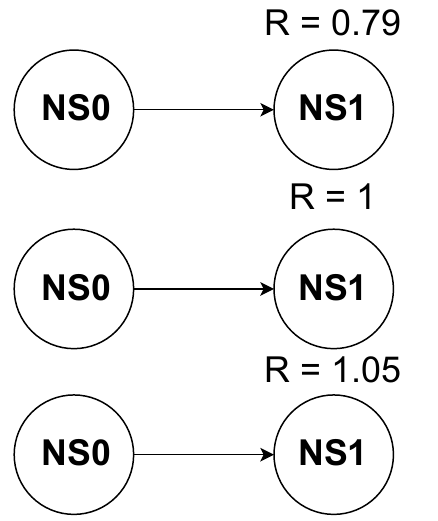}
            \caption[]%
            {{\small}}
            \label{fig:r_noise_diag}
        \end{subfigure}
        \begin{subfigure}[]{0.23\textwidth}
            \centering
            \includegraphics[width=\textwidth]{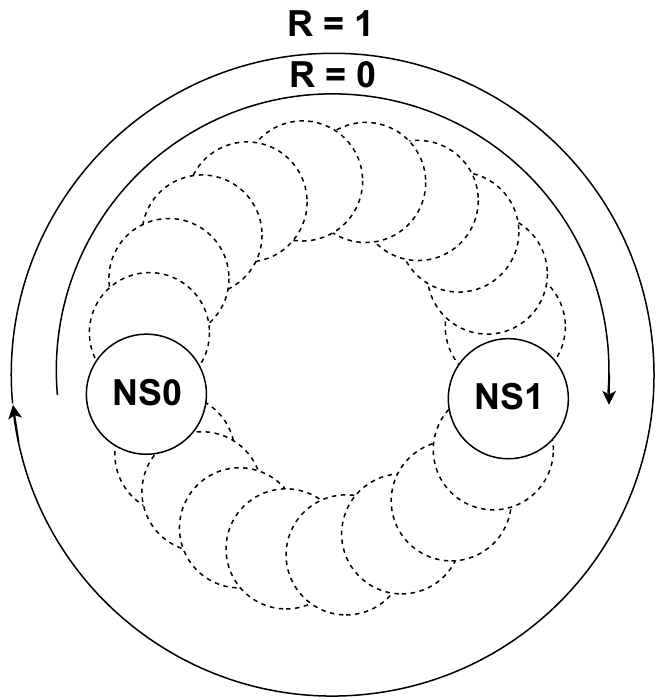}
            \caption[]%
            {{\small}}
            \label{fig:r_seq_diag}
        \end{subfigure}
        \begin{subfigure}[b]{0.45\textwidth}
            \centering
            \includegraphics[width=\textwidth]{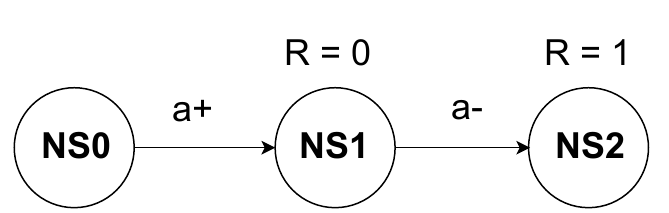}
            \caption[]%
            {{\small }}
            \label{fig:r_delay_diag}
        \end{subfigure}
        \begin{subfigure}[b]{0.45\textwidth}
            \centering
            \includegraphics[width=\textwidth]{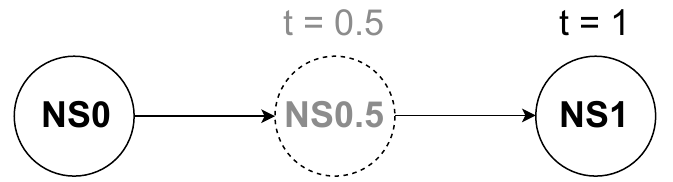}
            \caption[]%
            {{\small }}
            \label{fig:time_unit_diag}
        \end{subfigure}

        \caption[]
        {Some of the dimensions of hardness depicted visually to aid understanding. \chngdjair{Please note that the sub-figures do not use accurate graphical notation, but are rather meant to be guiding visualisations about how a particular dimension of hardness works}. We depict states of the system under consideration, i.e. the $N$-states as defined in the text. Not all $N$-states and actions are depicted to focus on the dimension of interest. Rewarding actions are shown as ``a+'' while actions shown as ``a-'' are not rewarding. Reward is shown as ``R'' and time unit as ``t''. (a) \textbf{$R$ density}: only 1 of 3 possible actions (a+) leads to a reward. (b) \textbf{$P$ noise}: A noise of 0.2 (split into 0.1 and 0.1 and shown with dotted lines) is shown to lead the agent to an $N$-state which is not the true next $N$-state. (c) \textbf{$R$ noise}: The \textit{same} transition leads to different rewards. (d) \textbf{Sequence Length}: Assume the task is to move in a circle, then executing a sequence of actions starting from $N$-state NS0 leading to $N$-state NS1 executes only a semi-circle and leads to $0$ reward (intermediate $N$-states along the sequence shown in dashed lines). Executing the full circle and returning to $N$-state NS0 leads to a reward of 1. \textit{Note}: there may be multiple such sequences corresponding to the task, but the emphasis here is on needing more than $1$ action, i.e., a sequence. (e) \textbf{$R$ delay}: The rewarding action (a+) leads to a reward not immediately but a step later than it was executed and this reward is achieved even though an action inconsequential to achieving the reward (a-) was performed in between. \textit{Note}: the reward would have been achieved a step later irrespective of which action was performed in the second step. (f) \textbf{Time Unit}: We depict a ``half'' action, i.e., performed for a \textit{time unit} that is half the default time unit, leading to an intermediate $N$-state.}
        \label{fig:dimensions_diag}
\end{figure}

We now describe many of the dimensions of hardness and briefly their implementations in \textit{MDP Playground}. Please see Figure \ref{fig:dimensions_diag} for guiding visualisatons.

\subsubsection{Reward Delay} In many situations, agents perform an action that leads to a rewarding trajectory but the agent is only rewarded in a \textit{delayed} manner \shortcite<see e.g.,>{rudder_arjonamedina2018} (see Figure \ref{fig:r_delay_diag}). For example, shooting at an enemy ship in \textit{Space Invaders} leads to rewards much later than the action of shooting. Any action taken and the resulting trajectory afterwards is inconsequential to obtaining the reward for destroying that enemy ship.
Regarding the $Q^*$ value, this means that if an incorrect information state is used, then updates performed for approximating $Q^*$ will tend to assign partial credit also to inconsequential information state features and actions. 
In \textit{MDP Playground}, the reward can be artificially delayed by a non-negative integer number of timesteps, $d$.

\subsubsection{Sequence Length} In many environments, \chngdjair{a reward is obtained for a \textit{sequence} of $N$-states resulting from a sequence of actions taken and not just the current $N$-state and action (see Figure \ref{fig:r_seq_diag}). A simple example is executing a tennis serve, where one needs a sequence of actions which leads to a rewarding $N$-state trajectory that results in a point, e.g., if an ace was served. In contrast to \textit{delayed rewards}, rewarding a sequence of $N$-states resulting from a sequence of actions addresses the trajectory followed as being consequential to obtaining a reward.} Learning sequences of actions as macro-actions or options is important, e.g., in Hierarchical Reinforcement Learning (HRL). \shortciteA{options_sutton1999between} present a framework for temporal abstraction in RL to deal with such sequences.
Regarding the $Q^*$ value, this means that if an incomplete information state is used, then updates performed for approximating $Q^*$ will tend to assign partial credit also to incomplete sequences. The agent may not realise that a whole sequence of actions is needed to reach a rewarding $N$-state trajectory and not just a part of the sequence. While agents can perform well asymptotically in the face of both delays and sequences, using the correct information state would very likely lead to much better sample efficiency and more stable learning. 
In \textit{MDP Playground}, for discrete environments, only specific sequences of $N$-states of positive integer length $n$ are rewardable.
Sequences consist of non-repeating $N$-states allowing for $\frac{(\|NS\| - \|T\|)!}{(\|NS\| - \|T\| - n)!}$ sequences.
For the continuous environment of moving to a target, $n$ is variable.

\subsubsection{Reward Density} Environments can also be characterised by their \textit{reward density}. When an environment has denser rewards (see Figure \ref{fig:r_density_diag}), one is more likely to obtain a supervisory reward signal. In sparse reward settings \shortcite{tackling_sparse_rewards_raluca_2019}, the reward is $0$ more frequently, e.g., in continuous control environments where a long trajectory is followed and then a single non-zero reward is received at its end. This also holds true for the example of the tennis serve above.
In \textit{MDP Playground}, for discrete environments, the \textit{reward density}, $rd$, is defined as the fraction of possible $N$-state sequences that are actually rewarded by the environment.
For continuous environments, density is controlled by having a sparse or dense environment using a \textit{make\_denser} configuration option. When the reward is dense, the reward for the current time step is the distance travelled towards the target since the last step. When the reward is sparse, a reward is only handed out when the agent reaches the target.

With regard to density, recall the tennis serve again. The point received by serving an ace would be a sparse reward. We as humans know to reward ourselves for executing only a part of the sequence correctly. Rewards in continuous control tasks to reach a target point \shortcite<e.g. in Mujoco,>{todorov2012mujoco}, are usually dense (such as the negative squared distance from the target). This lets the algorithm obtain a dense signal in space to guide learning, and it is well known \shortcite{sutton2018reinforcement} that it would be much harder for the algorithm to learn if it only received a single reward at the target point. The environments in MDP Playground have a configuration option, $make\_denser$, to allow this kind of reward shaping to make the reward denser and observe the effects on algorithms. To achieve this, when $make\_denser$ is $True$ for discrete toy environments, the environment gives a fractional reward if a fraction of a rewardable sequence is achieved and it does so for all the possible rewardable sequences an agent is currently on.

\subsubsection{Stochasticity} Another characteristic of environments that can significantly impact performance of agents is \textit{stochasticity}. The environment, i.e., dynamics $P$ and $R$, may be stochastic or may seem stochastic to the agent due to partial observability or sensor noise (see Figure \ref{fig:p_noise_diag}-\ref{fig:r_noise_diag}). A robot equipped with a rangefinder, for example, has to deal with various sources of noise in its sensors \shortcite{probabilistic_robotics_burgard_book_2005}.
In \textit{MDP Playground}, for discrete environments, we define \textit{transition noise} $t\_n \in [0,1]$; with probability $t\_n$, an environment transitions uniformly at random to an $N$-state that is not the \textit{true} next $N$-state given by $P$.  We define \textit{reward noise} as being normally distributed: ${r\_n} \sim \mathcal{N}(0, {\sigma^2}_{r\_n})$, and is added to the \textit{true} reward. For continuous environments, both ${t\_n}$ and ${r\_n}$ are normally distributed and are directly added to the $N$-states and rewards.\footnote{The noise functions can actually be passed as Python lambda functions and the described normally distributed noise is just the default setting.}
\footnote{\chngdjair{An important detail to note is that in case the actions are noisy, the effects of these noisy actions are seen in the resulting $N$-state trajectory and this is why in \textit{MDP Playground}, we reward sequences of $N$-states and not sequences of $N$-states \textit{and} actions.}}

\subsubsection{Irrelevant Features} Environments also tend to have a lot of \textit{irrelevant features} \shortcite{babi_irrelevant_features_rajendran_2018} that one need not focus on. This holds for all kinds of learners including Neural Networks (NNs). However, NNs can even fit random noise \shortcite{nns_memorization_iclr_17} and having irrelevant features is likely to degrade performance. For example, in certain racing car games, even though the whole screen is visible, concentrating on only the road would be more efficient without loss in performance.
In \textit{MDP Playground}, for discrete environments, \chngdjair{a new discrete dimension is concatenated and appended to the $N$-state and action spaces each. This dimension causes the dimensionality of $P$, associated with the NMRDP, to increase to 2. The first dimension corresponds to dynamics relevant to the reward, $P_{rel}$, while the second dimension has dynamics $P_{irr}$. $P_{rel}$ and $P_{irr}$ are independent of each other and when concatenated together form $P$, the transition function visible to the agent. However, only the discrete dimension corresponding to $P_{rel}$ is \textit{relevant} to calculate the reward function.} Similarly, in continuous environments, new irrelevant dimensions are added to $NS$ and $A$ that are not considered relevant to the reward. These irrelevant dimensions are visible in the observations which come from $O$.

\subsubsection{Representations} 
\begin{figure}[ht]
        \centering
        \begin{subfigure}[]{0.19\textwidth}
            \centering
            \includegraphics[width=\textwidth]{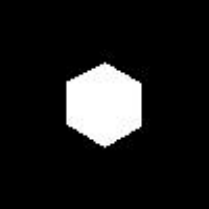}
            \caption[No transforms]%
            {{\small No transforms}}
            \label{fig:repr_learn_3_none}
        \end{subfigure}
        \begin{subfigure}[]{0.19\textwidth}   
            \centering 
            \includegraphics[width=\textwidth]{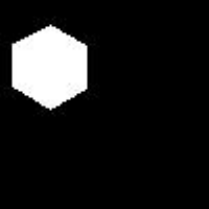}
            \caption[Shift]%
            {{\small Shift}}
            \label{fig:repr_learn_3_shift}
        \end{subfigure}
        \begin{subfigure}[]{0.19\textwidth}
            \centering
            \includegraphics[width=\textwidth]{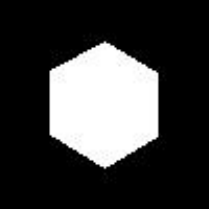}
            \caption[Scale]%
            {{\small Scale}}    
            \label{fig:append_repr_learn_3_scale}
        \end{subfigure}
        \begin{subfigure}[]{0.19\textwidth}   
            \centering 
            \includegraphics[width=\textwidth]{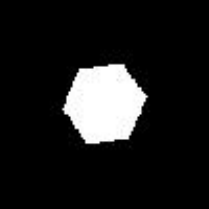}
            \caption[Rotate]%
            {{\small Rotate}}
            \label{fig:append_repr_learn_3_rotate}
        \end{subfigure}
        \begin{subfigure}[]{0.19\textwidth}
            \centering 
            \includegraphics[width=\textwidth]{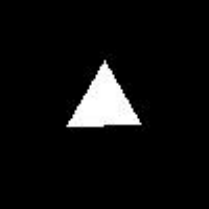}
            \caption[Flip]%
            {{\small Flip}}    
            \label{fig:append_repr_learn_0_flip}
        \end{subfigure}
        \begin{subfigure}[]{0.19\textwidth}
            \centering
            \includegraphics[width=\textwidth]{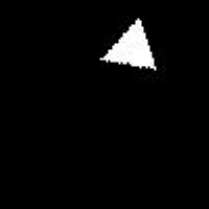}
            \caption[All transforms]%
            {{\small All transforms}}    
            \label{fig:append_repr_learn_0_all}
        \end{subfigure}
        \begin{subfigure}[]{0.38\textwidth}
            \centering
            \includegraphics[width=\textwidth]{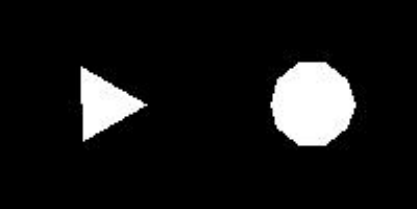}
            \caption[Image repr. irr. features]%
            {{\small Irrelevant features}}    
            \label{fig:image_repr_irr_features}
        \end{subfigure}
        \begin{subfigure}[]{0.38\textwidth}
            \centering
            \includegraphics[width=\textwidth]{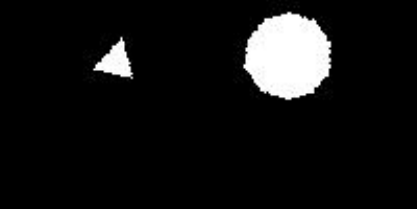}
            \caption[Image irrelevant features transforms]%
            {{\small Irrelevant features and transforms}}
            \label{fig:image_repr_irr_features_transforms}
        \end{subfigure}
        \begin{subfigure}[]{0.19\textwidth}
            \centering
            \includegraphics[width=\textwidth]{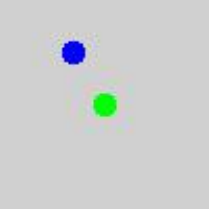}
            \caption[Cont. env.]%
            {{\small Agent + Target}}    
            \label{fig:cont_state_target}
        \end{subfigure}
        \begin{subfigure}[]{0.19\textwidth}
            \centering
            \includegraphics[width=\textwidth]{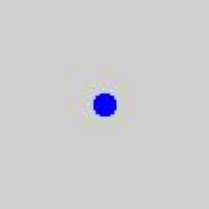}
            \caption[Cont. env.]%
            {{\small Agent at Target}}    
            \label{fig:cont_state_agent_at_target}
        \end{subfigure}
        \begin{subfigure}[]{0.19\textwidth}
            \centering
            \includegraphics[width=\textwidth]{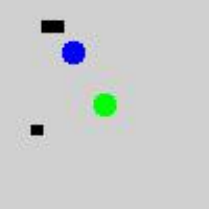}
            \caption[Cont. env.]%
            {{\small Terminal}}    
            \label{fig:cont_state_target_terminal_states}
        \end{subfigure}
        \begin{subfigure}[]{0.38\textwidth}
            \centering
            \includegraphics[width=\textwidth]{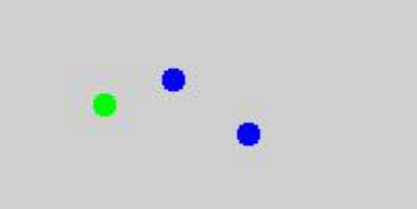}
            \caption[Cont. env.]%
            {{\small Irrelevant Features}}
            \label{fig:cont_env_irrelevant_image}
        \end{subfigure}

        \caption[ Various image representations ]
        {(a)-(f): \textbf{Discrete environments}: When using the dimension of hardness \textit{representations} with image representations, each categorical $N$-state corresponds to an image of a polygon (if the states were numbered beginning from $0$, each $N$-state $n$ corresponds to a polygon with $n + 3$ sides). Various transforms can be applied to the polygons randomly at each time step. Samples shown correspond to $N$-states $3$ and $0$. (g)-(h): \textbf{Discrete environments with irrelevant features}: When irrelevant features are enabled alongside image representations, an additional ``irrelevant'' image is stitched to the right of the relevant part and the same transforms are applied to each part. (i)-(k): \textbf{Continuous environments}: When using the dimension of hardness \textit{representations} with image representations in continuous environments, the agent is shown as a blue circle, the target point as a green circle and terminal $N$-states are black. The agent is rendered after the target as seen in the second sub-figure. \chngdjaircrc{(l): \textbf{Continuous environments with irrelevant features}: As for the discrete environments, the relevant part with $2$ dimensions is the left half of the image and the irrelevant is the right half. The agent exists in all $4$ dimensions and is visible in both halves of the image while the target exists in only the $2$ relevant dimensions and is visible only in the left half.}}
        \label{fig:repr_learn_images}
\end{figure}

Another aspect is that of \textit{representations}. The same underlying $N$-state may have many different external representations/observations from $O$, e.g., \textit{feature} space would be the $N$-space while \textit{pixel} space could be $O$ in Mujoco. Similarly, Atari games can use the underlying RAM state or images. For images, various image transformations \shortcite<\textit{shift}, \textit{scale}, \textit{rotate}, \textit{flip} and others; >{imagenet_c_hendrycks_iclr_2019} may manifest as observations of the same underlying $N$-state and can pose a challenge to learning. In \textit{MDP Playground}, for discrete environments, when image representations are enabled, each categorical $N$-state is associated with an image of a regular polygon which becomes the externally visible observation $o \in O$ to the agent (see Figure \ref{fig:repr_learn_3_none}-\ref{fig:image_repr_irr_features_transforms}). This image can be further transformed by \textit{shifting}, \textit{scaling}, \textit{rotating} or \textit{flipping}, which are applied at random to the polygon whenever an observation is generated. For continuous environments, image observations can be rendered for $2$D environments (see Figure \ref{fig:cont_state_target}-\ref{fig:cont_env_irrelevant_image}).

\subsubsection{Diameter} The \textit{diameter} of an MDP, i.e., the maximum distance between 2 states, is another significant dimension of hardness affecting performance and reachability of states \shortcite{jaksch_theory_regret_bounds_diameter_jmlr_2010}. In our case, we define the diameter for the states of the NMRDP, i.e., the $N$-states. In \textit{MDP Playground}, for discrete environments, for $diameter = d$, the set of $N$-states is set to be a $d$-partite graph, where, if we order the $d$ sets as $1, 2, .., d$, $N$-states from set $i$ will have actions leading to $N$-states in set $i+1$, with the final set $d$ having actions leading to $N$-states in set $1$. The number of actions for each $N$-state will, thus, be $(\textit{number of states}) / (d)$. This gives the discrete environments a grid-world like structure. For continuous environments, setting the dimension of hardness \textit{state space max} sets the bounds of the $N$-state space to $\pm \textit{state space max}$ and the $diameter = 2\sqrt{2} \,\, \textit{state space max}$.

\subsubsection{Time Unit and Action Range} For continuous control problems, we now describe two dimensions of hardness together. Firstly, \textit{action range} \shortcite{kanervisto_action_space_shaping_cog_2020}, or the action space, is the range of actions that the agent may take. \chngdjaircrc{For example, in the \textit{Gym} environment \textit{Pendulum}, the action applied is a scalar representing the torque applied to the free end of the pendulum and its range is $(-2.0, 2.0)$. The action range can be expected to have a significant impact on an agent's performance.} \chngdjair{For example, if the action range is too large, the exploration may take too long to find the optimal actions. On the other hand, if, for example, one is designing an MDP to solve a problem, if the action range is set to be too small, it may not contain the desired actions one is hoping to optimise for.} Similar to the action range, the \textit{time unit}, the discretisation of time (see Figure \ref{fig:time_unit_diag}) can be expected to have a significant impact on an agent's performance. For example, \shortciteA{biedenkapp-icml21} shows that there is an optimal number of times an action needs to be repeated. Similar to the action range, having too small a time unit would not capture such optima and having too large a time unit would lead to a much harder exploration problem. The \textit{time unit}, $t$, sets $P(ns, a) = ns + \int_{t} P_{cont}(ns, a)\; dt$ where $P_{cont}$ is the time-derivative of the dynamics function. The \textit{action range} sets $A \subset \mathbb{R}^a$ where $a$ is the action space dimensionality. \chngdjair{In order to avoid any confusion, we must clarify here that it is not our intention to suggest that standardised or benchmark problem definitions be changed to show new state of the art performances because changing the problem would invalidate the benchmark. What we want to say is that: 1) if the problem definition is given, e.g., a fixed action range, one might want to search in a subset of this range to learn quicker; the RL algorithm could decide to do so on a meta-level as is done with, e.g., \textit{reward shaping} on a meta-level \shortcite{zheng_2018_learning_intrinsic_rewards_pg,hu_neurips_2020_learning_utilize_shaping_rewards,zou_aaai_2021_learning_task_distribution_reward_shaping}; one could call it \textit{action space shaping} when doing it with the action space; 2) in some cases, the problem at hand may not have been defined in a desirable manner, leading to solutions which the people trying to solve the problem may not desire, and in such cases one might want to, e.g., in the case of a fixed action range, search outside this range if physically possible. This second case is more likely to occur for completely unknown or new environments where the problem definition itself may be unclear. We believe there is a difference between the problem people want to solve and the MDP formulation they use to write their problem in a way for standard RL agents to work. For example, the discount factor, $\gamma$, is frequently treated as a hyperparameter even though it is part of the MDP formulation and changing it changes the MDP being solved \shortcite{xu_2018_meta_gradient_rl,tessler_2020_tuning_gamma,franccois_2015_tuning_gamma}. As such, \shortciteA{xu_2018_meta_gradient_rl} tuned $\gamma$ with meta-gradients to try and solve their target problem. Another example is that of \shortciteA{runge_iclr_2019_learna}. They aimed to learn to design RNA structures, and formulated their MDP encoding such that it had hyperparameters that modified the state space and shaped the reward function. These hyperparameters of the MDP could be tuned alongside the RL algorithm’s hyperparameters in order to find a combination of MDP encoding and RL algorithm that solved the problem as well as possible; 3) finally, one might just want to play around with problem definitions to observe the effects on their agents, we do not propose that researchers release such results claiming that they improved the performance on the original benchmark problem, but such studies can still yield insights into the RL agent's behaviour.}

\subsubsection{Target Radius} Finally, an additional dimension of hardness for continuous control problems we describe is \textit{target radius} \shortcite{klink_self_paced_rl} which is a measure of the distance from the target within which we consider the target to have been successfully reached. This dimension of hardness is a fairly common use-case in continuous control environments, e.g., in \textit{OpenAI Gym}. The \textit{target radius} sets $T = \{ns\mid\|ns - ns_t \|_{2} < \textrm{\textit{target radius}}\}$, where $ns_t$ is the target point.



We now summarise the dimensions of hardness identified above (with the (PO)MDP component they directly impact in \textit{MDP Playground} in brackets):

\begin{minipage}{.320\textwidth}
\begin{itemize}
\denselist
\item Reward Delay ($R$, $MS$)
\item Sequence Length ($R$, $MS$)
\item Reward Density ($R$)
\item Stochasticity ($P$, $R$)
\end{itemize}
\end{minipage}
\begin{minipage}{.320\textwidth}
\begin{itemize}
\denselist
\item Irrelevant Features ($O$)
\item Representations ($O$)
\item Diameter ($P$, $NS$)
\end{itemize}
\end{minipage}%
\begin{minipage}{.320\textwidth}
\begin{itemize}
\denselist
\item Time Unit ($P$)
\item Action Range ($A$)
\item Target Radius ($T$)
\end{itemize}
\end{minipage}%

Only selected dimensions of hardness are included here, to aid in understanding and to show use-cases for \textit{MDP playground}. Trying to identify as many dimensions as possible has led to a very flexible platform and Table \ref{tab:default_dimension_values} lists all the dimensions of hardness and important configuration options of \textit{MDP Playground}. We would like to point out that what dimensions of hardness are important largely depends on the domain. For instance, in a video game domain, a practitioner may not want to inject any kind of noise into the environment, if their only aim is to obtain high scores, whereas in a domain like robotics adding such noise to a deterministic simulator could be crucial in order to obtain generalisable policies \shortcite{domain_rand_tobin_iros_17}. To maintain the flexibility of having as many dimensions of hardness as possible and yet keep the platform easy to use, \textit{default} values are set for dimensions that are not configured. This effectively \textit{turns off} those dimensions of hardness.

\section{MDP Playground}\label{sec:mdpp}



\textit{MDP Playground} consists of $4$ components: $1)$ Toy Environments $2)$ Complex Environment Wrappers $3)$ Experiments $4)$ Analysis. We now briefly describe the first two parts and describe how to experiment and analyse with \textit{MDP Playground} in detail in Section \ref{sec:experiments}.

\subsection{Toy Environments} The toy environments \chngdjaircrc{are \textit{Gym environments}. They} are cheap and encapsulate all the identified dimensions of hardness. The components of the MDP can be automatically generated according to the configured dimensions of hardness or they can be user-defined. We discuss how the automatically-generated MDPs function in detail in Section \ref{sec:mdpp:detailed} and the corresponding algorithm in Algorithms \ref{algorithm:generation_disc} and \ref{algorithm:generation_cont} in Appendix \ref{append_sec:algorithms}. Further, the underlying Markovian $M$-state is exposed in an \textit{augmented\_state} variable, which allows users to design agents that may try to identify the true underlying MDP $M$-state given the observations. Design decisions regarding the MDPs are discussed in detail in Section \ref{sec:design_decisions}.

\subsection{Complex Environment Wrappers} We further provide wrappers for \textit{Gym} environments which can be used to inject many of the dimensions of hardness into complex environments such as \textit{Atari}, \textit{Mujoco}, \textit{ProcGen} \shortcite{cobbe_procgen_icml_20} or any other \textit{OpenAI Gym} environment. These can be used to analyse agent behaviour in response to such dimensions of hardness.

\subsection{Code Sample}

\begin{wrapfigure}[16]{l}{.55\textwidth}
\begin{minipage}{.55\textwidth}
\begin{lstlisting}[language=Python]
from mdp_playground.envs import 
RLToyEnv, GymEnvWrapper
config = {
    'state_space_type': 'discrete',
    'action_space_size': 8,
    'delay': 1,
    'reward_noise': 0.25,
    }
env = RLToyEnv(**config)

ae = gym.make("QbertNoFrameskip-v4")
env = GymEnvWrapper(ae, **config)
\end{lstlisting}
\end{minipage}
\end{wrapfigure}

An environment instance is created as easily as passing a Python \texttt{dict}. \chngdjair{Users can set the values of dimensions in the \texttt{dict} with one line of code for each dimension.} As mentioned before, users only need to provide dimensions they are interested in. \chngdjair{The rest are set to \textit{vanilla} values. By \textit{vanilla}, we mean the dimension being set to its default value which turns it off in the case of most dimensions, e.g., setting delay, $d = 0$ or sequence length, $n = 1$. Please see Table \ref{tab:default_dimension_values} for a list of default values of the dimensions.} In the code sample below, we create a toy \textit{and} a complex environment using the wrapper with the same dimensions of hardness.

\subsection{Experiments and Analysis} The GitHub repository\footnote{\url{https://github.com/automl/mdp-playground}} describes how to run experiments and how to analyse them with plots in a Jupyter Notebook. Section \ref{sec:experiments} describes some experiments and analyses of this kind.

\subsection{Very Low-Cost Execution} Toy experiments with \textit{MDP Playground} are cheap, allowing academics without special hardware to perform insightful experiments. Wall-clock times depend a lot on the agent, network size (in case of NNs) and the dimensions of hardness used. Nevertheless, to give the reader an idea of the runtimes involved, DQN delay experiments from Section \ref{sec:experiments} performed using Ray RLLib~\shortcite{ray_liang2017rllib} with a network with $2$ hidden layers of $256$ units each) took on average $35$s for a \textit{complete} run of DQN for $20\,000$ environment steps.
In this setting, we restricted Ray RLLib and the underlying Tensorflow to run on \textit{one core of a laptop} (core-i7-8850H CPU --
the full CPU specifications can be found in Appendix \ref{sec:cpu_info}). This equates to roughly $300$ minutes for the \textit{entire} delay experiment shown in Figure \ref{fig:DQN_del_1d} which was plotted using $500$ runs ($100$ seeds $\times$ $5$ settings for \textit{delay}; these 500 runs could also be run in an embarrassingly parallel manner on a cluster e.g. in less than $2$ minutes with $500$ CPU-cores in total with one core for each seed). 
Even when using the more expensive continuous or image representation environments, runs were only about $3$-$5$ times slower.

\chngdjair{
\section{MDPs in MDP Playground}
\label{sec:mdpp:detailed}
In this section, we provide a more \textit{bottom-up}/\textit{implementation}-oriented presentation of \textit{MDP Playground} and its dimensions of hardness. We begin with describing the discrete toy environments (corresponding algorithm in Algorithm \ref{algorithm:generation_disc} in Appendix \ref{append_sec:algorithms}). We then describe the continuous toy environments (corresponding algorithm in Algorithm \ref{algorithm:generation_cont} in Appendix \ref{append_sec:algorithms}). Finally, we describe the complex environment wrappers.

\subsection{Discrete Toy Environments}
In the discrete toy environments, $NS$ consists of the set of states of the system, i.e., it contains no history. For $diameter = d$, $NS$ is set to be a $d$-partite graph, where, if we order the $d$ sets as $1, 2, .., d$, $N$-states from set $i$ will have actions leading to all $N$-states in set $i+1$, with the final set $d$ having actions leading to all $N$-states in set $1$. The number of actions for each $N$-state is, thus, $(\textit{number of states}) / (d)$. In setting the configuration, the user sets \textit{action space size} (which is equal to the size of each independent set) and the \textit{diameter}, which indirectly sets the size of $NS$. The $N$-states are labelled from $0$ to $\|NS\| - 1$ which we term \textit{global labelling}. The $N$-states in each independent set are labelled from $0$ to $\|NS\|/d - 1$ within that set ( this is obtained by modulo division of the global label with $\|A\|$ which we term  \textit{local labelling}. They are \textit{categorical} in that they are not ordered. As such, elements of both $NS$ and $A$ are 1-dimensional integers. In the case of the vanilla environment, $N$-states, $M$-states and observations coincide. The terminal $N$-state density can also be set by the user. The initial state distribution, $\rho_0$, is currently fixed to be uniform over the non-terminal $N$-states. The environments' episodes are restricted to a maximum of 100 timesteps by default. As an example, Figure \ref{fig:discrete_toy_env} shows a discrete toy environment with 6 $N$-states in total, partitioned into sets of 2 each. Please see the text in the figure caption for details on the example.

Based on the value the \textit{reward density} $rd$ and the \textit{sequence length} $n$ are set to, automatic generation of the MDP leads to randomly selecting $\lfloor rd * \textit{number of possible sequences} \rfloor$ (where $\lfloor \, \rfloor$ denotes the floor function) of non-repeating $N$-state sequences of length $n$ to be rewardable. While there are transitions possible to terminal $N$-states, the reward density and sequence lengths only consider non-terminal states to determine rewarding sequences, allowing for $\frac{(\|NS\| - \|T\|)!}{(\|NS\| - \|T\| - n)!}$ sequences. The reward handed out for following a rewardable sequence is $1$. Based on the value the reward delay $d$ is set to, we delay the reward at any timestep to be handed out $d$ steps later. In the case of reward delay and sequence lengths, the $N$-states remain the same as for the vanilla environment, but the $M$-states consist of $N$-state sequences of length $n$ from $d$ timesteps in the past. The observations remain the same as the $N$-states. Finally, when $make\_denser$ is $True$ for discrete toy environments, the environment gives a fractional reward if a fraction of a rewardable sequence is achieved and it does so for all the possible rewardable sequences an agent is currently on.

An additional point to note about episodic rewards in the presence of \textit{delay} and \textit{sequence length} in the toy environments is that the optimal episodic reward changes in the presence of these dimensions of hardness. In the case of delays, an agent cannot get the delayed reward at the end of an episode if the episode terminates before the delayed reward is handed out. In the case of sequence lengths, it is a bit tricky to handle rewardable sequences because the agent might be on multiple rewardable sequences at the same time and it is a very hard problem to determine the optimal episodic reward for larger sequence lengths in such cases. Since we are usually concerned only with agents trying to execute non-overlapping rewardable sequences, we provide the option \textit{reward\_every\_n\_steps} which is turned on by default. With this option, we hand out rewards for executing a rewardable sequence only if it finishes executing in a timestep of the environment that is divisible by $n$ if the sequence length is $n$ and there is no reward handed out if a rewardable sequence is completed at other timesteps (this does not affect other dimensions, e.g., there may be independent reward noise after every step or a terminal state reward at the end of the episode as described later). This means that even if the agent is on multiple rewardable sequences at once, they only get rewarded for one of them and the optimal episodic reward for the toy environment with 100 steps is $100/n$. \chngdjaircrc{For example, suppose the sequence length were $4$. If an agent in $N$-state $ns_0$ at timestep $0$ started executing a rewarding sequence and then reached $N$-states $ns_1, ns_2, ns_3$ which are all also the starting points of other rewarding sequence, it could then be on $4$ rewarding sequences at once for the given sequence length of $4$. If the agent successfully executes all $4$ sequences, it could be rewarded at timesteps $4, 5, 6$ and $7$ (if \textit{reward\_every\_n\_steps} were \textit{False}). This complicates the optimal reward calculation because given all the generated rewarding sequences, it would be hard to calculate the optimal sequence of overlapping rewarding sequences the agent needs to be on. However, if \textit{reward\_every\_n\_steps} were \textit{True}, the agent would only be rewarded after every $n^{th}$ step, so only after step $4$ in the above example. As a result, the optimal reward calculation in this case is fairly easy. It would be the episode length, i.e. 100, divided by $n$. This holds even when the diameter is $>1$, because an equal number of rewarding sequences per independent set of $N$-states is selected. However, this does change the reward scale of the optimal episodic reward.} To combat these changing episodic reward scales, we normalise the plots to have episodic reward scales of 100. This normalisation will be described in Section \ref{sec:experiments}.

\textit{Transition noise} $t\_n$ sets the fraction of times we transition to a connected $N$-state that is not the true next $N$-state as defined by the transition function $P$ for the current $N$-state, $ns$ and action $a$. \chngdjaircrc{In the presence of \textit{diameter}, this means that the next $N$-state belongs to the next independent set}. \textit{Reward noise} $\sigma_{r\_n}$ adds a normally distributed reward with standard deviation $\sigma_{r\_n}$ to the obtained reward at each time step. Stochasticity, thus, affects all of $N$-states, $M$-states and the observations.\footnote{Since the noise generation process is based on a pseudo-random number generator (PRNG), one could in principle claim that the $M$-state should, in the case of stochasticity have, additionally, the state of the PRNG as well because the next $M$-state depends on it. However, our intent is to treat the noise as \textit{true} noise and, as such, we ignore this technicality.}.

When \textit{irrelevant features} are turned on, the user sets the \textit{action space size} with a list of length 2. The first number in the list is used as the normal/relevant action space size as specified above. The second number is used in the same way to set the irrelevant action space size and the size of the irrelevant $N$-state space size using the same diameter as for the relevant spaces. The $N$-states and observations have the additional irrelevant dimension appended to the original $N$-states and observations, respectively. The $M$-states in this case are unchanged from when irrelevant features would be turned off. The dynamics function associated with the sub-space relevant to the reward, $P_{rel}$ and the dynamics function associated with the sub-space irrelevant to the reward, $P_{irr}$, are independent of each other. They are concatenated together to form the dynamics function, $P$ associated with the NMRDP. The dynamics associated with the MDP, however, are only $P_{rel}$\footnote{One could, naturally, also change the $M$-states and MDP to include the irrelevant features and still remain Markovian. However, in our MDP, we have decided to retain the minimal amount of information needed to obtain rewards.}. The irrelevant sub-space also does not have any terminal states. Finally, the same transition noise is applied to the irrelevant sub-space as the relevant sub-space.


When turning on \textit{image representations}, the observations become 2-dimensional greyscale images. The integer representation, $n$, of the underlying $N$-state space is associated with a polygon of $n + 3$ sides as shown in Figure \ref{fig:repr_learn_images} and this is used as the observation\footnote{Since the above image representations might only be useful for vertex detectors of polygons, we allow users to provide a directory with custom images that can be used as the representations for the $N$-states in case they have better ideas to test the representation learning aspect of their algorithm.}. This polygon is displayed in the centre of a $100 \times 100$ pixel image with the radius of the circle which circumscribes the polygon as $20$ pixels. Turning on \textit{image representations} changes only the observations. All of the dimensions of hardness discussed so far remain largely the same, otherwise. For example, the transition noise is only applied when transitioning within the underlying $N$-state space. There is no \textit{additional} noise in the image space, when transition noise is turned on, due to image representations also being turned on. In the case of irrelevant features also being turned on at same time as image representations, however, it may be ambiguous as to how image representations of the relevant and irrelevant sub-spaces are concatenated. In this case, we do so along axis 0, i.e., the “X-axis” of the image (in other words, width-wise) \chngdjaircrc{with the relevant part always being on the left side of the image}. \chngdjaircrc{Additionally, in the case when the \textit{diameter} is $>1$, there is more than one independent set and it may be ambiguous as to whether the global or the local label is used to create the polygons in the default image representations. We do so using the global label to be able to distinguish different $N$-states visually.}

As image observations are associated with different kinds of transforms such as \textit{shift}, \textit{scale}, \textit{rotate} and \textit{flip}, we allow these transforms to be applied \textit{when image representations are turned on}. While the default observation is of a polygon centred in the image with a fixed size, further turning on these transforms results in the observations associated with the \textit{same} underlying $N$-state to vary as shown in sub-figures \ref{fig:repr_learn_3_shift}-\ref{fig:append_repr_learn_0_all}.\footnote{We take care that the polygon does not translate out of the image and also print warnings in case the polygon size ends up being “too small” or “too large”.} This allows the representation learning aspect of the environment to potentially also be made harder. We also allow the quantisation of the \textit{shift} and \textit{rotate} transforms and the range of the \textit{scale} transform to be adjusted to have more fine-grained control over these. Finally, in the case of irrelevant features also being turned on at the same time as transforms for image representations, we apply the same transforms to the irrelevant part of the image as to the relevant part and then concatenate along the “X-axis” of the image \chngdjaircrc{with the relevant part always being on the left side of the image}.

\setcounter{footnote}{10}

\begin{figure}[ht!]  
        \centering
            \includegraphics[width=0.83\textwidth]{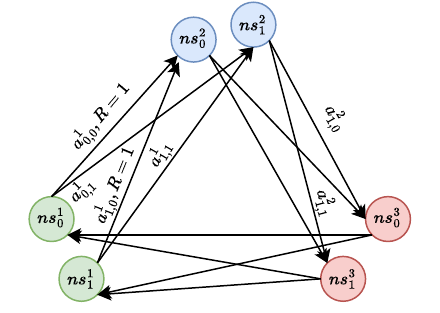}

        \caption[ Discrete toy environment ]
        {An example discrete toy environment with 6 $N$-states with diameter = 3\footnotemark, divided into 3 independent sets with 2 $N$-states each. Each $N$-state has 2 possible actions in its action space. The $N$-states have the set index for the set they belong to as superscript and the $N$-state index \textit{within} the set (the \textit{local label} as described in the text) as subscript. The actions have, additionally, the action index for the $N$-state they belong to as a second subscript. From each $N$-state in set $1$, one can take an action to enter each $N$-state in set $2$. The reward density is set to $1/6$ and the sequence length to $1$. Based on these settings, $N$-state sequences of length 1 would be rewardable, i.e., one needs to reach a rewardable $N$-state to obtain a reward of 1. Automatic generation of the MDP leads to randomly selecting 1 out of the 6 $N$-state sequences of length 1 being picked as rewardable. In this example, this leads to the only rewarding transitions being from $ns^{1}_{0}$ to $ns^{2}_{0}$ and $ns^{1}_{1}$ to $ns^{2}_{0}$ with the rewardable sequence of $N$-states being $ns^{2}_{0}$. The edges to it are labelled with reward $R = 1$, all other rewards are $0$. Not all edges are labelled to avoid clutter. Not all dimensions of hardness and no terminal $N$-states are included here for simplicity.}
        \label{fig:discrete_toy_env}
\end{figure}
\footnotetext{The diameter is 3, and not 2 as may seem at first sight, because the route from a starting vertex to the other vertices in the same set are of minimum length 3.}

Finally, the \textit{reward scale} scales the (possibly noisy) reward at each time step. After this, the \textit{reward shift} is added to the reward at each time step. The \textit{terminal state reward} scaled by the \textit{reward scale} is added to the reward at the end of the episode.

The \textit{Gym} environment returns observations as is usual. We, additionally, also return the whole $N$-state sequence starting from $d + n + 1$ time steps in the past in \textit{Gym}'s \texttt{info} variable, so that the agent might use these to validate its learning in different ways, e.g., checking if it can create the underlying $M$-state given the observation.

\subsection{Continuous Toy Environments}
We now describe the continuous toy environments in \textit{MDP Playground}.

The continuous toy environments consist of a point mass which is controllable using the agent's actions. The user can set the mass of the system using the \textit{inertia} configuration of the system. Similar to the discrete environments, $NS$ consists of the (non-Markovian) state of the system. In the case of the continuous toy environment, the elements of $NS$ and $A$ belong to $\mathbb{R}^{n_{dim}}$ where $n_{dim}$ is the dimensionality of the system, which is set by the user. The user also sets the bounds of the system by setting the \textit{state space max} and \textit{action space max} options. The dimensionality of $NS$ and $A$ is equal and each action dimension controls each $N$-state space dimension. The user can set the \textit{transition dynamics order}, $n_P$, of the system. This sets an action $a$, when applied, to set the $n_P^{th}$ order derivative of the position of the mass divided by the inertia for \textit{time unit} seconds. In the simplest case of $n_P = 1$ and $inertia = 1$, this sets the velocity of the point mass. Or, for example, in the case of $n_P = 2$, this sets the acceleration of the point mass. The \textit{time unit}, $t$, then sets $P(ns, a) = ns + \int_{t} P_{cont}(ns, a)\; dt$ where $P_{cont}$ is the time-derivative of the dynamics function. In this case where $a$ is constant for a fixed timestep, for current $N$-state, $ns$, the equations of motion can be analytically derived \chngdjaircrc{in closed form to be $ns^i_{t+1} = \sum\limits^{n_P-i}_{j = 0} ns^{i+j}_{t} \cdot \frac{1}{j!} \cdot time\_unit^{j}$ where an integer in the superscript represents the order of the derivative and not exponentiation}. For example, if $n_P = 1$, the equation is $ns = ns + ns^1 * t$, i.e., the next $N$-state is equal to the current $N$-state plus $ns^1$ where $ns^1 = a/inertia$, is set equal to the action, $a$. As in the case of discrete toy environments, the continuous toy environments' episodes are restricted to a maximum of 100 timesteps by default.

The reward function of the continuous environments is such that there is a \textit{target point} that needs to be reached. The difference between the distance of the current $N$-state from the target point and the previous $N$-state from the target point is handed out as the reward at each time step.\footnote{There is an additional reward function, \textit{moving along a line} as opposed to moving to a (target) point but this is still in the experimental phase. This is an attempt at having a \textit{sequential} task in the continuous environment and is described in the experimental dimensions section in Appendix \ref{append_sec:exptl_dims}.} Terminal $N$-state regions can be set by the user by providing \textit{terminal\_states} which \chngdjaircrc{are $n_{dim}$-dimensional points that specify centres of hypercube sub-spaces. Entering these hypercubes ends the episode.\footnote{\chngdjaircrc{The reason sub-spaces are used to implement terminal states in the continuous toy environments, as opposed to randomly sampling them from the whole $N$-state space as is the case with the discrete toy environments, is that, in principle, randomly sampling terminal states in continuous spaces would mean they are virtually impossible for the agent to reach. In practice, though, due to the finite precision of floating point numbers, one could contemplate doing the latter. However, we believe it would be a lot of overhead for little gain and so decided to not implement it in this manner.}}} The length of the side of the hypercube is set by \textit{term\_state\_edge}. The initial $N$-state of the system is set to be sampled uniformly from non-terminal regions of the space.

The \textit{reward delay} in the continuous toy environments works as in the discrete ones. The \textit{sequence length} is variable for the task of reaching a target point and depends on the actions taken, so it cannot be set by the user. The \textit{reward density} can be controlled in the sense of whether the reward is handed out after each time step or only once the target point is reached. This is done by turning the configuration option $make\_denser$ on or off. The \textit{transition noise} differs from the discrete environments. The \textit{transition noise}, $\sigma_{t\_n}$, and \textit{reward noise}, $\sigma_{r\_n}$, both add normally distributed noise with standard deviation $\sigma_{t\_n}$ or $\sigma_{r\_n}$ to the $N$-state or reward, respectively, at each time step. Stochasticity, thus, affects all of $N$-states, $M$-states and the observations.

\textit{Irrelevant features} can be specified by using the \textit{relevant indices} option and specifying a dimensionality of the system, $n_{dim}$, which is greater than the number of relevant indices. So, the continuous environments can have more than 1 relevant dimension and more than 1 irrelevant dimension. In contrast to the discrete toy environment, the relevant and irrelevant parts have the same transition dynamics. The only difference between the relevant and irrelevant parts in the continuous case is that the former are considered for calculating the reward, i.e., the dimensionality of the target point is equal to the number of relevant dimensions and the agent needs to try and reach the target point in the relevant sub-space of the system. The irrelevant sub-space also does not have any terminal regions. Irrelevant features, thus, affects the $N$-state and observations but the $M$-states are unchanged from when irrelevant features would be turned off.

\textit{Image representations} are only available for 2-dimensional spaces. They represent the point mass as a blue circle with radius 5 pixels and the target point as a green circle with radius 5 pixels. Terminal regions are represented as black and the background is grey as shown in Figures \ref{fig:cont_state_target}-\ref{fig:cont_env_irrelevant_image}. They only change the observations of the system and do not affect the $N$-states or the $M$-states. In case irrelevant features are also turned on, as in the case of the discrete environments, there are relevant and irrelevant parts to the image that are stitched together along the ``X-axis'' with the relevant part on the left (each part is fixed to be 2-dimensional). We do not allow transforms for image representations for continuous environments as there would be no way to distinguish between an action and a \textit{shift} transform. 

The \textit{target radius} is the distance from the \textit{target point} where it is considered to have been reached. In addition to the user set terminal regions, reaching the target point ends the episode. The \textit{action space max} option mentioned above sets the \textit{action range} of the system. The user can additionally also opt to add an \textit{action loss weight} to penalise the L2-norm of the actions taken during the control task. This subtracts the \textit{action loss weight} multiplied by the L2-norm of the action from the reward at the current time step.

Finally, the \textit{reward scale, reward shift} and \textit{terminal state reward} work in the same way as for the discrete toy environments.

\subsection{Complex Environment Wrappers}
We now describe the complex environment wrappers which allow a subset of the dimensions of hardness of \textit{MDP Playground} to be controlled in complex environments such as \textit{Atari, Mujoco} and \textit{Procgen}. They function in largely the same way as the toy environments, however, they only implement a subset of the dimensions of hardness. An overview of which of these are implemented for complex environments can be found in Table \ref{tab:default_dimension_values}.

The \textit{reward delay} works the same way as in the toy environments, except towards the end of the episode, when any delayed reward that was not yet handed out is handed out at the end of the episode. This is to maintain the same episodic reward scales for the environment when comparing across delays. For the toy environments we decided to normalise the episodic rewards post hoc (i.e., only in the plots) and not hand out all the delayed reward at the end of an episode because 1) we know the optimal rewards for the toy environments whereas we do not know the optimal rewards for complex environments; 2) rewards in the toy environment experiments we performed were much denser than the complex environments and handing out all the delayed reward at once in the toy environment would bias the terminal samples stored in the replay buffer a lot compared to the complex environments. 

The \textit{transition noise} for discrete environments is applied by choosing an action that is not the selected action from the action space.\chngdjaircrc{\footnote{\chngdjaircrc{The transition noise for the discrete complex environments wrappers is modelled in the action space and not in the state space (as was the case for the toy environments), as it is not possible to know all the possible next discrete states in advance for the complex environments.}}} The \textit{transition noise} for continuous environments and the \textit{reward noise} for both discrete and continuous environments are applied in the same way as for the toy environments. The \textit{irrelevant features} for continuous environments creates a toy continuous environment (with the same dynamics as the ones described above and with its own dimensions of hardness which are set to their defaults if not set) whose $N$-states are appended to the $N$-states of the continuous environment for every transition and whose reward is ignored. The \textit{reward scale, reward shift} and \textit{terminal state reward} work in the same way as for the toy environments.

The \textit{action range} and \textit{time unit} wrappers are specific to \textit{Mujoco}. For both of these dimensions of hardness, the scalar value passed in the configuration is used to multiply the base environments' default values for that dimension of hardness. With respect to the \textit{action space max} (which controls the \textit{action range}), the new action range is achieved by multiplying the base \textit{Gym Mujoco} environments' \texttt{action\_max} and \texttt{action\_min} by the \textit{action space max} passed in the configuration for the wrapper. This works with any \textit{Mujoco} environment.

\begin{table}
 \caption{Dimensions of hardness and additional configuration options in MDP Playground with default values. For the wrappers, $+$ next to a \checkmark signifies that the dimension only works on the \textit{continuous} case, $*$ that it only works on \textit{continuous Mujoco} environments. (\textit{Note}: Formatting and colour of the text: \textbf{bold} dimensions signify applicability on both continuous and discrete toy environments, plain text: only on discrete toy, \textit{italic}: only on continuous toy, coloured ones have wrappers implemented: \blue{blue}: the wrappers can be used any environment, \red{red}: only with \textit{continuous} environments, \green{green}: only with \textit{Mujoco} environments.)}
 \label{tab:default_dimension_values}
\begin{center}
 \begin{tabular}{|| p{0.31\textwidth} | c | c | c | p{0.29\textwidth} ||} 
 \hline
 Dimension of Hardness & Discrete & Continuous & Wrapper & Default \\ [0.5ex] 
 \hline\hline
 \textbf{\blue{Reward Delay}} & \checkmark & \checkmark & \checkmark & 0 \\ 
 \hline
 \text{Sequence Length} & \checkmark & & & 1 \\
 \hline
 \text{Reward Density} & \checkmark &  &  & 0.25 \\
 \hline
 \textbf{\blue{Transition Noise}} & \checkmark & \checkmark & \checkmark & 0 \\
 \hline
 \textbf{\blue{Reward Noise}} & \checkmark & \checkmark & \checkmark & 0 \\
 \hline
 \textbf{\red{Irrelevant Features}} & \checkmark & \checkmark & \checkmark+ & False \\
 \hline
 \textbf{Image Representations} & \checkmark & \checkmark & & False \\
 \hline
 Image Transforms & \checkmark &  &  & None \\
 \hline
 \textbf{\blue{Reward Scale}} & \checkmark & \checkmark & \checkmark & 1 \\
 \hline
 \textbf{\blue{Reward Shift}} & \checkmark & \checkmark & \checkmark & 0 \\
 \hline
 \textbf{Make denser} & \checkmark & \checkmark &  & False \\
 \hline
 \text{$N$-State space size}, $\|NS\|$ & \checkmark &  &  & 8 \\
 \hline
 \text{Action space size}, $\|A\|$ & \checkmark &  &  & 8 \\
 \hline
 \text{Reward every n steps} & \checkmark &  &  & True \\
 \hline
 \text{Diameter} & \checkmark &  &  & 1 \\
 \hline
 \text{Terminal state density} & \checkmark &  &  & 0.25 \\
 \hline
 \textbf{\blue{Terminal state reward}} & \checkmark & \checkmark & \checkmark & 0 \\
 \hline
 \textit{Target Radius} &  & \checkmark &  & 0.05 \\
 \hline
 \textit{Target Point} &  & \checkmark &  & Origin \\
 \hline
 \textit{$N$-State Space Range} &  & \checkmark &  & [-10, 10] \\
 \hline
 \textit{\green{Action Range}} &  & \checkmark & \checkmark* & [-1, 1] \\
 \hline
 \textit{\green{Time Unit}} &  & \checkmark & \checkmark* & 1 \\
 \hline
 \textit{Action Loss Weight} &  & \checkmark &  & 0 \\
 \hline
 \textbf{Initial State Distribution} & \checkmark & \checkmark &  & Uniform over non-terminal states \\
 \hline
 \textit{Transition Dynamics Order} &  & \checkmark &  & 1 \\
 \hline
 \textit{Inertia} &  & \checkmark &  & 1 \\
 \hline
 \textit{Terminal regions} &  & \checkmark &  & Circle of radius \textit{target radius} around target point \\
 \hline
\end{tabular}
\end{center}
\end{table}

With respect to the new \textit{time unit}, it is achieved by multiplying the base \textit{Gym Mujoco} environments' \texttt{frame\_skip} and not \textit{Mujoco}'s internal timestep. This means that if the base environments' \texttt{frame\_skip} is 5, for example, the smallest new \textit{time unit} achievable using our wrapper would be $1/5$ times the default time unit of the base environment and that larger time units would be integral multiples of this smallest achievable time unit. We decided not to achieve the new \textit{time unit} by modifying \textit{Mujoco}'s internal timestep, although this would have technically been possible, because that would change the numerical integration done by \textit{Mujoco} and thus would introduce more noise into the dynamics of the environment. Further, while this wrapper also works with any \textit{Mujoco} environment in the way described above, we would advise the reader to be careful that this might interfere with the reward function of the base environment. For the environments \textit{HalfCheetahV3, Pusher} and \textit{Reacher}, since changing the \textit{time unit} to be 5 times smaller, for example, means that we get 5 rewards in the original amount of \textit{real} time per time step and these rewards are of the same scale, we multiply each of the 5 rewards by $0.2$ to achieve \textit{episodic} reward on the same scale as the original environment. For \textit{HalfCheetahV3}, for example, the rewards are of the same scale in the new time unit because its reward consists of 2 parts: the velocity and the control cost. So, for a given velocity and given applied control having rewards 5 times more frequently in the original \textit{real} time step would mean a reward scale that is 5 times greater and so the need for the correction. \textit{Pusher} and \textit{Reacher} have similar reasoning and, therefore, we would caution the user to be careful when setting the \textit{time unit} dimension of hardness in the complex \textit{Mujoco} wrapper\footnote{A similar concern does \textit{not} exist for the toy continuous environments because in that case, the reward is the amount of distance moved from the current position to the target point. So even if the time step is made smaller, the reward scale becomes proportionally smaller because the agent can only travel proportionally smaller distances towards the goal on average in smaller timesteps.}.
}


\section{Design Decisions for Toy Environments}\label{sec:design_decisions}
We now discuss some of the design decisions underlying the toy environments to shed more light on their design.

\subsection{Discrete Environment Generation} Once the values for the dimensions of hardness are set, for the case of auto-generated discrete toy environments, as mentioned above, $P$ is generated by selecting for each $N$-state $ns$ in independent set $i$, for each action $a$, a random successor state $ns'$ from independent set $i+1$. This results in a regular grid-world like structure for $P$. For $R$, we select the $num_r$ rewardable sequences randomly for a given reward density $rd$ based on all the sequences possible under the generated $P$.
A motivating reason for unit testing in such a regularly structured environment is that all the RL agents we are aware of do not themselves take the structure of the environment into account and are designed for general $P$s and $R$s. Because of this, once the toy environment's dimensions are set, the structure of the environment is set and the agents should show similar behaviour on all such environments and this is exactly what we observed in our experiments when run with different seeds for the environment generation.

\subsection{Regular Structure for Unit Tests} In principle, it is always possible to design \textit{adversarial} $P$s \shortcite{adversarial_P_game_trees_dana_1983,adversarial_P_ramanujan_aaai_2010} which can be made arbitrarily hard to solve. Suppose there is an environment where a large reward is placed in an unknown and deliberately unexpected location. \chngdjair{For many use cases, evaluating an agent on such an environment may give us a misleading measure of the type of agent intelligence we are hoping to measure.} This is, in some cases, also a problem with many complex environments, e.g., \textit{HalfCheetah} has a bug that allows the agent to reach infinite speed and obtain enormous rewards \shortcite{bo_mbrl_zhang-aistats21}. \textit{qbert} has a bug which allows the agent to achieve a very large number of points \shortcite{back_to_basics_canonical_es_chrabaszcz_ijcai_18}. \textit{breakout} has a scenario where, if an agent creates a hole through the bricks, it can achieve a very large number of points. Even though the latter \textit{can} be a sign of desired behaviour, it skews the distribution of rewards and introduces a \textit{lot of} variance in the evaluation. There is the additional danger that the blackbox nature of complex environments can lead researchers to draw inferences that may be biased by their intuition \shortcite{kirkeboen_intuition_bias_front_psych_2017}. For example, the agent strategy of creating a tunnel to target bricks in the top for \textit{breakout} has been challenged multiple times \shortcite{atrey_saliency_counterfactual_iclr_2020,tosch_toybox_arxiv_2019}. 
As \shortciteA{irpan_rl_doesnt_work_blogpost} sums it up: \textit{If my reinforcement learning code does no better than random, I have no idea if it’s a bug, if my hyperparameters are bad, or if I simply got unlucky.}
Thus, having a very complex $P$ or $R$ itself can introduce ``noise'' into the evaluation of agents and require many iterations of training before we can see the agent learning. For unit testing, especially, such complexity and variance in evaluation is likely undesirable. One needs quick insights on whitebox environments and thus, we believe it is beneficial to have a simple and regular structure.

Finally, a \textit{more} regular structure is imposed as opposed to the usual gridworld because, though, semantically meaningful, such gridworlds have small irregularities around edges which makes them hard to keep consistent with all the other dimensions and begins to introduce the kind of ``noise'' that was discussed for more complex environments above.

\subsection{Independently Controllable Dimensions} Algorithms like DQN \shortcite{dqn_mnih2013playing} have been applied to many varied environments and produce very variable performance across these. In some simple environments, DQN's performance exceeds human performance by large amounts, but in other environments, such as Montezuma's revenge, performance is very poor. For some of these environments, e.g. Montezuma's revenge, we need a very specific sequence of actions to get a reward. For others, there are different delays in rewards. A problem with evaluating on these environments is that we have either no control over their difficulty or little control such as with having different difficulty levels. But even these difficulty levels do not isolate the confounding factors that are present at the same time and do not allow us to control the confounding factors \textit{individually}. We make that possible with our dimensions. We do not try to capture, say credit assignment or generalisation as dimensions because these are not independently controllable. They are to be dealt with at a higher level, something that is like an \textit{integration} test, rather than a \textit{unit} test.



\section{Experimenting with MDP Playground}
\label{sec:experiments}

We now discuss in detail how to experiment on \textit{MDP Playground} to understand and debug agents.

\subsection{Experimental Setup}
We used the following experimental setup for this section. \textbf{Agents}: We evaluated \textit{Rllib} implementations \shortcite{ray_liang2017rllib} of DQN \shortcite{dqn_mnih2013playing}, Rainbow DQN \shortcite{hessel2017rainbow}, and A3C \shortcite{a3c_mnih2016asynchronous} on discrete environments and DDPG \shortcite{ddpg_lillicrap2015continuous}, TD3 \shortcite{td3_fujimoto2018} and SAC \shortcite{sac_haarnoja2018} on continuous environments over grids of values for the dimensions of hardness. \chngdjair{The information state used by the agent was always the observation given out by the environment.} Hyperparameters (including neural network architectures used) and the tuning procedure used are specified in Appendices \ref{sec:hpo_append} and \ref{sec:append_tuned_hyperparams}. We used fully connected networks except for image representations where we used Convolutional Neural Networks (CNNs).

\subsubsection{Environments}
For the \textbf{discrete toy} experiments, we created a simple toy task with $\|S\|$ and $\|A\|$ set to $8$. The \textit{diameter} was set to $1$. Thus, an agent could transition from an $N$-state to every other $N$-state. The reward density was such that when the sequence length was 1, only one sequence of a single $N$-state was rewarding, so that the task in the vanilla environment was as easy as choosing the right action to reach the rewarding $N$-state at every time step. For the \textbf{continuous toy} experiments, \chngdjair{we set the state and action space dimensionalities to $2$. The state space range for each dimension was $[-10, 10]$ while the default action space range was $[-1, 1]$. The episodes would terminate when an algorithm would reach the \textit{target point}, or after at most $100$ timesteps. The agent had to reach the same target point $[0, 0]$, from random starting positions.} \chngdjair{For these toy experiments, the dimension under consideration had a grid of values, each of which is shown in the figures for the corresponding experiment below. All other dimensions of hardness were set to their default values mentioned in Table \ref{tab:default_dimension_values}. We evaluated 100 seeds for each of these experiments. We ran the DQN variants for $20k$ timesteps and A3C for $150k$ timesteps. The reason for selecting a different number of timesteps was dependent on how long it took for the different agents to converge and is discussed further in Appendix \ref{sec:append_max_timesteps_discussion}. We ran all the continuous control task agents for $20k$ timesteps.}
For the \textbf{complex} experiments, we used \textit{Atari} and \textit{Mujoco}. For \textit{Atari}, we ran the agents on \textit{Beam Rider}, \textit{Breakout}, \textit{Qbert} and \textit{Space Invaders}. For \textit{Mujoco}, we ran the agents on \textit{HalfCheetahV3}, \textit{Pusher} and \textit{Reacher} using \textit{mujoco-py}. We evaluated $5$ seeds for $10M$ steps ($40M$ frames) for Atari, $500k$ steps for \textit{Pusher} and \textit{Reacher} and $3M$ for \textit{HalfCheetah}.

\chngdjair{\subsubsection{Analysis}
For the experiments with toy environments, in the presence of \textit{delay} and \textit{sequence length}, we normalise the plots so that the optimal episodic reward is 100. For delay, we do so by multiplying the episodic reward by $100 / (100 - delay)$. For sequence length, we do so by multiplying the episodic reward by the sequence length, $n$. \chngdjaircrc{In the bar plots, we plot the Area Under the Curve (AUC) of the training curves. As a result, the reward scales between the learning curves and bar plots would differ.} For all bar plots, we apply Bonferroni corrections \shortcite{colas_2018_seeds_rl} for a significance level of 0.05 (Bonferroni corrections adjust the significance level of 0.05 as 0.05 / (\textit{number of comparisons being performed}), e.g. $0.05 / \binom{5}{2}$ if there are 5 bars in a plot).}


\subsection{Understanding Existing Agents}\label{sec:understanding_agents}
We hope our platform can help analyse and understand RL agents much better, especially due to the ease of performing experiments with fine-grained control over the injected dimensions of hardness.
We first describe an experiment where experiments on toy environments alone led us to interesting insights. We then describe some experiments where we compare and contrast the performances on the toy and complex environments for further interesting insights.

\subsubsection{Image representations}
\begin{figure}[t]
        \centering
        \begin{subfigure}[b]{0.435\textwidth}
            \centering
            \includegraphics[width=\textwidth]{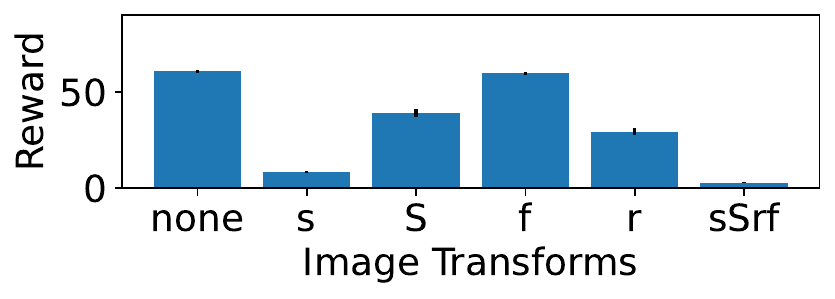}
            \caption[DQN]%
            {{\small DQN}}    
            \label{fig:dqn_repr_learn}
        \end{subfigure}
        \begin{subfigure}[b]{0.435\textwidth}   
            \centering 
            \includegraphics[width=\textwidth]{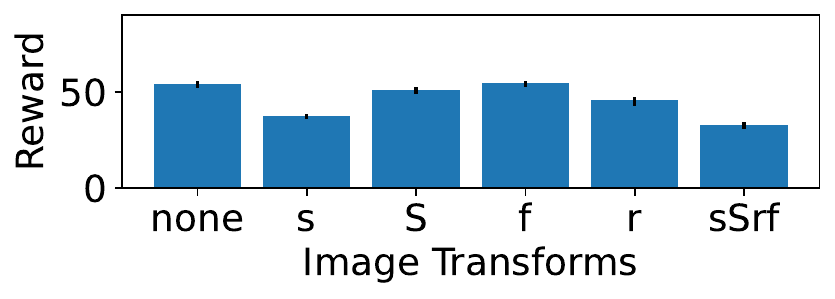}
            \caption[Rainbow]%
            {{\small Rainbow}}    
            \label{fig:rainbow_repr_learn}
        \end{subfigure}
        \begin{subfigure}[b]{0.435\textwidth}
            \centering
            \includegraphics[width=\textwidth]{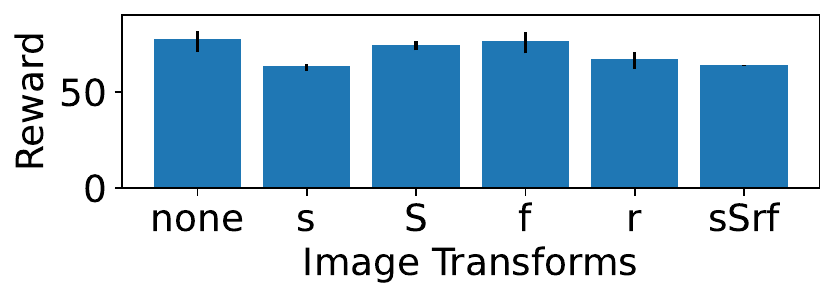}
            \caption[A3C]%
            {{\small A3C}}    
            \label{fig:a3c_repr_learn}
        \end{subfigure}
        \begin{subfigure}[b]{0.31\textwidth}
            \centering
            \includegraphics[width=\textwidth]{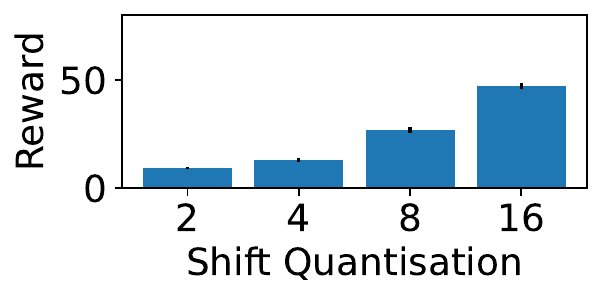}
            \caption[DQN]%
            {{\small DQN shift}}    
            \label{fig:dqn_sh_quant}
        \end{subfigure}

        \caption[ AUC of episodic reward at the end of training ]
        {AUC of episodic reward at the end of training for the different agents when varying \textbf{image representation transforms} on the discrete toy environments. `s' denotes \textit{shift}, `S' denotes \textit{scale}, `f' denotes \textit{flip}, `r' denotes \textit{rotate} \chngdjaircrc{and `sSrf' denotes all $4$ of these transforms in the X-axis ticks in the first three sub-figures and \textit{Shift Quantisation} in the fourth sub-figure} represents quantisation (in pixels) of the \textit{shift}s in the DQN experiment for this. Error bars represent bootstrapped confidence intervals with Bonferroni corrections for a significance level of 0.05.}
        \label{fig:repr_learn}
\end{figure}

As an example of what a researcher might want to experiment with, \textit{MDP Playground} allows turning on image representations for the categorical states in the discrete toy environments. It further allows applying transforms to these images to have multiple possible representations for the same state. We applied various transforms (\textit{shift}, \textit{scale}, \textit{rotate} and \textit{flip}) one at a time and also all at once to study the effects of these. We observed that the more transforms that are applied to the images at once, the harder it is for agents to learn, as can be seen in Figures \ref{fig:dqn_repr_learn}-\ref{fig:a3c_repr_learn}. This was to be expected since there are many more combinations to generalise over for the agent.

\begin{figure}[t]
        \centering
        \begin{subfigure}[]{0.435\textwidth}
            \centering
            \includegraphics[width=\textwidth]{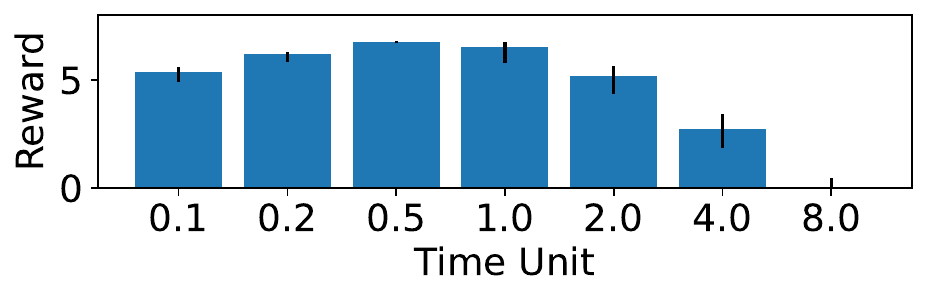} 
            \caption[DDPG]%
            {{\small toy}}    
            \label{fig:ddpg_time_unit_rew}
        \end{subfigure}
        \begin{subfigure}[]{0.435\textwidth}
            \centering
            \includegraphics[width=\textwidth]{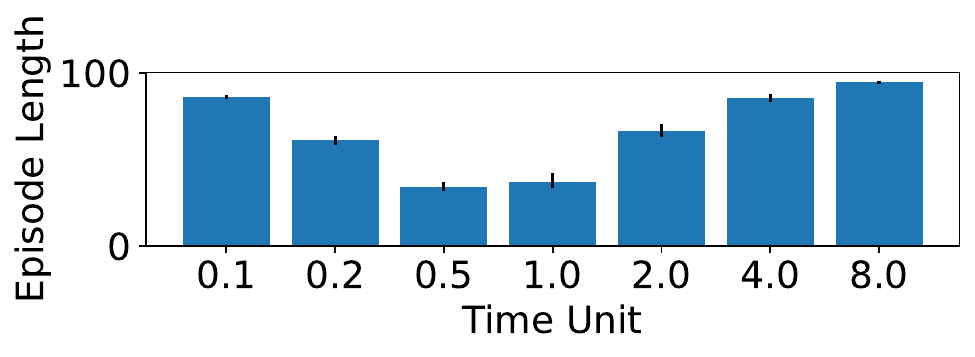}
            \caption[DDPG]%
            {{\small toy}}
            \label{fig:ddpg_time_unit_len}
        \end{subfigure}
        \begin{subfigure}[]{0.435\textwidth}   
            \centering 
            \includegraphics[width=\textwidth]{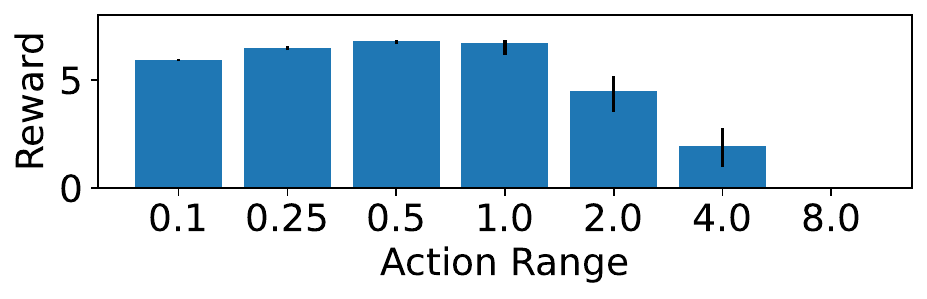} 
            \caption[DDPG]%
            {{\small toy}}
            \label{fig:ddpg_action_space_max_rew}
        \end{subfigure}
        \begin{subfigure}[]{0.435\textwidth}   
            \centering 
            \includegraphics[width=\textwidth]{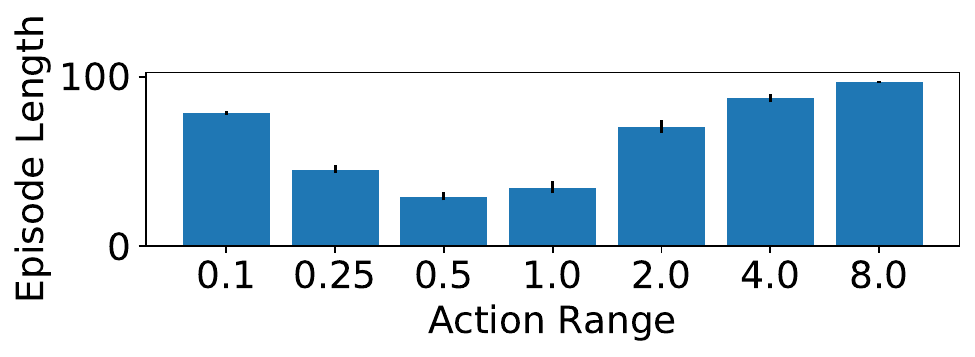}
            \caption[DDPG]%
            {{\small toy}}    
            \label{fig:ddpg_action_space_max_len}
        \end{subfigure}
        \begin{subfigure}[]{0.435\textwidth}   
            \centering 
            \includegraphics[width=\textwidth]{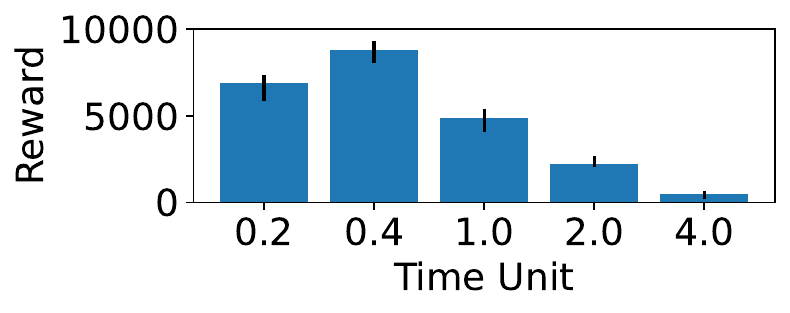} 
            \caption[DDPG]%
            {{\small HalfCheetah}}
            \label{fig:ddpg_time_unit_halfcheetah}
        \end{subfigure}
        \begin{subfigure}[]{0.435\textwidth}
            \centering
            \includegraphics[width=\textwidth]{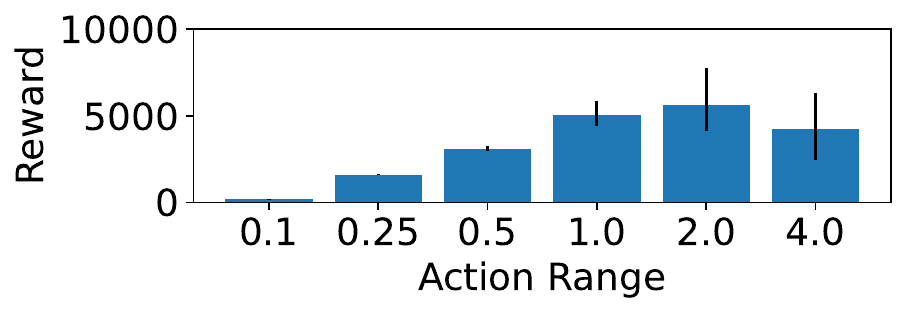} 
            \caption[DDPG]%
            {{\small HalfCheetah}}    
            \label{fig:ddpg_act_max_halfcheetah}
        \end{subfigure}\hspace{0.05\textwidth}%

        \caption[ AUC of episodic reward and lengths at the end of training ]
        {AUC of selected metric at the end of training for DDPG \textbf{(a)}: episodic reward with \textbf{time unit} on the toy environment \textbf{(b)}: episode length with \textbf{time unit} on the toy environment. \textbf{(c)}: episodic reward with \textbf{action range} on the toy environment \textbf{(d)}: episode length with \textbf{action range} on the toy environment. \textbf{(e)}: episodic reward with \textbf{time unit} on a complex (HalfCheetah) environment (\textit{time unit} values are relative to the defaults). \textbf{(f)}: episodic reward with \textbf{action range} on a complex (HalfCheetah) environment (\textit{action range} values are relative to the defaults). Error bars represent bootstrapped confidence intervals with Bonferroni corrections for a significance level of 0.05.}
        \label{fig:ddpg_continuous}
\end{figure}
        
\begin{figure}[t]
        \centering
        \begin{subfigure}[]{0.435\textwidth}
            \centering
            \includegraphics[width=\textwidth]{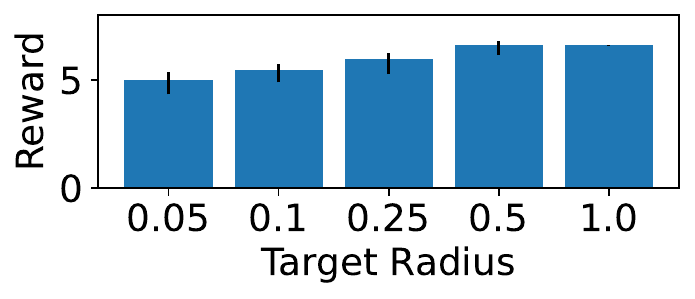}
            \caption[DDPG]%
            {{\footnotesize Episodic reward}}    
            \label{fig:ddpg_target_radius_rew}
        \end{subfigure}\hspace{0.05\textwidth}%
        \begin{subfigure}[]{0.435\textwidth}
            \centering
            \includegraphics[width=\textwidth]{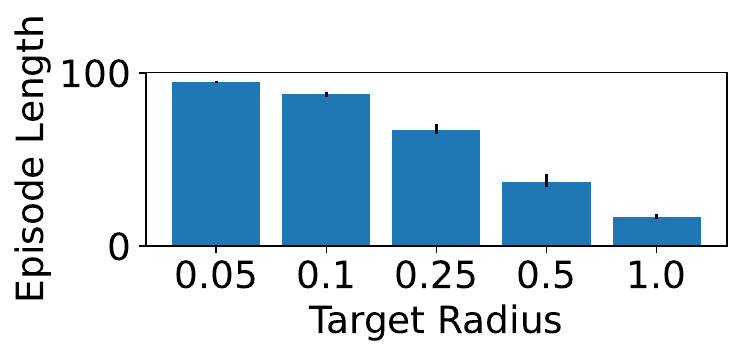}
            \caption[DDPG]%
            {{\footnotesize Episode length}}    
            \label{fig:ddpg_target_radius_len}
        \end{subfigure}
        
        \caption[ AUC of episodic reward and lengths at the end of training ]
        {AUC of selected metric at the end of training for DDPG when varying \textbf{target radius} on the toy environment \textbf{(a)}: episodic reward \textbf{(b)}: episodic length. Error bars represent bootstrapped confidence intervals with Bonferroni corrections for a significance level of 0.05.}
        \label{fig:ddpg_target_radius}
\end{figure}

However, it was unexpected to us that the most difficult transform for the agents to deal with was \textit{shift}. Despite the spatial invariance learned in CNNs \shortcite{cnn-spatial-invariance1}, our results imply that this variation seems to be the hardest one to adapt to. The results seemed to suggest that the more the total different values that are possible for a transform, the harder it is to adapt to. \textit{Shift} had the most possible different values that could be applied to the images as compared to the other transforms. This fact seems to suggest that at least some amount of memorisation is taking place in the NNs.

As these trends were strongest in DQN, we evaluated further ranges for the individual transforms for DQN. Therefore, we quantised the \textit{shift}s to have fewer possible values.
Figure \ref{fig:dqn_sh_quant} shows that DQN's performance improved with increasing quantisation (i.e., fewer possible values) of \textit{shift}.
We noticed similar trends for the other transforms as well, although not as strong as they do not have as many different values as \textit{shift} (see Figures \ref{fig:dqn_scale_range}-\ref{fig:dqn_ro_quant} in Appendix \ref{sec:append_addnl_reward_plots}).
We emphasise that in a more complex setting, we would have easily attributed some of these results to chance but in the setting where we had individual control over the dimensions of hardness, our platform allowed us to dig deeper in a controlled manner.

\subsubsection{Time unit and Action Range}
As another example of a dimension of hardness a researcher might be interested in controlling, \textit{MDP Playground} allows varying the \textit{time unit} of how often the agent acts in a continuous environment. We varied this dimension of hardness in the toy continuous environment and observed that the \textit{time unit} has an optimal value which has a significant impact on performance (Figure \ref{fig:ddpg_time_unit_rew} for DDPG; and Figures \ref{fig:sac_time_unit_rew} and \ref{fig:td3_time_unit_rew} in the appendix for SAC and TD3). The \textit{time unit} can be neither too small nor too large since there is an optimal \textit{time unit} for which we should repeat the same action \shortcite{biedenkapp-icml21}. Based on this finding, we also \chngdjaircrc{analysed} the \textit{time unit} for complex environments (Figure \ref{fig:ddpg_time_unit_halfcheetah}, Figures \ref{fig:append_ddpg_time_unit_halfcheetah}-\ref{fig:sac_time_unit_halfcheetah}, \ref{fig:ddpg_time_unit_pusher}-\ref{fig:sac_time_unit_pusher} and \ref{fig:ddpg_time_unit_reacher}-\ref{fig:sac_time_unit_reacher} in the appendix), finding similar results on all environments. In fact, for all three agents, (Figures \ref{fig:ddpg_time_unit_halfcheetah}, Figures \ref{fig:sac_time_unit_halfcheetah} and \ref{fig:td3_time_unit_halfcheetah} in the appendix), we see peak performance on \textit{HalfCheetah} for a \textit{time unit} $0.4$ smaller than that of the \textit{expert}-designed vanilla environment. There were gains of even $100\%$ in performance in some cases. \chngdjair{We emphasise again that we do not advocate modifying the problem definition of such benchmark problems to claim new state of the art results, but rather for analysing problems.} A further insight to be had is that for simpler environments like the toy, \textit{Pusher} and \textit{Reacher}, the effect of the selection of the \textit{time unit} was not as pronounced as for a more complex environment like \textit{HalfCheetah}. This makes intuitive sense as one can expect a narrower range of values to work for more complex environments. \chngdjair{This shows that it may be even more important to study the effect of such dimensions of hardness for more complex environments, especially when defining the problem for a completely new task.}



\chngd{
We observed similar trends for \textit{action range} as for \textit{time unit}, in that there was an optimal value of \textit{action range}, i.e., that it can be neither too small nor too large. For the toy environment, please see Figure \ref{fig:ddpg_action_space_max_rew} for DDPG and Figures \ref{fig:sac_action_space_max_rew} and \ref{fig:td3_action_space_max_rew} in the appendix for SAC and TD3. Similarly as for \textit{time unit}, more pronounced trends for \textit{action range} can be seen for HalfCheetah in Figure \ref{fig:ddpg_act_max_halfcheetah} for DDPG and in Figures \ref{fig:td3_act_max_halfcheetah}-\ref{fig:sac_act_max_halfcheetah} for TD3 and SAC in the appendix. Figures \ref{fig:ddpg_act_max_pusher}-\ref{fig:sac_act_max_pusher} and \ref{fig:ddpg_act_max_reacher}-\ref{fig:sac_act_max_reacher} in the appendix show trends largely similar to the toy environment for all three agents on \textit{Pusher} and \textit{Reacher}.
%
This supports the insight gained on our toy environment that setting this value differently may lead to significant gains for an agent. 
Similar to the \textit{time unit}, in the expert-designed environment setting of \textit{Reacher} in \textit{Gym}, we found that all three algorithms performed best for an \textit{action range} $0.25$ times the value found in \textit{Gym} (Figures \ref{fig:ddpg_act_max_reacher}, \ref{fig:td3_act_max_reacher}, \ref{fig:sac_act_max_reacher} in the appendix).
The difference in performances across the different values of \textit{action range} is much greater in the complex environments. We believe this is due to correlations within the multiple degrees of freedom as opposed to a rigid object in the toy environment. 
}

To the best of our knowledge, the impact of \textit{time unit} and \textit{action range} while developing agents and defining problems is under-researched because the standard benchmark environments have been pre-configured by experts. \chngdjair{From the above discussion, it is clear that playing around with pre-configured values can lead to new insights even in \textit{known} environments and that we need to pay attention to these dimensions of hardness in a completely \textit{unknown} environment when defining the MDP formulation of a problem, if we want agents to perform as desired. And even when the problem definition is set, one might still want their agent to search within a subset of the action range or want their agent to repeat actions over multiple timesteps \shortcite{biedenkapp-icml21}.}

\subsubsection{Target Radius}
The \textit{target radius} is a value which is generally set to a small enough value to be able to say that the algorithm has reached the target. We noticed in the continuous toy environment that, for small values of the target radius, all the continuous control agents oscillated around the target while trying to reach it exactly. This can be observed in Figure \ref{fig:ddpg_target_radius_rew} and \ref{fig:ddpg_target_radius_len} for DDPG (and in Figures \ref{fig:td3_target_radius_rew} and \ref{fig:td3_target_radius_len} for TD3 and Figures \ref{fig:sac_target_radius_rew} and \ref{fig:sac_target_radius_len} for SAC in the appendix), where we note that even though the task \textit{was} learnt for different \textit{target radii}, the episode lengths were much longer for shorter target radii. This was because the agents kept oscillating outside the target radius. Even for such a simple task all evaluated algorithms failed to adapt to performing fine-grained control near the target. We believe that the agents did not learn to perform fine-grained control close to the target because the agents need to see data points closer to the goal to train their NNs, which highlights how data intensive model-free algorithms can be.

        

\begin{figure}[t]
        \centering
        \begin{subfigure}[b]{0.325\textwidth}
            \centering
            \includegraphics[width=\textwidth]{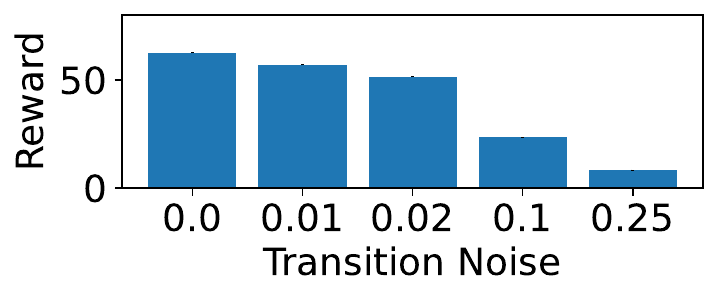}
            \caption[DQN]%
            {{\small toy train}}    
            \label{fig:dqn_p_noise_1d}
        \end{subfigure}
        \begin{subfigure}[b]{0.325\textwidth}
            \centering
            \includegraphics[width=\textwidth]{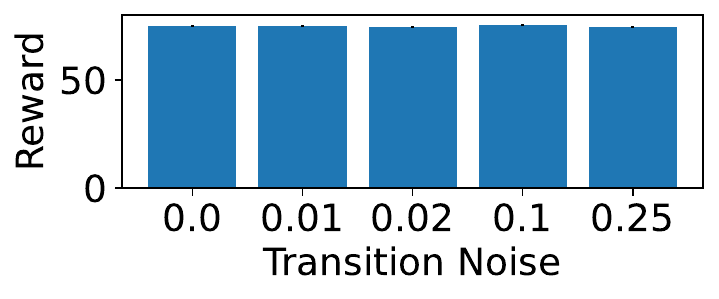}
            \caption[DQN]%
            {{\small toy noise-free}}    
            \label{fig:dqn_p_noise_eval_1d}
        \end{subfigure}
        \begin{subfigure}[b]{0.325\textwidth}
            \centering
            \includegraphics[width=\textwidth]{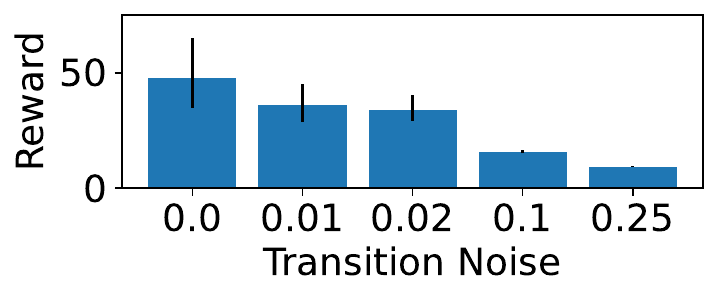}
            \caption[]%
            {{\small Breakout}}
            \label{fig:dqn_breakout_p_noise}
        \end{subfigure}
        
        \caption[ AUC of episodic reward and lengths at the end of training ]
        {AUC of episodic reward for DQN at the end of training when varying \textbf{transition noise} \textbf{(a)}: on the toy environment while training. \textbf{(b)}: on the toy environment when rolling out the policy that was learned on the noisy environment on a noise-free setting \textbf{(c)} on \textit{Breakout}. Error bars represent bootstrapped confidence intervals with Bonferroni corrections for a significance level of 0.05.}
        \label{fig:dqn_p_noise}
\end{figure}

\begin{figure}[t]
        \centering
        \begin{subfigure}[b]{0.325\textwidth}
            \centering
            \includegraphics[width=\textwidth]{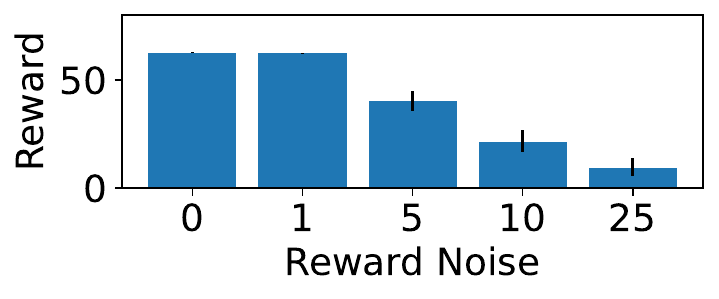}
            \caption[DQN]%
            {{\small toy train}}    
            \label{fig:dqn_r_noise_1d}
        \end{subfigure}
        \begin{subfigure}[b]{0.325\textwidth}
            \centering
            \includegraphics[width=\textwidth]{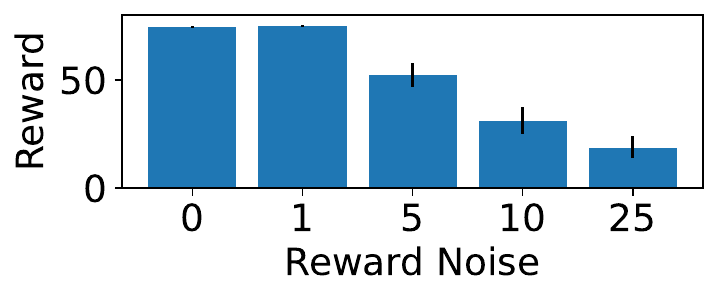}
            \caption[DQN]%
            {{\small toy noise-free}}    
            \label{fig:dqn_r_noise_eval_1d}
        \end{subfigure}
        \begin{subfigure}[b]{0.325\textwidth}
            \centering
            \includegraphics[width=\textwidth]{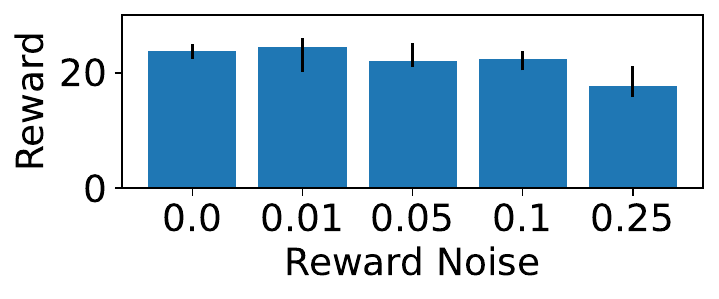}
            \caption[]%
            {{\small Breakout}}
            \label{fig:dqn_breakout_r_noise}
        \end{subfigure}

        \caption[ AUC of episodic reward and lengths at the end of training ]
        {AUC of episodic reward for DQN at the end of training when varying \textbf{reward noise} \textbf{(a)}: on the toy environment while training. \textbf{(b)}: on the toy environment when rolling out the policy that was learned on the noisy environment on a noise-free setting \textbf{(c)}: on \textit{Breakout}. Please note that this run on \textit{Breakout} had reward clamping turned off, which is on by default in Ray and thus had a very different episodic reward range compared to runs with other dimensions of hardness on \textit{Breakout}. Error bars represent bootstrapped confidence intervals with Bonferroni corrections for a significance level of 0.05.}
        \label{fig:dqn_r_noise}
\end{figure}

\subsubsection{Transition and Reward Noise} 
Experimenting with \textit{transition noise} and \textit{reward noise} allowed us to observe that performance dropped on the toy environments for all agents as one might expect (Figure \ref{fig:dqn_p_noise_1d} and \ref{fig:dqn_r_noise_1d}; Figures \ref{fig:algs_p_noise_train_1d}-\ref{fig:algs_r_noise_train_1d} in the appendix). Performance degrades gradually as more and more noise is injected. It is interesting that, during training, all the algorithms seem to be more sensitive to noise in the transition dynamics compared to the reward dynamics: transition noise values as low as 0.02 lead to a clear handicap in learning while for the reward dynamics (with the \textit{reward scale} being $1.0$) reward noise standard deviation of $\sigma_{r\_n} = 1$ still resulted in learning progress.

\chngdjaircrc{Interestingly, when we look at the performances on the noisy environment versus when rolling out the policy that was learned on the noisy environment on a noise-free setting for \textit{transition} noise in Figures \ref{fig:dqn_p_noise_1d} and \ref{fig:dqn_p_noise_eval_1d} (Figures \ref{fig:algs_p_noise_train_1d} and \ref{fig:algs_p_noise_eval_1d} for all agents in the appendix), we see that the training performance of the algorithms is more sensitive to noise during training than the eventual performance of the same policy on the noise-free setting. It is a non-trivial insight that injecting noise into the \textit{transition} function still leads to good learning for the noise-free version of the transition noise environment (as displayed in the noise-free rollout plots). A second interesting insight is when comparing the above pairs of figures from the transition noise environments to the corresponding ones for the \textit{reward} noise versions of the environments
Figures \ref{fig:dqn_r_noise_1d} and \ref{fig:dqn_r_noise_eval_1d} (Figures \ref{fig:algs_r_noise_train_1d} and \ref{fig:algs_r_noise_eval_1d} for all agents in the appendix). We observe that injecting noise into the \textit{reward} function does \textit{not} lead to good learning for the noise-free version of the reward noise environment. This is in contrast to the insight from the transition noise versions of the environments. We conjecture that this could be due to the noise from different noisy transitions in the replay buffer/training data cancelling each other out in the gradient-based updates while learning for the transition noise but not for the reward noise. It could also be that transition noise helps the agent \textit{explore} more while reward noise does not.}


We observed also for the Atari environments for all three agents that performance dropped on average when injecting \textit{transition noise} (Figure \ref{fig:dqn_breakout_p_noise} for DQN on \textit{Breakout}; Figures \ref{fig:dqn_beam_rider_p_noise}-\ref{fig:a3c_beam_rider_p_noise}, \ref{fig:append_dqn_breakout_p_noise}-\ref{fig:a3c_breakout_p_noise}, \ref{fig:dqn_qbert_p_noise}-\ref{fig:a3c_qbert_p_noise} and \ref{fig:dqn_space_invaders_p_noise}-\ref{fig:a3c_space_invaders_p_noise} for all agents and environments in the appendix). However, we also observe that performance drop is different for different environments. This is to be expected as there are other dimensions of hardness that we cannot control or measure for the complex environments. We further observed that for some of the environments, transition noise actually even helped performance (e.g. Figure \ref{fig:rainbow_space_invaders_p_noise} for Rainbow on \textit{Space Invaders}). This has also been observed in prior work \shortcite{bench_mbrl_wang2019benchmarking}. This can happen when the exploration policy was not tuned optimally since inserting transition noise is almost equivalent to $\epsilon$-greedy exploration for low values of noise. Finally, we observed similar trends for \textit{reward noise} (Figure \ref{fig:dqn_breakout_r_noise} for DQN on \textit{Breakout}; Figures \ref{fig:dqn_beam_rider_r_noise}-\ref{fig:a3c_beam_rider_r_noise},  \ref{fig:append_dqn_breakout_r_noise}-\ref{fig:a3c_breakout_r_noise},  \ref{fig:dqn_qbert_r_noise}-\ref{fig:a3c_qbert_r_noise} and \ref{fig:dqn_space_invaders_r_noise}-\ref{fig:a3c_space_invaders_r_noise} for all agents and environments in the appendix). 

\chngdjair{Another interesting insight is that if one compares the X-axis ticks for the toy and complex environments for \textit{reward noise} (e.g. Figures \ref{fig:dqn_r_noise_1d} and \ref{fig:dqn_breakout_r_noise}), the scale of the \textit{reward noise} is very different in the toy and complex environments. This, we believe, is because the complex environments are much more sparse \chngdjaircrc{- if one rolls out a random agent, or even a trained one, on the complex environments, the non-zero rewards are far less frequent than in the toy environments we experimented with. While it is naturally possible to design toy environments to be much more representative in terms of trying to match the hardness dimensions of complex environments, it tends to happen that this leads to them becoming much harder to learn on as well and, as a result, losing their ability to be quick testbeds}. Training the agents on them with the same scale of \textit{reward noise} as the toy environments led to no learning at all, so we had to select far smaller values to show the trends. Another important detail to note about the reward noise experiments on Atari environments is that Atari environments in Ray have what they term \textit{reward clipping} turned on by default. This sets the reward at every time step to 0, -1 or +1 depending on the sign of the reward obtained\footnote{It is called reward \textit{clipping} in there, but \textit{clamping} might be a better term}. This interferes with the reward noise experiments because even very small noisy rewards are clamped according to their sign, e.g., $+0.01$ would be set to $+1$ and then the meaning of reward noise is different to what we intended to inject. As such, we had to change reward clipping to instead actually have ``clip'' behaviour, i.e. to clip rewards greater than $+1$ to $+1$ and clip rewards less than $-1$ to $-1$, to perform meaningful experiments with reward noise. This led to very different episodic reward ranges compared to runs with other dimensions of hardness, e.g. on \textit{Breakout} (compare reward scales in Figures \ref{fig:dqn_breakout_p_noise} and \ref{fig:dqn_breakout_r_noise}). This difference seemed to occur only for DQN and Rainbow but not for A3C (please compare the reward scales in the bottom row of Figure \ref{fig:breakout_perfs} in the appendix with the reward scales in its upper rows) which is an interesting avenue for future research regarding behaviour of these agents in the presence or absence of reward clipping. Ideally, needless to say, we want our agents to be robust to different scales of reward and to not have to resort to something like reward clipping.}


\begin{figure}[t]
        \centering
        \begin{subfigure}[]{0.345\textwidth}
            \centering
            \includegraphics[width=\textwidth]{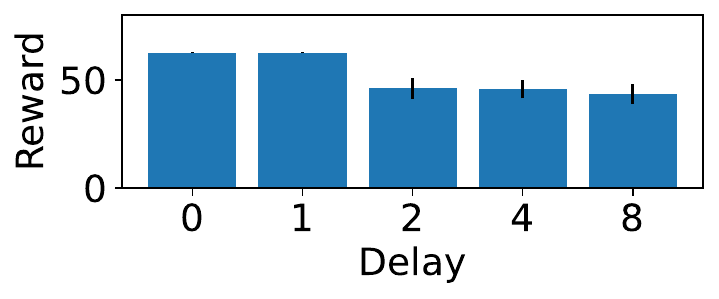}
            \caption[DQN]%
            {{\small toy}}    
            \label{fig:DQN_del_1d}
        \end{subfigure}
        \begin{subfigure}[]{0.345\textwidth}
            \centering
            \includegraphics[width=\textwidth]{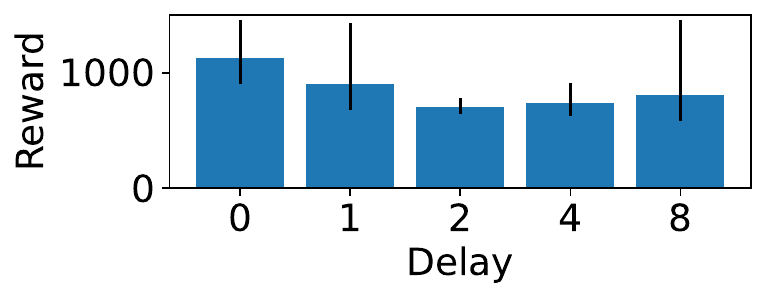} 
            \caption[]%
            {{\small Qbert}}
            \label{fig:dqn_qbert_del}
        \end{subfigure}
        \begin{subfigure}[]{0.295\textwidth}
            \centering
            \includegraphics[width=\textwidth]{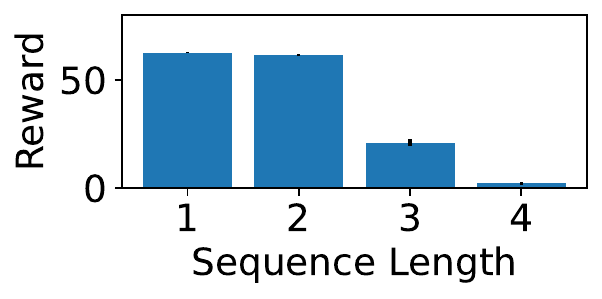}
            \caption[DQN]%
            {{\small Sequence Length}}    
            \label{fig:DQN_seq_1d a}
        \end{subfigure}
        
        \caption[ AUC of episodic reward and lengths at the end of training ]
        {AUC of episodic reward for DQN at the end of training when varying  \textbf{(a)}: \textbf{delay} on the toy environment. \textbf{(b)}: \textbf{delay} on \textit{Qbert}. \textbf{(c)}: \textbf{sequence length} on the toy environment. Error bars represent bootstrapped confidence intervals with Bonferroni corrections for a significance level of 0.05.}
        \label{fig:dqn_del_seq}
\end{figure}

\begin{figure}[t]
        \centering
            \includegraphics[width=0.34\textwidth]{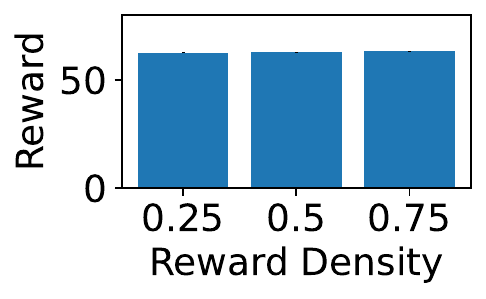}
        
        \caption[ AUC of episodic reward and lengths at the end of training ]
        {AUC of episodic reward for DQN at the end of training when varying \textbf{(a)}: \textbf{reward density} on the toy environment. Error bars represent bootstrapped confidence intervals with Bonferroni corrections for a significance level of 0.05.}
        \label{fig:dqn_density}
\end{figure}

\subsubsection{Reward Delay and Sequence Length}
Experimenting with \textit{delay} also led to similar high-level trends as for \textit{transition noise}: performance dropped on the toy environments for all agents (Figure \ref{fig:DQN_del_1d} for DQN; Figure \ref{fig:algs_del_1d} for all agents in the appendix). We additionally also plot the \textit{learning curves}, when varying delay, in Figure \ref{fig:dqn_del_1d_train_curves}. We see how training proceeds much more smoothly and is less variant across different seeds for the vanilla environment (where the delay is $0$) and that the variance across seeds is very large for delay $=2$. Such noisy learning curves are typically associated with RL and clearly partial observability, as is the case here, is a contributing factor even for a toy environment. Sometimes, the agents still managed to perform well, which emphasises that agents \textit{can} perform well even when having a non-Markov information state and shows how \textit{tuning} seeds can lead to good results \shortcite<see e.g., Figure~5 in>{deep_rl_that_matters}.

\begin{sidewaysfigure*}
        \centering
        \begin{subfigure}[]{\textwidth}
            \centering
            \includegraphics[width=\textwidth]{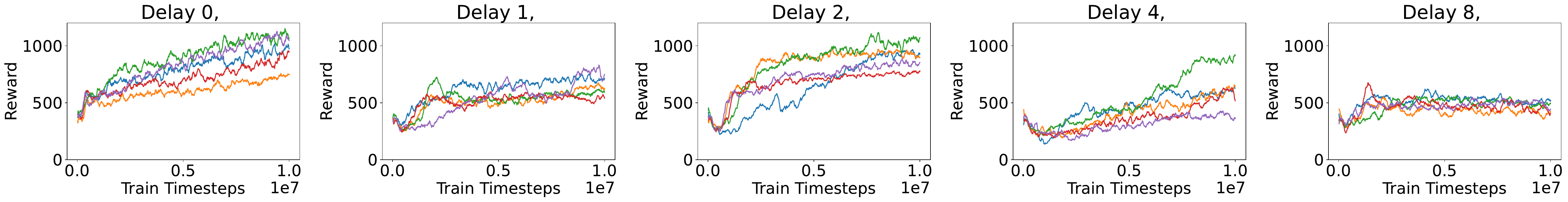}
            \caption[]%
            {{\small DQN on Beam Rider}}
            \label{fig:dqn_beam_rider_del_train_curves}
        \end{subfigure}
        \begin{subfigure}[]{\textwidth}
            \centering
            \includegraphics[width=\textwidth]{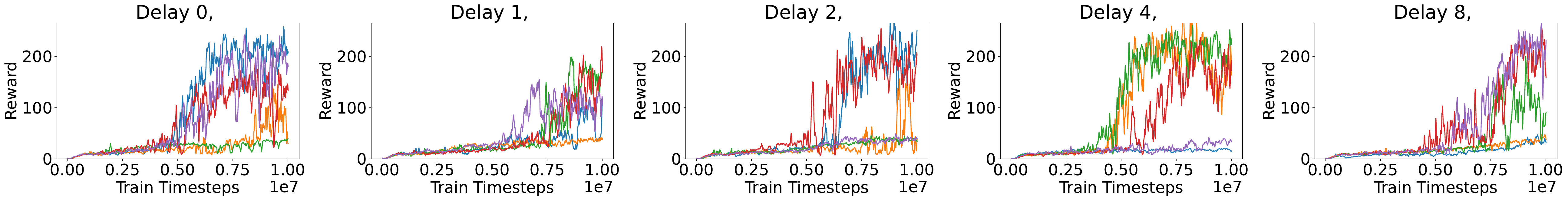}
            \caption[]%
            {{\small DQN on Breakout}}
            \label{fig:dqn_breakout_del_train_curves}
        \end{subfigure}
        \begin{subfigure}[]{\textwidth}
            \centering
            \includegraphics[width=\textwidth]{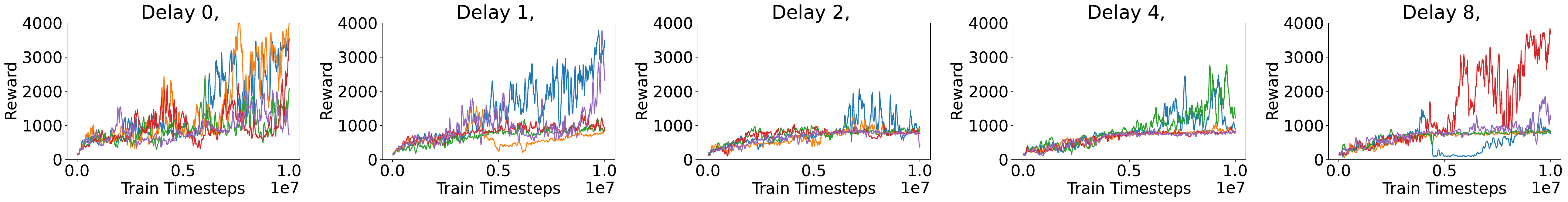}
            \caption[]%
            {{\small DQN on Qbert}}
            \label{fig:dqn_qbert_del_train_curves}
        \end{subfigure}
        \begin{subfigure}[]{\textwidth}
            \centering
            \includegraphics[width=\textwidth]{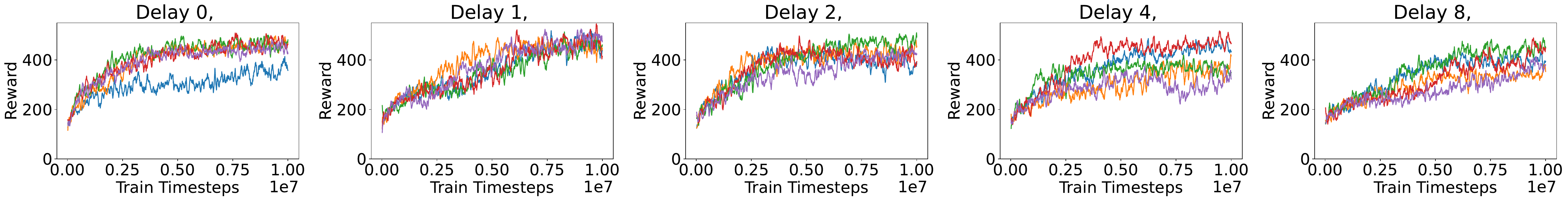}
            \caption[]%
            {{\small DQN on Space Invaders}}
            \label{fig:dqn_space_invaders_del_train_curves}
        \end{subfigure}
        
        \caption[]
        {Training Learning Curves for DQN on Atari environments with \textit{delay}. Please note that each different colour corresponds to one of $5$ seeds in each subplot.}
        \label{fig:dqn_train_curves_delay}
\end{sidewaysfigure*}

\begin{sidewaysfigure*}
        \centering
        \begin{subfigure}[]{\textwidth}
            \centering
            \includegraphics[width=\textwidth]{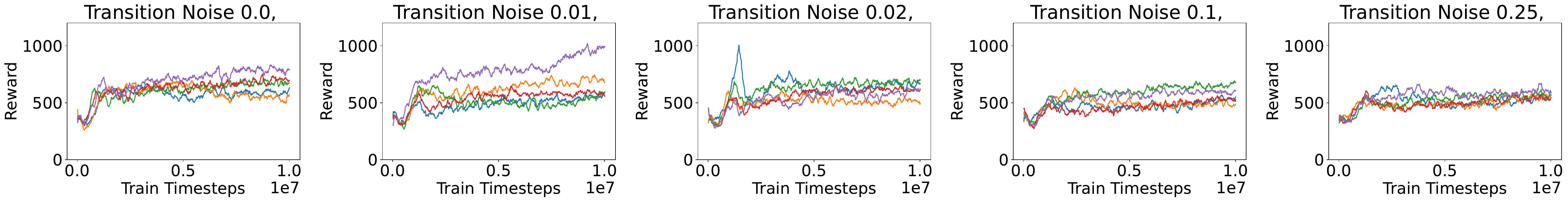}
            \caption[]%
            {{\small DQN on Beam Rider}}
            \label{fig:dqn_beam_rider_p_noise_train_curves}
        \end{subfigure}
        \begin{subfigure}[]{\textwidth}
            \centering
            \includegraphics[width=\textwidth]{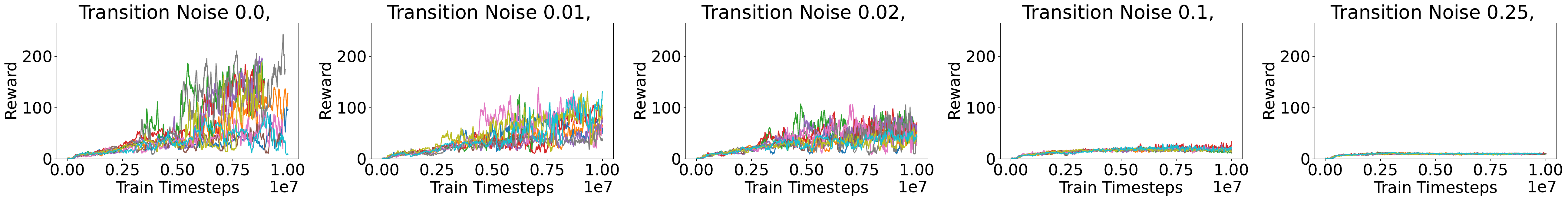}
            \caption[]%
            {{\small DQN on Breakout}}
            \label{fig:dqn_breakout_p_noise_train_curves}
        \end{subfigure}
        \begin{subfigure}[]{\textwidth}
            \centering
            \includegraphics[width=\textwidth]{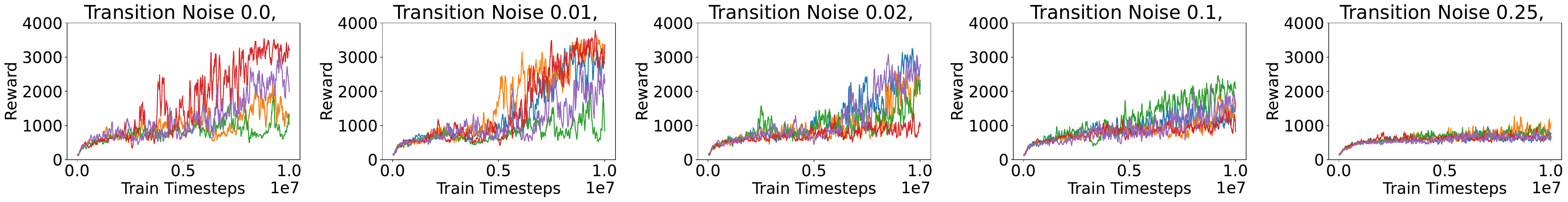}
            \caption[]%
            {{\small DQN on Qbert}}
            \label{fig:dqn_qbert_p_noise_train_curves}
        \end{subfigure}
        \begin{subfigure}[]{\textwidth}
            \centering
            \includegraphics[width=\textwidth]{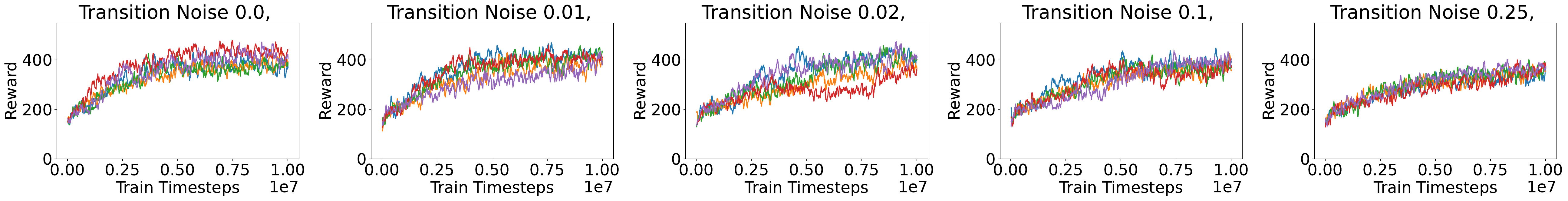}
            \caption[]%
            {{\small DQN on Space Invaders}}
            \label{fig:dqn_space_invaders_p_noise_train_curves}
        \end{subfigure}
        
        
        \caption[]
        {Training Learning Curves for DQN on Atari environments with \textit{transition noise}. Please note that each different colour corresponds to one of $5$ seeds in each subplot.}
        \label{fig:dqn_train_curves_p_noise}
\end{sidewaysfigure*}

\begin{figure*}[ht]
        \centering
            \centering
            \includegraphics[width=\textwidth]{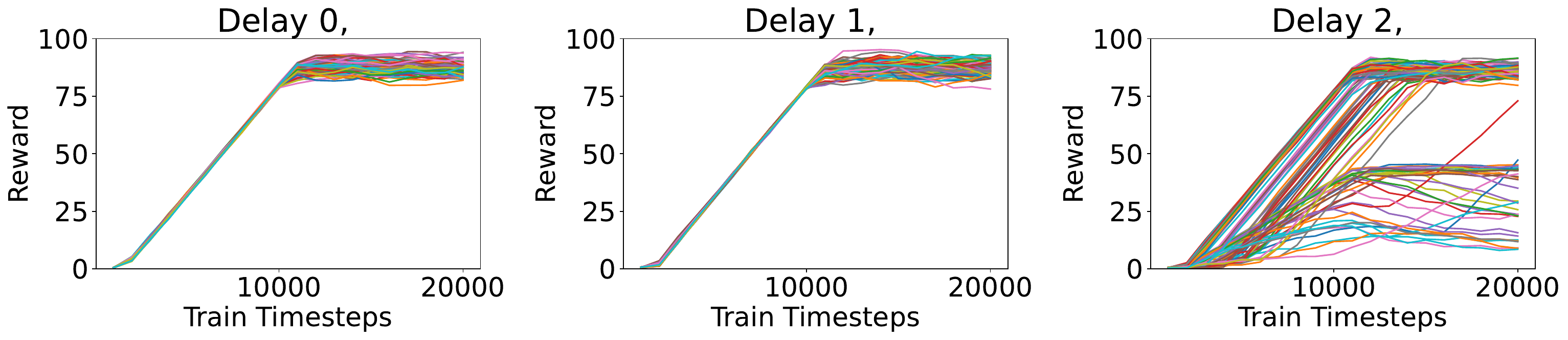} 
            \caption[DQN]%
            {{\small Train Learning Curves for $100$ runs with different seeds for DQN when varying \textbf{delay}. Please note that each different colour corresponds to one of $100$ seeds in each subplot. We do not show the curves for delay = 4 and delay = 8 which follow the same trends to improve readability.}}
            \label{fig:dqn_del_1d_train_curves}
\end{figure*}

We also note that, on average, performance dropped for the Atari environments (Figure \ref{fig:dqn_qbert_del} for DQN on \textit{Qbert}; Figures \ref{fig:dqn_beam_rider_del}-\ref{fig:a3c_beam_rider_del}, \ref{fig:dqn_breakout_del}-\ref{fig:a3c_breakout_del}, \ref{fig:dqn_qbert_del append_fig}-\ref{fig:a3c_qbert_del} and \ref{fig:dqn_space_invaders_del}-\ref{fig:a3c_space_invaders_del} for all agents and environments in the appendix).
However, the trends for \textit{delay} were noisier than for \textit{transition noise}. We believe this is because the situation for delays is more nuanced in the case of complex environments: many considered environments tend to have repetitive sequences which would make the effect of injecting delays noisier. Interestingly, many of the learning curves with delays inserted, are indistinguishable from normal learning curves (e.g., DQN on Atari in Figure \ref{fig:dqn_train_curves_delay}). We believe that, in addition to the motivating examples from Section \ref{sec:motivations_dimensions}, this is empirical evidence that delays are already present in these environments and so inserting them does not cause the curves to look vastly different. In contrast, when we look at learning curves for transition noise (e.g., DQN on Atari in Figure \ref{fig:dqn_train_curves_p_noise}), we observe that, as we inject more and more noise, training tends to a smoother curve as the agent tends towards becoming a completely random agent. Pointers to more such learning curves can be found in Appendix \ref{sec:append_complex_envs_learn_curves}.

Even more nuanced in the case of \textit{delay} is the comparison of effects on the different environments: the greatest drops in performance, on average, were on \textit{Qbert}, followed by \textit{Beam Rider}, \textit{Space Invaders} and \textit{Breakout} (see again Figures \ref{fig:dqn_beam_rider_del}-\ref{fig:a3c_beam_rider_del}, \ref{fig:dqn_breakout_del}-\ref{fig:a3c_breakout_del}, \ref{fig:dqn_qbert_del append_fig}-\ref{fig:a3c_qbert_del} and \ref{fig:dqn_space_invaders_del}-\ref{fig:a3c_space_invaders_del} in the appendix); for \textit{Breakout}, in many instances, we do not even see any performance drops. We believe this is because large delays from rewarding trajectory to reward are already present in \textit{Breakout}, which means that inserting more delays does not have as large an effect as in \textit{Qbert}. Agents are most strongly affected in \textit{Qbert} which, upon looking at gameplay, we believe has the least delays from rewarding trajectory to reward. It is interesting that inserting \textit{delays} for Rainbow on \textit{Qbert} seems to really impact its performance, more so than any of the other Atari experiments. 

To further analyse and compare the effects of the dimensions of hardness on the toy and complex environments for the Atari experiments, we use the Spearman rank correlation coefficient between corresponding toy and complex experiments for performance across different values of the dimension of hardness (Tables \ref{tab:delay_rank_corr}-\ref{tab:r_noise_rank_corr} in the appendix list the individual rank correlation for each experiment (for different agents, environments and dimensions of hardness). The Spearman correlation was $>=0.7$ for $27$ out of $36$ experiments and a positive correlation for eight of the remaining nine. DQN with delays added on breakout was the only experiment with correlation $0$, which further supports our hypothesis above that delays are already present in the greatest amount in breakout. Further, the average rank correlation over 12 experiments (3 agents x 4 Atari environments) was $0.867$ for \textit{transition noise}, $0.733$ for \textit{reward noise} and $0.617$ for \textit{reward delay} which supports the observation above that trends were noisier for delay.

Results for experiments with different \textit{sequence length}s on the toy environment are qualitatively similar to the ones for delay. However, we observe (see Figure \ref{fig:DQN_seq_1d a} for DQN, Figure \ref{fig:algs_seq_len_1d} for all agents in the appendix) that sequence length has a more drastic effect in terms of degradation of performance. 

\subsubsection{Irrelevant Features}
\begin{figure}[ht]
        \centering

        \begin{subfigure}[]{0.30\textwidth}   
            \centering 
            \includegraphics[width=\textwidth]{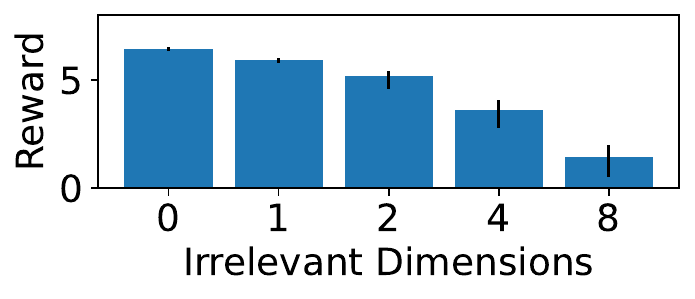}
            \caption[SAC]%
            {{\small SAC toy}}
            \label{fig:sac_irr_dims_rew}
        \end{subfigure}
        \begin{subfigure}[]{0.30\textwidth}   
            \centering 
            \includegraphics[width=\textwidth]{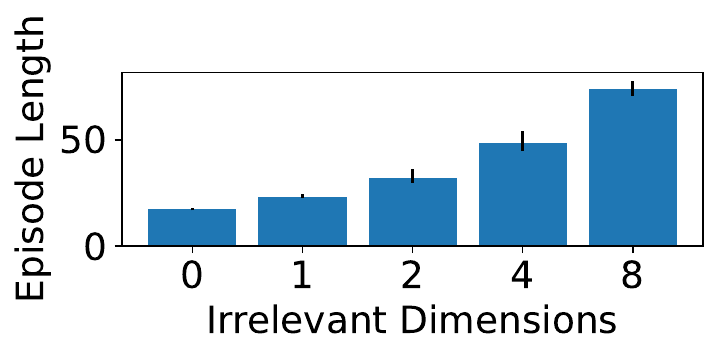}
            \caption[SAC]%
            {{\small SAC toy}}
            \label{fig:sac_irr_dims_len}
        \end{subfigure}
        \begin{subfigure}[]{0.384\textwidth}
            \centering
            \includegraphics[width=\textwidth]{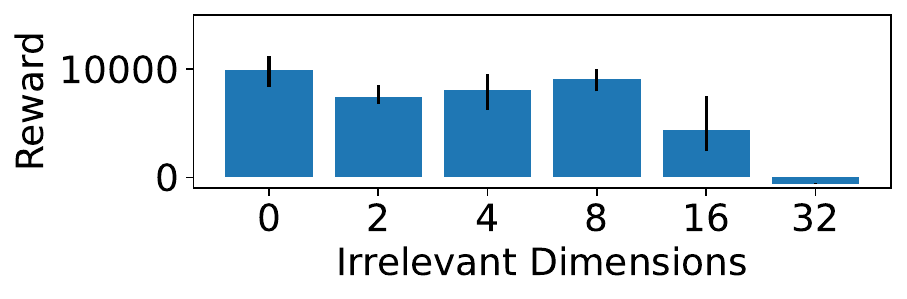}
            \caption[]%
            {{\footnotesize SAC HalfCheetah}}    
            \label{fig:sac_irr_dims_halfcheetah}
        \end{subfigure}

        \caption[]
        {AUC of selected metric for SAC at the end of training when varying \textbf{irrelevant dimensions} \textbf{(a)}: episodic reward on the toy environment. \textbf{(b)}: episode length on the toy environment. \textbf{(c)}: episodic reward on \textit{HalfCheetah}. Error bars represent bootstrapped confidence intervals with Bonferroni corrections for a significance level of 0.05.}
        \label{fig:sac_irr_dims}
\end{figure}

Introducing \textit{irrelevant dimensions} in to the toy continuous environments, while keeping the number of relevant dimensions fixed to $2$, decreased agent performance (Figure \ref{fig:sac_irr_dims_rew} for SAC; Figure \ref{fig:ddpg_irr_dims_rew append_fig}-\ref{fig:sac_irr_dims_rew append_fig} for all agents in the appendix). This gave us the insight that irrelevant features interfere with the learning~process. It was not just the episodic reward that was worse but also the mean episodic length (Figure \ref{fig:sac_irr_dims_len}; Figures \ref{fig:ddpg_irr_dims_len}-\ref{fig:append_sac_irr_dims_len} for all agents in the appendix), which meant that even though the agents were reaching the target point within the episode, they were taking longer to do so. We performed similar experiments with SAC on a complex environment (\textit{HalfCheetah}) and observed similar trends for performance (Figure \ref{fig:sac_irr_dims_halfcheetah}).

\subsubsection{Reward Density}
Figure \ref{fig:dqn_density} shows the results for DQN (Figure \ref{fig:algs_spar_train_1d} for all agents in the appendix) when controlling the \textit{reward density} on the toy environment. It is interesting that the density does not seem to impact the performance almost at all for any of the agents. We believe that is because once any of these agents finds a rewarding sequence (which is a single $N$-state in the case of a sequence length of 1), they do not really need to change their policy because the other rewarding sequences also result in the same reward. To allow for a more varied reward distribution to explore more interesting directions in such cases, we have an experimental dimension of hardness, the \textit{reward distribution} (for more information on how this dimension works, please see Appendix \ref{append_sec:exptl_dims}). While the above experiment might make it seem that the dimension of hardness reward density is not useful when the reward handed out is always 1, Section \ref{sec:debugging} shows an interesting use-case for it in the presence of $\epsilon$-greedy exploration.

\subsubsection{Further Discussion}
For some readers, it might feel self-evident that injecting many of these dimensions causes difficulties for agents. However, to the best of our knowledge, no other work has tried to collect so many dimensions of hardness in one place and study them comprehensively and what aspects transfer from toy to more complex environments. 

We believe that the \textit{transfer} of the hardness dimensions from toy to complex environments occurs because the algorithms we have tested are environment agnostic and usually do not take aspects of the environment into account.
Q-learning for instance is based on TD-errors and the Bellman equation. The equation is agnostic to the environment and while adding deep learning may help agents learn representations better, it does not remove the problems inherent in deep learning.
While it is nice to have general algorithms that may be applied in a blackbox fashion, by studying the dimensions we have listed and their effects on environments, we may gain deeper insights into what is needed to design better agents.


An additional comment can be made about comparing the continuous complex to toy comparisons to the discrete complex to toy comparisons. The ``noise'' in comparing the toy to the complex discrete environments was higher compared to comparing the continuous toy to complex environments and we believe this is due to the discrete environments being much more sparse and having many more \textit{lucky areas} that can be exploited as with the \textit{qbert} bug and \textit{breakout} strategy mentioned. In comparison, continuous environments usually employ a dense reward formulation in which case the value functions are likely to be continuous leading to less noisy comparisons. 

The experiments in this section also show how the complex environment wrappers allow researchers, who are curious, to study the robustness of their agents to these dimensions of hardness on complex environments, without having to fiddle with lower-level code.





\begin{figure}[ht]
        \centering
        \begin{subfigure}[]{0.45\textwidth}   
            \centering 
            \includegraphics[width=\textwidth]{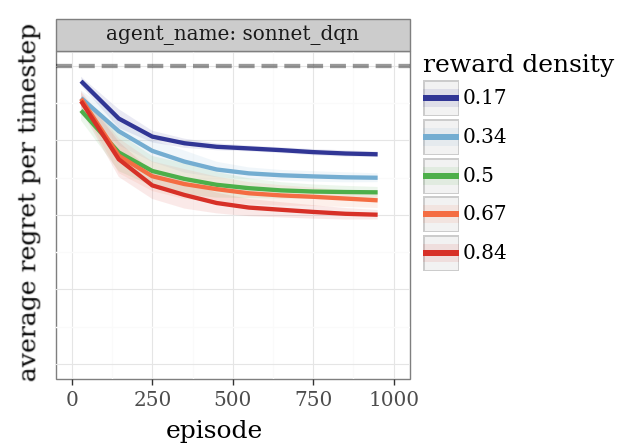}
            \caption[]%
            {{\small bsuite DQN}}    
            \label{fig:bsuite_r_sparsity}
        \end{subfigure}
        \begin{subfigure}[]{0.42\textwidth}
            \centering
            \includegraphics[width=\textwidth]{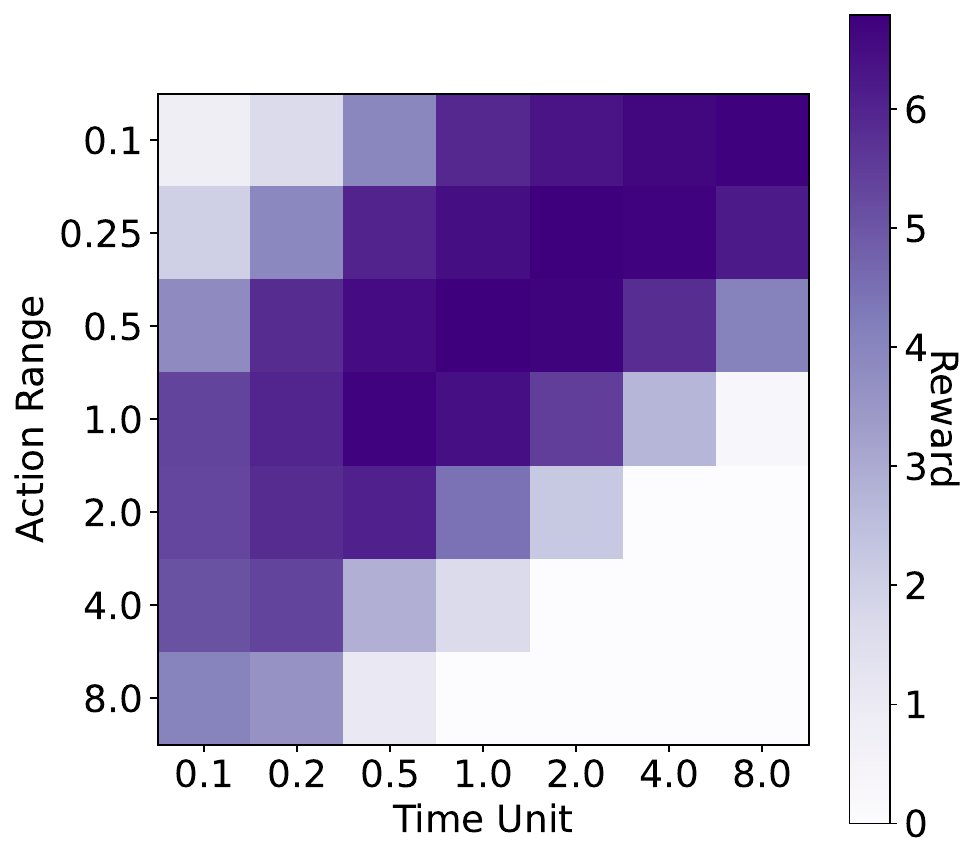}
            \caption[DDPG]%
            {{\small 2 dimensions together}}    
            \label{fig:ddpg_p_order_2_action_max_time_unit_main_paper}
        \end{subfigure}

        \caption[ ]
        {Analysing and Debugging. \textbf{(a)}: Varying \textit{reward density} on the toy environment for the \textit{bsuite} DQN agent. \textbf{(b)}: Varying 2 dimensions of hardness together - \textbf{action range} and \textbf{time unit}. 
        }

        \label{fig:radar_plots_and_others}
\end{figure}





\subsection{Debugging Agents}\label{sec:debugging}
Analysing how an agent performs under the effect of various dimensions of hardness can reveal unexpected aspects of the agent. For instance, when using bsuite agents \shortcite{bsuite_osb2019behaviour}, we noticed that when we varied our environment's \textit{reward density}, the performance of the bsuite DQN agent got worse as the reward got sparser (see Figure \ref{fig:bsuite_r_sparsity}). This did not occur for other bsuite agents. This behaviour with \textit{MDP Playground} occurs because the DQN agent used $\epsilon$-greedy while the other agents explored using other methods. $\epsilon$-greedy causes the agent to perform a completely random a certain fraction of the time which then leads to entering a terminal state a proportional fraction of the time and consequently to rewards in proportion to the reward density. Such insights can easily go unnoticed if the environments used are too complex. The simplicity of our toy environments helps debug such cases.

\begin{figure}[ht]
        \centering
        \begin{subfigure}[b]{0.435\textwidth}
            \centering
            \includegraphics[width=\textwidth]{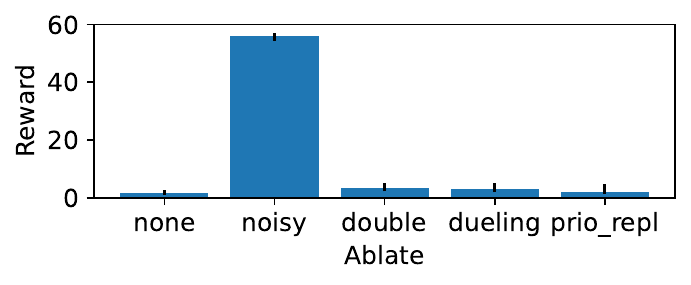}
            \caption[]%
            {{\small toy}}    
            \label{fig:ray_rllib_noisy_nets_toy}
        \end{subfigure}
        \begin{subfigure}[b]{0.435\textwidth}
            \centering
            \includegraphics[width=\textwidth]{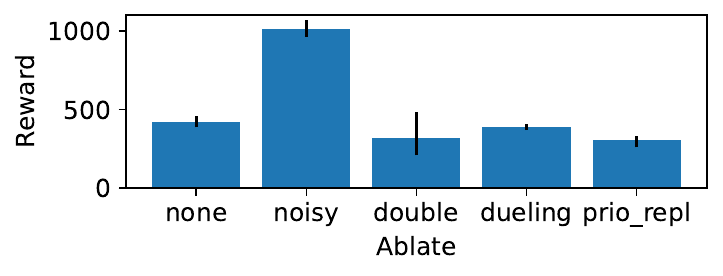}
            \caption[]%
            {{\small Beam Rider}}    
            \label{fig:ray_rllib_noisy_nets_beam_rider}
        \end{subfigure}

        \caption[]
        {Ablations of Rainbow on \textit{image representation} environments. X-axis labels mention which component was ablated\chngdjaircrc{, i.e., turned off}: \textit{noisy} represents noisy nets, \textit{double} represents Double DQN, \textit{dueling} represents Dueling Networks and \textit{prio\_repl} represents prioritised replay. Runs on \textit{Beam Rider} were only till 3M timesteps in this experiment. Turning noisy nets \textbf{off} improved performances on both toy and complex environments significantly. Error bars represent bootstrapped confidence intervals with Bonferroni corrections for a significance level of 0.05.}
        \label{fig:ray_rllib_noisy_nets_broken}
\end{figure}

In another example, in one of the Ray versions we used (\texttt{0.9.0.dev0}), we observed that DQN was performing well on the \textit{image representations} environment while Rainbow was performing poorly. We were quickly able to \textit{ablate} additional Rainbow hyperparameters on the toy environments and found that the noisy nets \shortcite{noisy_nets_fortunato_iclr_2018} implementation was broken (see Figure \ref{fig:ray_rllib_noisy_nets_toy}). We then tested and observed the same on a more complex environment, \textit{Beam Rider} (see Figure \ref{fig:ray_rllib_noisy_nets_beam_rider}). This shows how easily and quickly agents can be debugged to see if something major is broken.\footnote{An example of how hard it can be to debug RL agents can be found in this GitHub
issue for bsuite: \url{https://github.com/deepmind/bsuite/issues/20}.} This, in combination with their low computational cost, also makes a case to use the toy environments in Continuous Integration (CI) tests on code repositories.

\subsection{Further Use Cases}\label{sec:addnl_use_cases}
We now discuss further interesting use cases for \textit{MDP Playground}.

\subsubsection{Multiple Dimensions} It is possible to vary multiple dimensions of hardness at the same time in the same base environment in \textit{MDP Playground}. For instance, Figure \ref{fig:ddpg_p_order_2_action_max_time_unit_main_paper} shows the interaction effect (an inversely proportional relationship) between the \textit{action range} and the \textit{time unit} in the continuous toy environment with DDPG. This is an interesting insight and shows that the agent's \textit{action range} should depend on the \textit{time unit} or vice versa. On thinking more about this, it makes sense because: 1) if actions that are too large are allowed for large time steps, the agent might go out of control, while if large actions are allowed only for shorter time steps, the agent can still remain stable because it has fine-grained control; 2) allowing a large space of actions for large time steps increases the search space for optimal actions and make it a harder control problem. Plots for more such experiments can be found in Appendix \ref{sec:append_2d_heatmaps}, including varying both $P$ and $R$ \textit{noise}s together and varying \textit{delay} and \textit{sequence length} together in the toy discrete environments.

\subsubsection{Design and Analyse Experiments} We allow the user the ability to inject dimensions of hardness into toy or complex environments in a fine-grained manner. This can be used to define custom experiments with the dimensions of hardness. The results can be analysed in the accompanying \textit{Jupyter notebook}\footnote{\url{https://github.com/automl/mdp-playground/blob/master/plot_experiments.ipynb}}. 
%


\subsubsection{Ground Truth} Since \textit{MDP Playground} has the ground truth of the underlying MDPs available for the toy environments, it is a useful tool for designing agents that run on POMDPs and try to extract the underlying features of an MDP that make it fully observable. The ground truth could also be used, for instance, in model-based RL to test how well true aleatoric uncertainties are learned by a probabilistic model on a toy environment with stochasticity injected.


\subsubsection{Tabular Agents} The fact that the toy environments are so simple also allows the platform to be used to compare tabular and deep agents. To this end, we also evaluated the tabular baselines Q-learning \shortcite{sutton2018reinforcement}, Double Q-learning \shortcite{double_q_Hasselt_2010_nips} and SARSA \shortcite{sutton2018reinforcement} on the discrete non-image based environments with similar qualitative results as to those for deep agents. Results can be seen in Appendix \ref{sec:tabular_baselines}.


The experiments here are only a glimpse into the power and flexibility of \textit{MDP Playground}. Users can also upload custom functions for $P$ and $R$ and custom image representations $O$ (instead of the polygons from Figure \ref{fig:repr_learn_images}) and our platform takes care of injecting the other dimensions of hardness for them (wherever possible).

\section{Related Work}\label{sec:related_work}
Many of the other environments mentioned in the main paper are largely vision-based, which means that a large part of their problem solving receives benefits from advances in the vision community while our environments try to tackle pure RL problems in their most toy form. This also means that our experiments are extremely cheap, making them a good platform to test out new algorithms' robustness to different challenges in RL.

A parallel and independent work along similar lines as the MDP Playground, which was released a month before ours on arXiv, is the Behaviour Suite for RL (bsuite, \shortciteR{bsuite_osb2019behaviour}). In contrast to our \textit{generated} environments, bsuite \textit{collects} simple RL environments from the literature that are representative of various types of problems which occur in RL and tries to characterise RL algorithms. \textit{Bsuite} can be seen as an intermediate step between our toy MDPs and more complex environments, because the toy environments that \textit{bsuite} collects are more complex than ours and do not always have controllable dimensions of hardness that can be independently injected. This makes \textit{bsuite}'s dimensions not individually  controllable and \textit{atomic} like ours. Fine-grained control is a feature that sets our platform apart. \textit{bsuite} has a collection of \textit{presets} chosen by experts which work well but would be much harder to play around with. While \textit{MDP Playground} also has good presets through default values defined for experiments, it is much easier to configure. 
\textit{Bsuite}'s experiments are also more expensive than ours; while \textit{bsuite} is already quite cheap to run, \textit{MDP Playground} experiments are over an order of magnitude cheaper.
Furthermore, while \textit{bsuite} offers \textbf{no continuous control environments}, \textit{MDP Playground} provides discrete and continuous environments. This is important because several agents like DDPG, TD3, SAC are designed for continuous control. Unlike their framework, where currently there is no toy environment for Hierarchical RL (HRL) algorithms, the rewardable sequences that we describe also fits very well with HRL. Finally, in contrast to \textit{bsuite}, \textit{MDP Playground} offers wrappers to inject many of the identified dimensions of hardness into complex environments. We use these wrappers to also compare the identified trends on toy and complex environments. An important distinction between the two platforms could be summed up by saying that they try to characterise \textit{algorithms} while we try to characterise \textit{environments} with the aim that new adaptable algorithms can be developed that can tackle environments with specific challenges.

Toybox~\shortcite{tosch_toybox_arxiv_2019} and Minatar~\shortcite{young_minatar_arxiv_2019} are also cheap platforms with similar goals of gaining deeper insights into RL agents. However, their games target the specific \textit{Atari} domain and, like \textit{bsuite}, they are complementary to our approach as they have more specific environments and do not offer fine-grained control over dimensions of hardness.

We found the work of \shortciteA{andersson_benchmark_dimensions_icml_ws_2018} the most similar work to ours in spirit. In contrast to our platform, they only target continuous environments. We capture their dimensions of hardness in a different manner and offer many more dimensions with fine-grained control. Furthermore, their code is not open-source, making their work harder to build on by the community.

\shortciteA{mdp_hardness_nips_2014} defines a novel theoretical metric for defining hardness of MDPs and captures this hardness when the true state of the MDP is known. However, a large part of the hardness in our MDPs comes from the agent not knowing the optimal information state to use. It would be interesting to design a metric which captures this aspect of hardness as well.

Our platform allows formulating problems in terms of the identified dimensions and we feel this is a very human-understandable way of defining problems or specifying tasks. \shortciteA{littman_Task_Specifications_GLTL_2017} defines a Geometric Linear Temporal Logic (GLTL) specification language to formally specify tasks for MDPs and RL environments. They also share our motivation in making it easier and more natural to specify tasks.

Further related work includes \textit{Procgen}~\shortcite{cobbe_procgen_icml_20} and \textit{Obstacle~Tower}~\shortcite{obstacle_tower_Juliani_2019arXiv190201378J}.
Procgen adds various heterogeneous environments and tries to quantify generalisation in RL.
In a similar vein, Obstacle Tower provides a generalisation challenge for problems in vision, control, and planning. Unlike ours, the scope of these platforms is not being a testbed. As such, they do not capture dimensions of hardness that can be independently injected with fine-grained control over them, which we view as an important aspect when testing agents. \shortciteA{challenges_real_world_rl_arxiv_2003_11881} provides some overlapping dimensions of hardness with our platform but only for continuous environments, and has no toy environments.

\subsection{MDP Playground in Relation to MNIST}
MNIST \shortcite{mnist_lecun-mnisthandwrittendigit-2010} captured some key aspects of hardness required for computer vision (CV) which made it a good testbed for designing and debugging CV algorithms - the webpage for the dataset (\url{http://yann.lecun.com/exdb/mnist/} mentions some distortions for MNIST (similar to our image representation transforms): \textit{distortions are random combinations of shifts, scaling, skewing, and compression}. \shortciteA{mnist_c_mu_arxiv_19} captures $15$ such distortions to benchmark out-of-distribution robustness in CV. However, being a good testbed does not mean that MNIST can be used to directly learn models for much more specific CV applications such as classification of plants or medical image analysis. It captures many aspects that are general to CV problems but not specific ones, similar to our toy environments for RL.



\section{Limitations of the Approach and its Ethical and Societal Implications}
\label{sec:limitations_and_societal_impact}
While we do not see any limitations regarding the complex environment wrappers in \textit{MDP Playground} beyond the limitations of complex environments themselves, we would like to caution against some limitations for the toy environments. Toy environments in \textit{MDP Playground} are meant to be analysis and debug testbeds and not for tuning the final agent hyperparameters (HPs) for use on complex environments. They are extremely cheap compared to complex environments and (as one would expect), can only be used to draw high-level insights and are likely not as differentiating as complex environments for many of the finer changes between RL agents. As also mentioned in \shortciteA{benchmark_lottery_arxiv_21}, a lot of tricks may be involved in achieving SOTA scores on complex benchmarks and the toy environments cannot be expected to be predictive of performance on such benchmarks. Finally, Multi-Agent RL, Multi Objective RL, and Time Varying MDPs (amongst others) are beyond the scope of the currently implemented toy environments. \chngdjaircrc{We also note here that we currently consider partial observability only in the reward dynamics and not in the transition dynamics. Introducing partial observability also in the transition dynamics is an interesting avenue for future work.}

\chngdjaircrc{Another limitation is that in the complex environment wrappers, it is not always possible to control the exact ``amount'' of hardness in complex environments. Using the wrappers, however, always increases this existing amount. For dimensions of hardness like \textit{transition noise} and \textit{reward noise}, one can use the wrappers to control the exact amount of hardness. However, for dimensions of hardness like \textit{delay} and \textit{sequence length} which are usually already present in a complex environment and variable across the state space, it is not possible to control their exact amounts of hardness with the wrappers as we do not know how much of these exist already in the environment. It is a very hard problem to determine how much of each of them exists already and being able to determine this could go a long way to solving the RL problem itself, so there is no trivial way to get around this limitation of the wrappers in \textit{MDP Playground}.}

Further, high-dimensional control problems where there are interaction effects between degrees of freedom are not captured in the toy rigid body continuous control problem as this is the domain of complex environments and beyond the scope of this platform.

In terms of the broader impact on society and ethical considerations, we foresee no direct impact, only indirect consequences through RL since our work promotes standardisation and reproducibility. An additional environmental impact would be that, at least, prototyping and testing of agents could be made cheaper, reducing carbon emissions but also hopefully accelerating RL research.


\section{Conclusion and Future Work}
\label{sec:conclusion}
We introduced a platform to analyse and debug RL agents with fine-grained control over various dimensions of hardness. The platform allows low-cost experiments on toy environments and analysing the effects of dimensions of hardness on complex environments.
We studied the performance of various agents and gained insights on toy and complex environments by varying these dimensions. To the best of our knowledge, we are also the first to perform a principled study of how significant aspects such as non-Markov information states, irrelevant features, representations and low-level dimensions of hardness, like time discretisation, affect agent performance. Finally, we also described debugging capabilities of the platform.

We believe there is great potential to design agents that adapt the time unit, action range, etc. dynamically and test them on vanilla environments. To the best of our knowledge agents currently do not try to adapt most of these dynamically and yet as we know from observing humans, humans are adaptive to these dimensions of hardness. Previous theoretical breakthroughs initially came on very small toy examples while \textit{deep} breakthroughs came, \chngdjaircrc{in some cases}, decades later. If we are able to make agents more robust to some of these dimensions of hardness on toy examples, we may later be able to make it work on complex environments. Such agents could also improve explainability of RL agents because, as we know, humans can also provide rough estimates of the dimensions of hardness to explain their actions. For example, if a human buys a lottery ticket and wins the lottery a week later, they can say that it was the action of buying with a week's delay that got them their reward. This also has links to causality.

 \chngdjaircrc{An interesting distinction can also be made between agents' robustness to perturbations to a single task and agents learning to generalise across a distribution of tasks}. Having ground truth and fine-grained control over the dimensions of hardness allows \textit{MDP Playground} to also be used as a testbed for MetaRL which learns over distributions of tasks/environments. For example, it could be studied how robust MAML \shortcite{maml_finn_icml17} is to wider distributions of the same task, by simply varying the range of a dimension of hardness of \textit{MDP Playground}. It could also be used to study how the number of gradient descent steps performed from the common initialisation found by MAML interacts with the distribution it performs well over. While the exact values from such analyses may not carry over to complex domains, it is interesting to analyse whether such effects exist on a simpler domain and to then possibly formalise them theoretically \chngdjaircrc{and come up with numerical quantifications of the effects and their transfer across environments.}

Finally, we would like \textit{MDP Playground} to be a community-driven effort. It is open-source for the benefit of the RL community at \url{https://github.com/automl/mdp-playground}. While we tried to identify as many dimensions of hardness as possible, it is unlikely that we have captured \textit{all} such dimensions that can be independently injected in RL environments. We have some further dimensions of hardness currently in the experimental phase and we describe these in Appendix \ref{append_sec:exptl_dims}. We welcome more dimensions of hardness that readers think will help us encapsulate further challenges in RL and will add them to \textit{MDP Playground} based on the community's opinions. Given the current brittleness of RL agents \shortcite{deep_rl_that_matters}, and many claims that have been challenged \shortcite{atrey_saliency_counterfactual_iclr_2020,tosch_toybox_arxiv_2019}, we believe RL agents need to be tested on a lower and more basic level to gain insights into their inner workings.

\acks{The authors gratefully acknowledge support by BMBF grant DeToL, by the Bosch Center for Artificial Intelligence, by the European Research Council (ERC) under the European Union’s Horizon 2020 research and innovation programme under grant no. 716721, by the state of Baden-W\"{u}rttemberg through bwHPC, by the German Research Foundation (DFG) through grant no INST 39/963-1 FUGG and by TAILOR, a project funded by EU Horizon 2020 research and innovation programme under GA No 952215. \chngdjaircrc{The authors acknowledge funding through the research network ``Responsive and Scalable Learning for Robots Assisting Humans'' (ReScaLe) of the University of Freiburg. The ReScaLe project is funded by the Carl Zeiss Foundation.} They would like to thank their group, especially Joerg, Steven, Samuel, for helpful feedback and discussions. \chngdjaircrc{The authors are also grateful to the reviewers for their many helpful insights and feedback that helped improve the paper.} Raghu would like to additionally thank Michael Littman for his feedback and encou
ragement and the RLSS 2019, Lille organisers and participants who he had interesting discussions with.}








\appendix

\section{Experimental Dimensions in MDP Playground}\label{append_sec:exptl_dims}
We list here some dimensions that are still experimental in \textit{MDP Playground}. 

\begin{itemize}
\denselist
\item[$\ast$] Only for discrete toy environments:
    \begin{itemize}
    \denselist
    \item[$\bullet$] \textbf{Reward Distribution}: When the reward distribution option is turned on, it can be specified as a list with 2 floats. These 2 values are interpreted as a closed interval and taken as the end support points of a categorical distribution with support points equally spaced along the interval. Each support point is assigned randomly to one of the automatically generated rewardable sequences, so that when a rewardable sequence is achieved, the reward handed out is the value of the support point. Further different kinds of reward distributions that could lead to interesting insights are an interesting avenue to develop this dimension of hardness further.
    \end{itemize}
\item[$\ast$] Only for continuous toy environments:
    \begin{itemize}
    \denselist
    \item[$\bullet$] \textbf{Reward Function}: The reward function is currently fixed to \texttt{move\_to\_a\_point} which represents the task of moving the point mass to a \textit{target point}. Additionally, we have a toy task of moving along a line which can be specified by setting the reward function to \textit{move\_along\_a\_line}. For this task, we hand out greater rewards the closer a point object is to moving linearly for $n$ consecutive steps where $n$ is the sequence length. The linear movement can be along \textit{any} direction, the important thing is that $n$ consecutive movements of the point mass should be in a straight line to achieve the reward. This allows evaluating agents on a sequential/navigation task of moving along a line. However, it is currently in the experimental phase because it turned out to be too hard for the agents and we probably need to find a better formulation for the task so that it allows us more valuable insights into agents.
    \end{itemize}
\end{itemize}

\section{Additional Details on Toy Experiments}\label{sec:append_addnl_reward_plots}
We provide some additional detail on the experiments in the main paper here, including the remaining plots referenced in the discussion in the main paper.


\subsection{Selecting Total Timesteps for Discrete Toy Runs}\label{sec:append_max_timesteps_discussion}

\begin{figure}[ht]
        \centering
            \centering
            \includegraphics[width=0.45\textwidth]{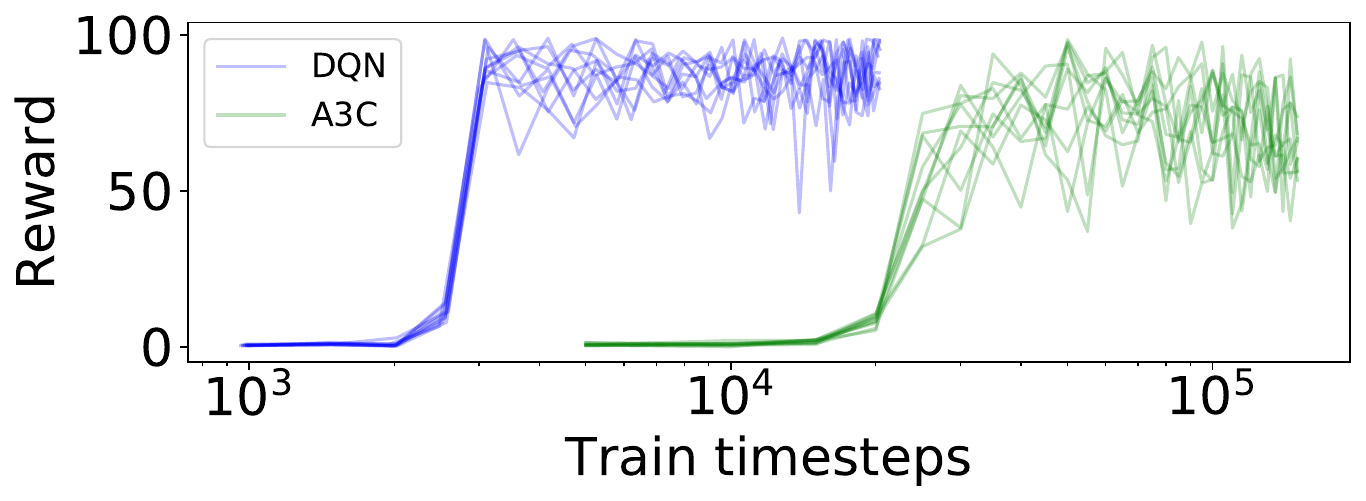}
            \caption[DQN]%
            {{\small Evaluation rollouts for DQN and A3C in the vanilla environment which shows that DQN learns faster than A3C in terms of the number of timesteps.}}
            \label{fig:dqn_a3c_combined_eval_curves}
\end{figure}

We ran the experiments and plot the results for DQN variants up to $20\,000$ environment timesteps and the ones for A3C variants up to $150\,000$ time steps since A3C took longer\footnote{In terms of environment steps. Wallclock time used was still similar.} to learn as can be seen in Figure \ref{fig:dqn_a3c_combined_eval_curves}. We refrain from fixing a single number of timesteps for our environments (as, e.g., bsuite does), since the study of different trends for different families of algorithms will require different numbers of timesteps. Policy gradient methods such as A3C tend to be slower in terms of time steps compared to value-based approaches such as DQN. Throughout, we always run $100$ seeds of all algorithms to obtain reliable results. We repeated many of our experiments with an independent set of $100$ seeds and obtained the same qualitative results.

\clearpage
\subsection{Additional Plots}
We provide here additional plots for the various experiments we ran on \textit{MDP Playground}.
\begin{figure}[ht]
        \centering
        \begin{subfigure}[]{0.435\textwidth}
            \centering
            \includegraphics[width=\textwidth]{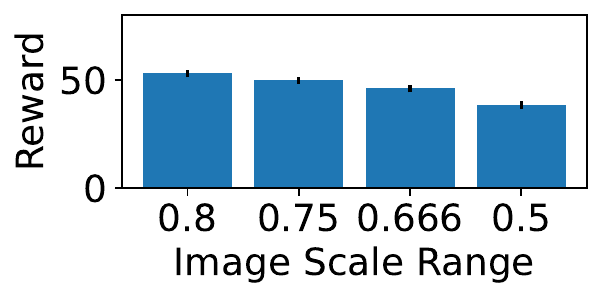}
            \caption[DQN]%
            {{\small Scale Range}} \label{fig:dqn_scale_range}
        \end{subfigure}
        \begin{subfigure}[]{0.465\textwidth}
            \centering
            \includegraphics[width=\textwidth]{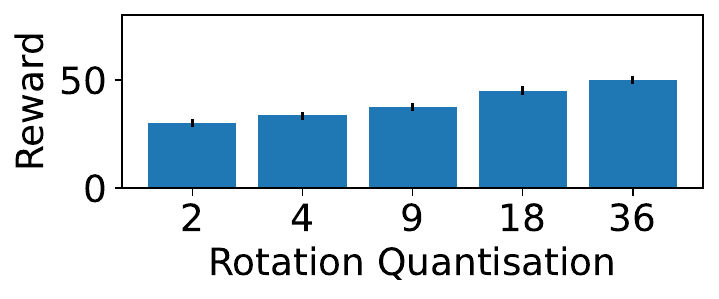}
            \caption[DQN]%
            {{\small Rotate}}
            \label{fig:dqn_ro_quant}
        \end{subfigure}

        \caption[ AUC of episodic reward at the end of training ]
        {AUC of episodic reward at the end of training for DQN when varying \textbf{the quantisation/range of transforms for image representations}. \textbf{(a)}: \textit{Image Scale Range} represents \textit{scaling} ranges. The X-axis ticks are the lower end of the image scale range, the upper end is just the inverse of the tick value shown. \textbf{(b)}: \textit{Rotation quantisation} represents quantisation of the \textit{rotation}s. Error bars represent bootstrapped confidence intervals with Bonferroni corrections for a significance level of 0.05.}
        \label{fig:append_repr_learn}
\end{figure}

\begin{figure}[ht]
        \centering
        \begin{subfigure}[]{0.325\textwidth}
            \centering
            \includegraphics[width=\textwidth]{figures/dqn_del_train_final_reward_delay_episode_reward_mean_1d.pdf}
            \caption[DQN]%
            {{\small DQN}}    
            \label{fig:DQN_del_1d a}
        \end{subfigure}
        \begin{subfigure}[]{0.325\textwidth}   
            \centering 
            \includegraphics[width=\textwidth]{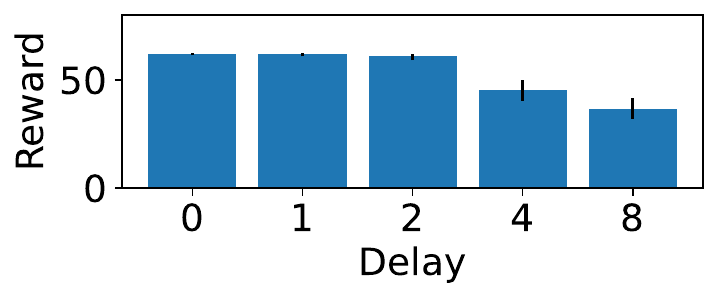}
            \caption[Rainbow]%
            {{\small Rainbow }}    
            \label{fig:rainbow_del_1d b}
        \end{subfigure}
        \begin{subfigure}[]{0.325\textwidth}
            \centering
            \includegraphics[width=\textwidth]{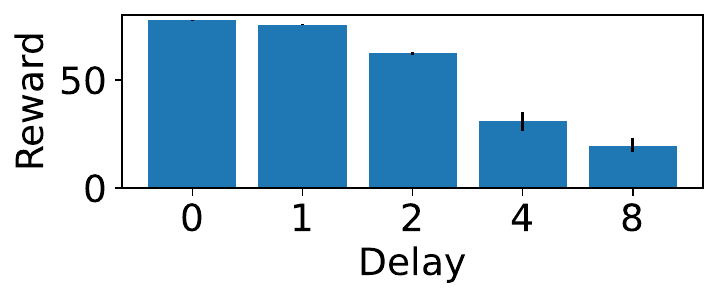}
            \caption[A3C]%
            {{\small A3C }}
            \label{fig:a3c_del_1d c}
        \end{subfigure}
        
        \caption[ AUC of episodic reward at the end of training ]
        {AUC of episodic reward at the end of training for DQN, Rainbow, A3C when varying \textbf{delays} on the discrete toy environment. Error bars represent bootstrapped confidence intervals with Bonferroni corrections for a significance level of 0.05.}
        \label{fig:algs_del_1d}
\end{figure}

\begin{figure}[ht]
        \centering
        \begin{subfigure}[]{0.325\textwidth}
            \centering
            \includegraphics[width=\textwidth]{figures/dqn_seq_train_final_reward_sequence_length_episode_reward_mean_1d.pdf}
            \caption[DQN]%
            {{\small DQN }}
            \label{fig:append_DQN_seq_1d a}
        \end{subfigure}
        \begin{subfigure}[]{0.325\textwidth}   
            \centering 
            \includegraphics[width=\textwidth]{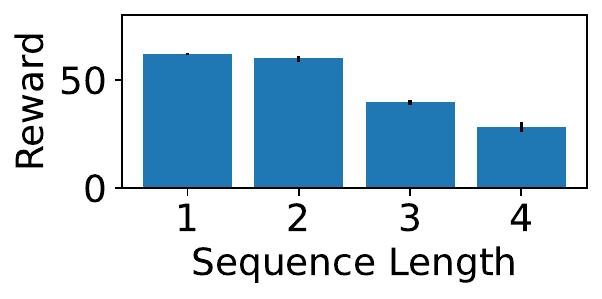}
            \caption[Rainbow]%
            {{\small Rainbow }}    
            \label{fig:rainbow_seq_1d b}
        \end{subfigure}
        \begin{subfigure}[]{0.325\textwidth}
            \centering
            \includegraphics[width=\textwidth]{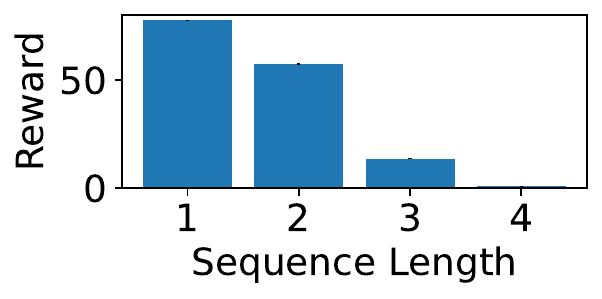}
            \caption[A3C]%
            {{\small A3C }}    
            \label{fig:a3c_seq_1d c}
        \end{subfigure}

        \caption[ AUC of episodic reward at the end of training ]
        {AUC of episodic reward at the end of training for DQN, Rainbow, A3C when varying \textbf{sequence lengths} on the discrete toy environment. Error bars represent bootstrapped confidence intervals with Bonferroni corrections for a significance level of 0.05.}
        \label{fig:algs_seq_len_1d}
\end{figure}




\begin{figure}[ht]
        \centering
        \begin{subfigure}[]{0.325\textwidth}
            \centering
            \includegraphics[width=\textwidth]{figures/dqn_p_noise_train_final_reward_transition_noise_episode_reward_mean_1d.pdf}
            \caption[DQN]%
            {{\small DQN}}    
            \label{fig:append_dqn_p_noise_1d}
        \end{subfigure}
        \begin{subfigure}[]{0.325\textwidth}   
            \centering 
            \includegraphics[width=\textwidth]{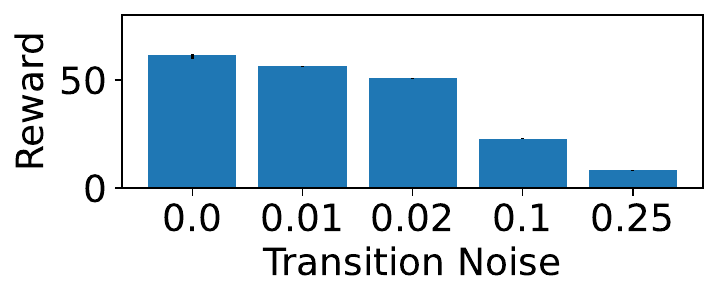}
            \caption[Rainbow]%
            {{\small Rainbow}}    
            \label{fig:rainbow_p_noise_1d append_fig}
        \end{subfigure}
        \begin{subfigure}[]{0.325\textwidth}
            \centering
            \includegraphics[width=\textwidth]{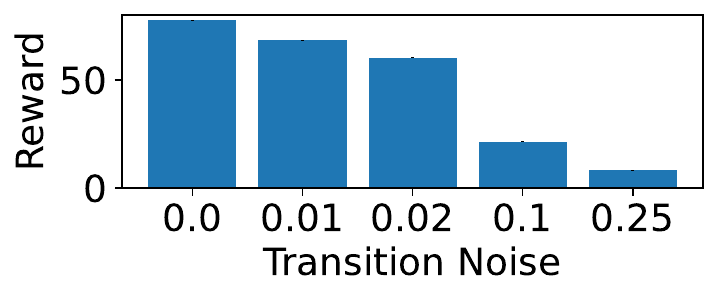}
            \caption[A3C]%
            {{\small A3C}}    
            \label{fig:a3c_p_noise_1d}
        \end{subfigure}

        \caption[ AUC of episodic reward at the end of training ]
        {AUC of episodic reward at the end of training for DQN, Rainbow, A3C when varying \textbf{transition noise} on the discrete toy environment. Error bars represent bootstrapped confidence intervals with Bonferroni corrections for a significance level of 0.05.}
        \label{fig:algs_p_noise_train_1d}
\end{figure}

\begin{figure}[ht]
        \centering
        \begin{subfigure}[]{0.325\textwidth}
            \centering
            \includegraphics[width=\textwidth]{figures/dqn_r_noise_train_final_reward_reward_noise_episode_reward_mean_1d.pdf}
            \caption[DQN]%
            {{\small DQN}}    
            \label{fig:append_dqn_r_noise_1d}
        \end{subfigure}
        \begin{subfigure}[]{0.325\textwidth}   
            \centering 
            \includegraphics[width=\textwidth]{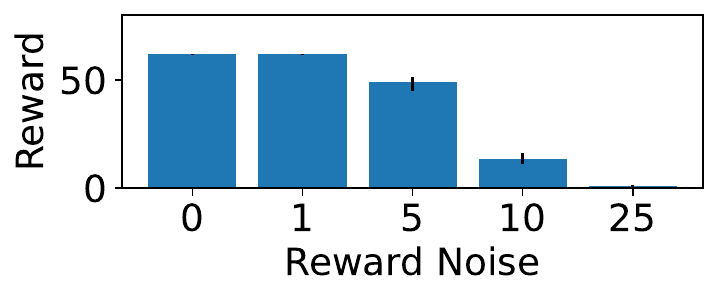}
            \caption[Rainbow]%
            {{\small Rainbow}}    
            \label{fig:rainbow_r_noise_1d}
        \end{subfigure}
        \begin{subfigure}[]{0.325\textwidth}
            \centering
            \includegraphics[width=\textwidth]{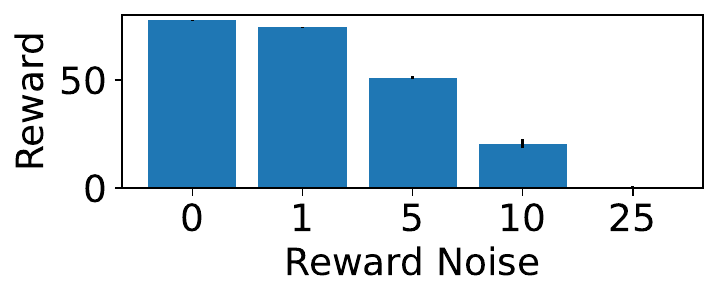}
            \caption[A3C]%
            {{\small A3C}}    
            \label{fig:a3c_r_noise_1d}
        \end{subfigure}

        \caption[ AUC of episodic reward at the end of training ]
        {AUC of episodic reward at the end of training for DQN, Rainbow, A3C when varying \textbf{reward noise} on the discrete toy environment. Error bars represent bootstrapped confidence intervals with Bonferroni corrections for a significance level of 0.05.}
        \label{fig:algs_r_noise_train_1d}
\end{figure}

\begin{figure}[ht]
        \centering
        \begin{subfigure}[]{0.325\textwidth}
            \centering
            \includegraphics[width=\textwidth]{figures/dqn_p_noise_eval_final_reward_transition_noise_episode_reward_mean_1d.pdf}
            \caption[DQN]%
            {{\small DQN}}    
            \label{fig:append_dqn_p_noise_eval_1d}
        \end{subfigure}
        \begin{subfigure}[]{0.325\textwidth}   
            \centering 
            \includegraphics[width=\textwidth]{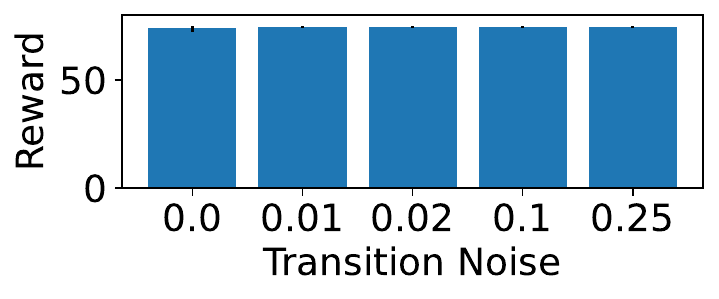}
            \caption[Rainbow]%
            {{\small Rainbow}}    
            \label{fig:rainbow_p_noise_eval_1d}
        \end{subfigure}
        \begin{subfigure}[]{0.325\textwidth}
            \centering
            \includegraphics[width=\textwidth]{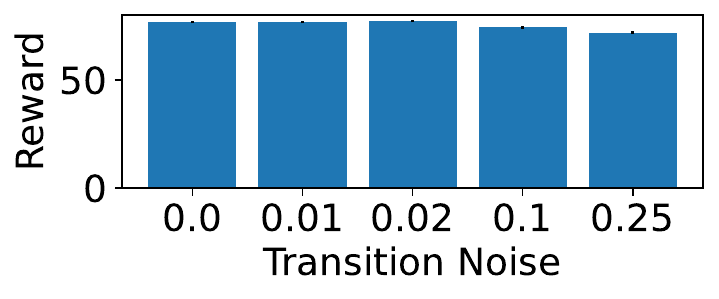}
            \caption[A3C]%
            {{\small A3C}}    
            \label{fig:a3c_p_noise_eval_1d}
        \end{subfigure}

        \caption[ Mean episodic evaluation rollout reward at the end of training ]
        {AUC of episodic reward when rolling out the policy that was learned on the noisy environment on a noise-free setting at the end of training for DQN, Rainbow, A3C when varying \textbf{transition noise} on the discrete toy environment. Error bars represent bootstrapped confidence intervals with Bonferroni corrections for a significance level of 0.05.}
        \label{fig:algs_p_noise_eval_1d}
\end{figure}

\begin{figure}[ht]
        \centering
        \begin{subfigure}[]{0.325\textwidth}
            \centering
            \includegraphics[width=\textwidth]{figures/dqn_r_noise_eval_final_reward_reward_noise_episode_reward_mean_1d.pdf}
            \caption[DQN]%
            {{\small DQN}}    
            \label{fig:append_dqn_r_noise_eval_1d}
        \end{subfigure}
        \begin{subfigure}[]{0.325\textwidth}   
            \centering 
            \includegraphics[width=\textwidth]{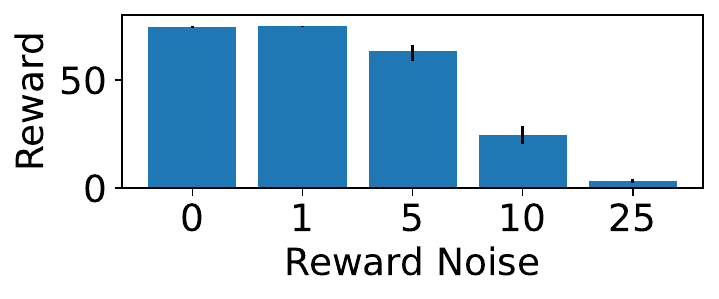}
            \caption[Rainbow]%
            {{\small Rainbow}}    
            \label{fig:rainbow_r_noise_eval_1d}
        \end{subfigure}
        \begin{subfigure}[]{0.325\textwidth}
            \centering
            \includegraphics[width=\textwidth]{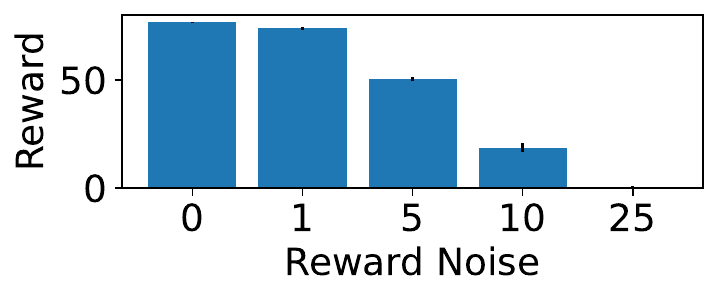}
            \caption[A3C]%
            {{\small A3C}}    
            \label{fig:a3c_r_noise_eval_1d}
        \end{subfigure}

        \caption[ Mean episodic evaluation rollout reward at the end of training ]
        {AUC of episodic reward when rolling out the policy that was learned on the noisy environment on a noise-free setting at the end of training for DQN, Rainbow, A3C when varying \textbf{reward noise} on the discrete toy environment. Error bars represent bootstrapped confidence intervals with Bonferroni corrections for a significance level of 0.05.}
        \label{fig:algs_r_noise_eval_1d}
\end{figure}

\begin{figure}[ht]
        \centering
        \begin{subfigure}[]{0.325\textwidth}
            \centering
            \includegraphics[width=\textwidth]{figures/dqn_sparsity_train_final_reward_reward_density_episode_reward_mean_1d.pdf}
            \caption[DQN]%
            {{\small DQN}}    
            \label{fig:append_dqn_spar_1d}
        \end{subfigure}
        \begin{subfigure}[]{0.325\textwidth}   
            \centering 
            \includegraphics[width=\textwidth]{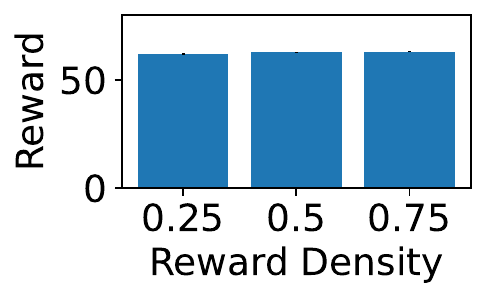}
            \caption[Rainbow]%
            {{\small Rainbow}}    
            \label{fig:rainbow_spar_1d}
        \end{subfigure}
        \begin{subfigure}[]{0.325\textwidth}
            \centering
            \includegraphics[width=\textwidth]{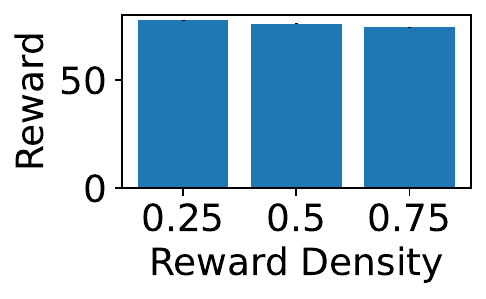}
            \caption[A3C]%
            {{\small A3C}}    
            \label{fig:a3c_spar_1d}
        \end{subfigure}

        \caption[ AUC of episodic reward at the end of training ]
        {AUC of episodic reward at the end of training for DQN, Rainbow, A3C when varying \textbf{reward density} on the discrete toy environment. Error bars represent bootstrapped confidence intervals with Bonferroni corrections for a significance level of 0.05.}
        \label{fig:algs_spar_train_1d}
\end{figure}

\begin{figure}[ht]
        \centering
        \begin{subfigure}[]{0.325\textwidth}
            \centering
            \includegraphics[width=\textwidth]{figures/ddpg_move_to_a_point_target_radius_train_final_reward_target_radius_episode_reward_mean_1d.pdf}
            \caption[DDPG]%
            {{\footnotesize DDPG}}    
            \label{fig:append_ddpg_target_radius_rew}
        \end{subfigure}
        \begin{subfigure}[]{0.325\textwidth}
            \centering
            \includegraphics[width=\textwidth]{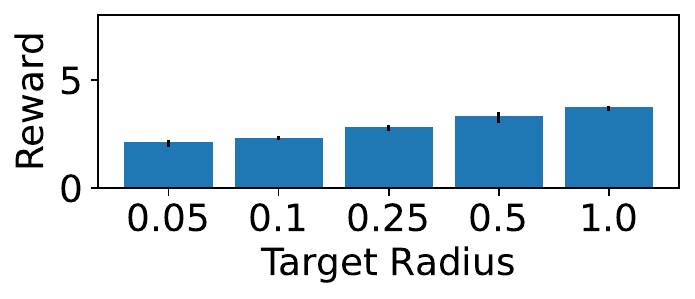}
            \caption[TD3]%
            {{\small TD3}}    
            \label{fig:td3_target_radius_rew}
        \end{subfigure}
        \begin{subfigure}[]{0.325\textwidth}
            \centering
            \includegraphics[width=\textwidth]{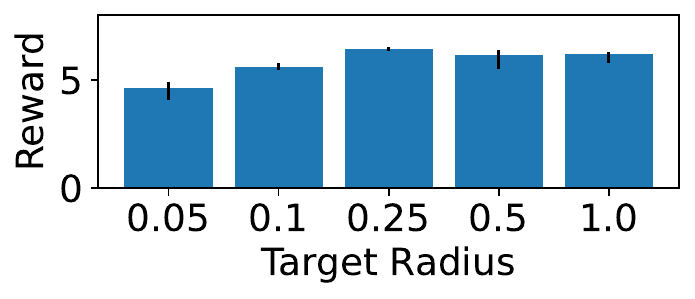}
            \caption[SAC]%
            {{\small SAC}}    
            \label{fig:sac_target_radius_rew}
        \end{subfigure}

        \begin{subfigure}[]{0.325\textwidth}
            \centering
            \includegraphics[width=\textwidth]{figures/ddpg_move_to_a_point_target_radius_train_final_reward_target_radius_episode_len_mean_1d.pdf}
            \caption[DDPG]%
            {{\small DDPG}}
            \label{fig:append_ddpg_target_radius_len}
        \end{subfigure}
        \begin{subfigure}[]{0.325\textwidth}
            \centering
            \includegraphics[width=\textwidth]{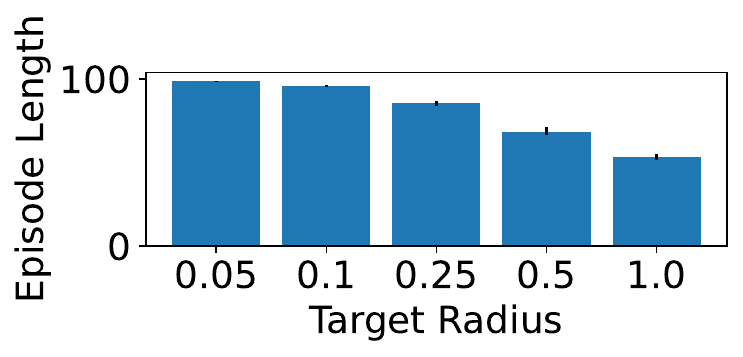}
            \caption[TD3]%
            {{\small TD3}}
            \label{fig:td3_target_radius_len}
        \end{subfigure}
        \begin{subfigure}[]{0.325\textwidth}
            \centering
            \includegraphics[width=\textwidth]{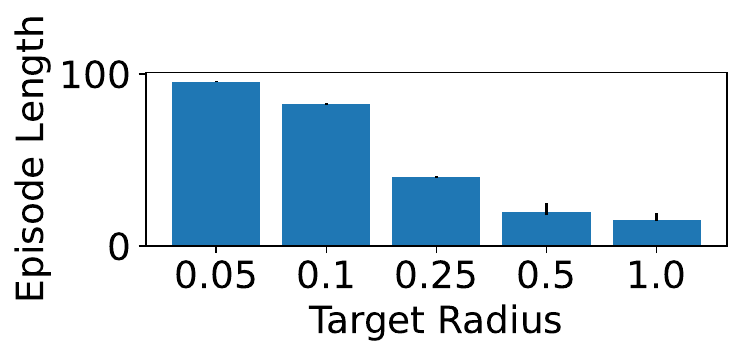}
            \caption[SAC]%
            {{\small SAC}}
            \label{fig:sac_target_radius_len}
        \end{subfigure}

        \caption[ AUC of episodic reward and lengths at the end of training ]
        {AUC of episodic reward (top) and lengths (bottom) for DDPG, TD3, SAC at the end of training when varying \textbf{target radius} on the discrete continuous environment. Error bars represent bootstrapped confidence intervals with Bonferroni corrections for a significance level of 0.05.}
        \label{fig:target_radius_toy_continuous}
\end{figure}

\begin{figure}[ht]
        \centering
        \begin{subfigure}[]{0.325\textwidth}   
            \centering 
            \includegraphics[width=\textwidth]{figures/ddpg_move_to_a_point_action_max_train_final_reward_action_space_max_episode_reward_mean_1d.pdf}
            \caption[DDPG]%
            {{\small DDPG}}    
            \label{fig:append_ddpg_action_space_max_rew}
        \end{subfigure}
        \begin{subfigure}[]{0.325\textwidth}   
            \centering 
            \includegraphics[width=\textwidth]{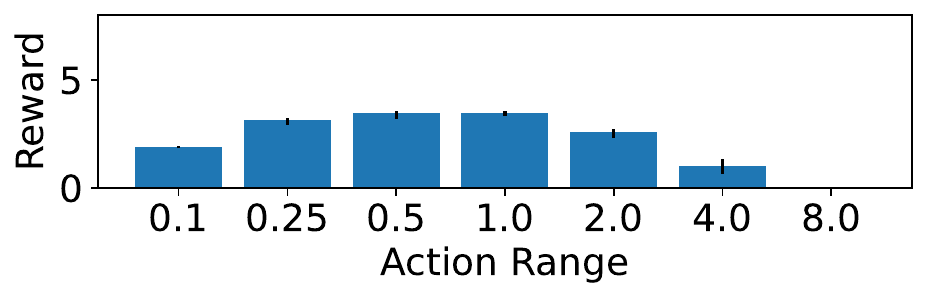}
            \caption[TD3]%
            {{\small TD3}}    
            \label{fig:td3_action_space_max_rew}
        \end{subfigure}
        \begin{subfigure}[]{0.325\textwidth}   
            \centering 
            \includegraphics[width=\textwidth]{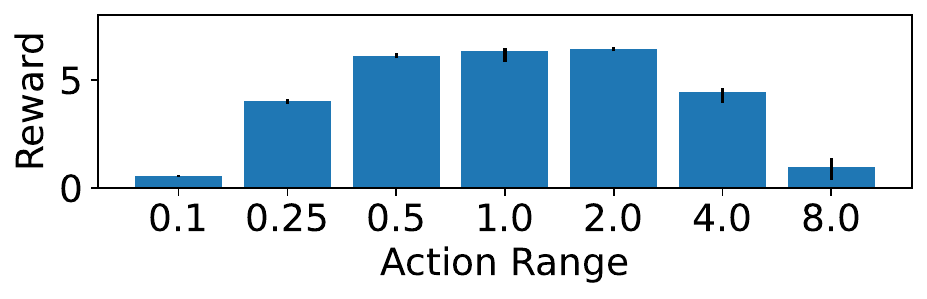}
            \caption[SAC]%
            {{\small SAC}}    
            \label{fig:sac_action_space_max_rew}
        \end{subfigure}

        \begin{subfigure}[]{0.325\textwidth}   
            \centering 
            \includegraphics[width=\textwidth]{figures/ddpg_move_to_a_point_action_max_train_final_reward_action_space_max_episode_len_mean_1d.pdf}
            \caption[DDPG]%
            {{\small DDPG}}    
            \label{fig:append_ddpg_action_space_max_len}
        \end{subfigure}
        \begin{subfigure}[]{0.325\textwidth}   
            \centering 
            \includegraphics[width=\textwidth]{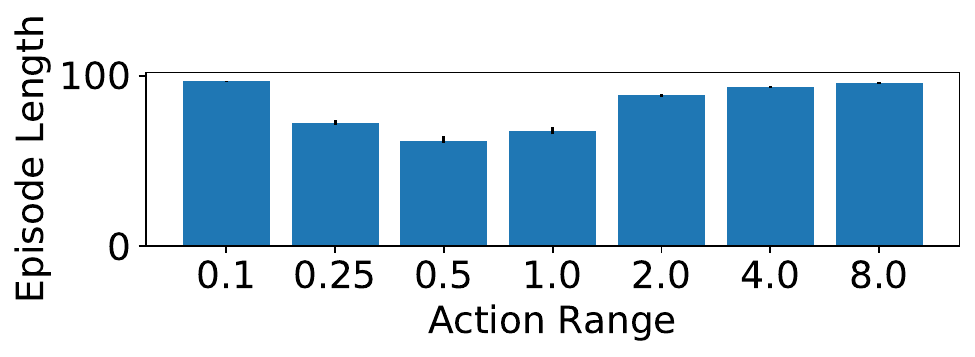}
            \caption[TD3]%
            {{\small TD3}}    
            \label{fig:td3_action_space_max_len}
        \end{subfigure}
        \begin{subfigure}[]{0.325\textwidth}   
            \centering 
            \includegraphics[width=\textwidth]{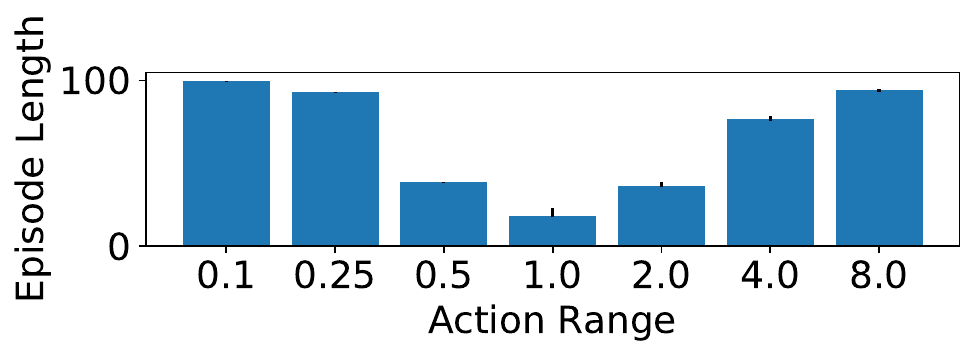}
            \caption[SAC]%
            {{\small SAC}}    
            \label{fig:sac_action_space_max_len}
        \end{subfigure}

        \caption[ AUC of episodic reward and lengths at the end of training ]
        {AUC of episodic reward (above) and lengths (below) for DDPG, TD3, SAC at the end of training when varying \textbf{action range} on the discrete continuous environment. Error bars represent bootstrapped confidence intervals with Bonferroni corrections for a significance level of 0.05.}
        \label{fig:act_max_toy_continuous}
\end{figure}

\begin{figure}[ht]
        \centering
        \begin{subfigure}[]{0.325\textwidth}
            \centering
            \includegraphics[width=\textwidth]{figures/ddpg_move_to_a_point_time_unit_train_final_reward_time_unit_episode_reward_mean_1d.pdf}
            \caption[DDPG]%
            {{\small DDPG}}    
            \label{fig:ddpg_time_unit_rew append_fig}
        \end{subfigure}
        \begin{subfigure}[]{0.325\textwidth}
            \centering
            \includegraphics[width=\textwidth]{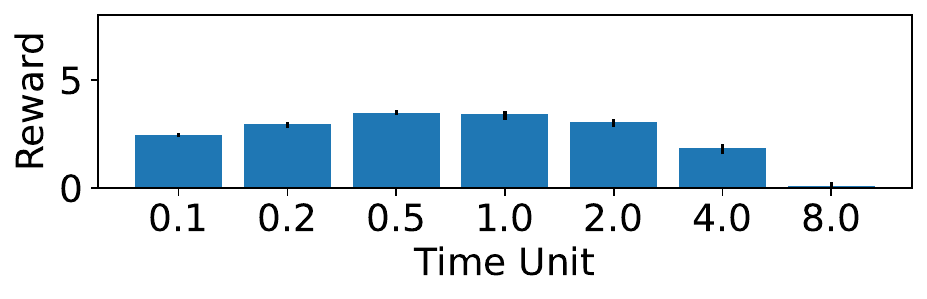}
            \caption[TD3]%
            {{\small TD3}}    
            \label{fig:td3_time_unit_rew}
        \end{subfigure}
        \begin{subfigure}[]{0.325\textwidth}
            \centering
            \includegraphics[width=\textwidth]{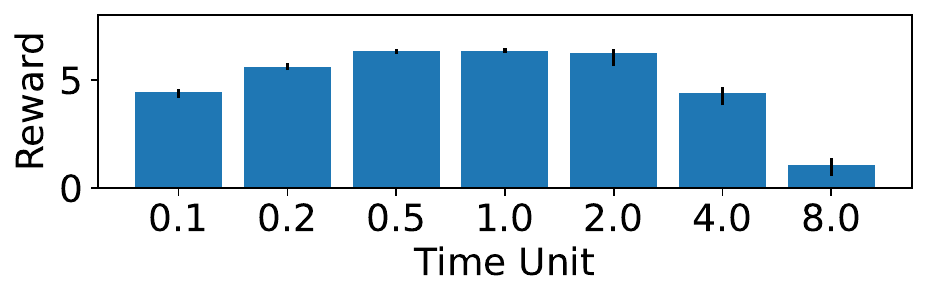}
            \caption[SAC]%
            {{\small SAC}}    
            \label{fig:sac_time_unit_rew}
        \end{subfigure}

        \begin{subfigure}[]{0.325\textwidth}
            \centering
            \includegraphics[width=\textwidth]{figures/ddpg_move_to_a_point_time_unit_train_final_reward_time_unit_episode_len_mean_1d.pdf}
            \caption[DDPG]%
            {{\small DDPG}}
            \label{fig:append_ddpg_time_unit_len}
        \end{subfigure}
        \begin{subfigure}[]{0.325\textwidth}
            \centering
            \includegraphics[width=\textwidth]{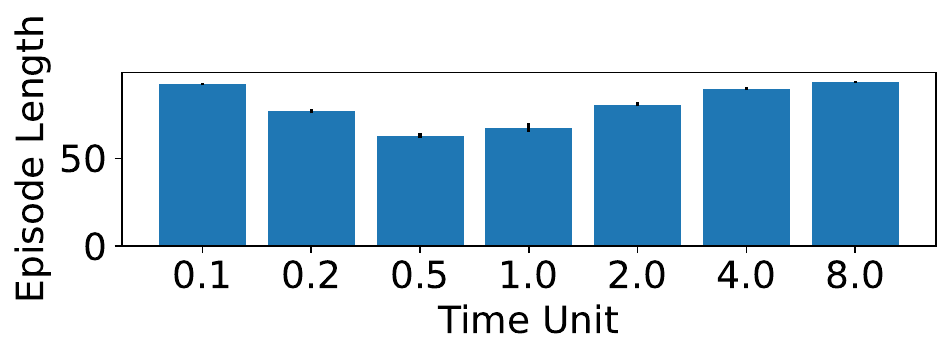}
            \caption[TD3]%
            {{\small TD3}}
            \label{fig:td3_time_unit_len}
        \end{subfigure}
        \begin{subfigure}[]{0.325\textwidth}
            \centering
            \includegraphics[width=\textwidth]{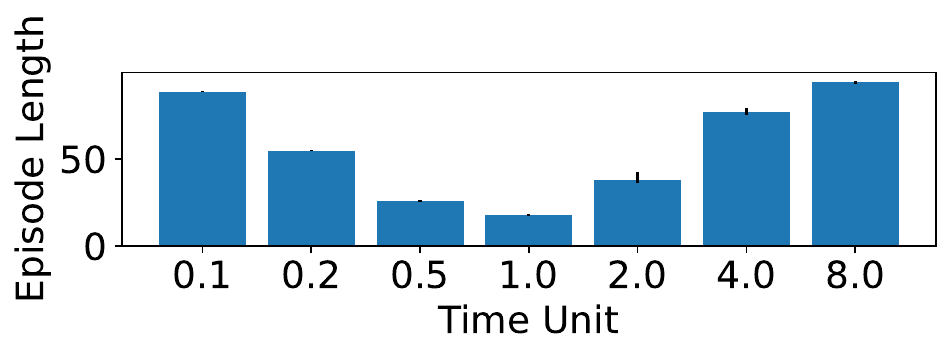}
            \caption[SAC]%
            {{\small SAC}}
            \label{fig:sac_time_unit_len}
        \end{subfigure}

        \caption[ AUC of episodic reward and lengths at the end of training ]
        {AUC of episodic reward (above) and lengths (below) for DDPG, TD3, SAC at the end of training when varying \textbf{time unit} on the discrete continuous environment. Error bars represent bootstrapped confidence intervals with Bonferroni corrections for a significance level of 0.05.}
        \label{fig:time_unit_toy_continuous}
\end{figure}

\begin{figure}[ht]
        \centering
        \begin{subfigure}[]{0.325\textwidth}   
            \centering 
            \includegraphics[width=\textwidth]{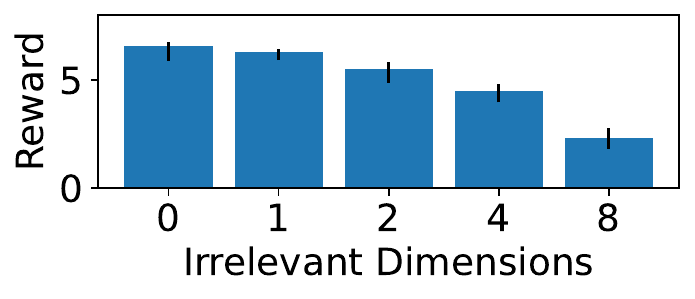}
            \caption[DDPG]%
            {{\small DDPG}}
            \label{fig:ddpg_irr_dims_rew append_fig}
        \end{subfigure}
        \begin{subfigure}[]{0.325\textwidth}   
            \centering 
            \includegraphics[width=\textwidth]{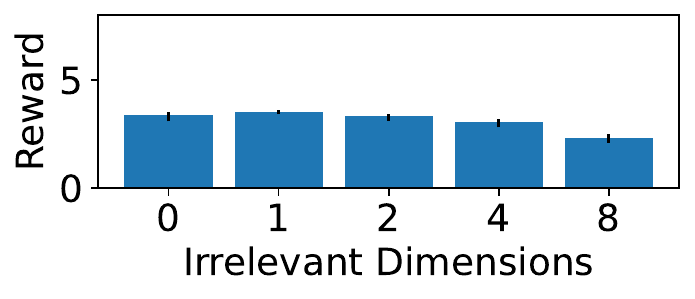}
            \caption[TD3]%
            {{\small TD3}}
            \label{fig:td3_irr_dims_rew}
        \end{subfigure}
        \begin{subfigure}[]{0.325\textwidth}   
            \centering 
            \includegraphics[width=\textwidth]{figures/sac_move_to_a_point_irr_dims_train_final_reward_state_space_dim_episode_reward_mean_1d.pdf}
            \caption[SAC]%
            {{\small SAC}}
            \label{fig:sac_irr_dims_rew append_fig}
        \end{subfigure}

        \begin{subfigure}[]{0.325\textwidth}   
            \centering 
            \includegraphics[width=\textwidth]{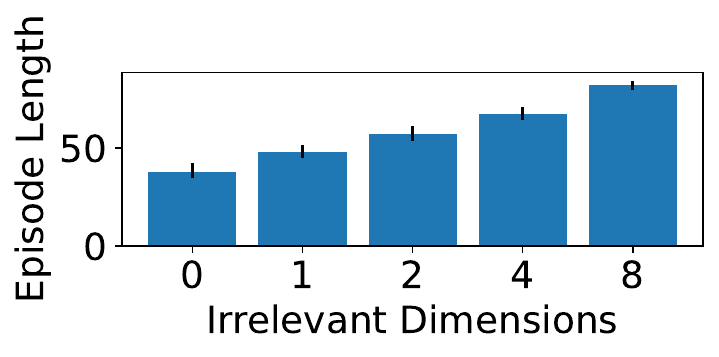}
            \caption[DDPG]%
            {{\small DDPG}}
            \label{fig:ddpg_irr_dims_len}
        \end{subfigure}
        \begin{subfigure}[]{0.325\textwidth}   
            \centering 
            \includegraphics[width=\textwidth]{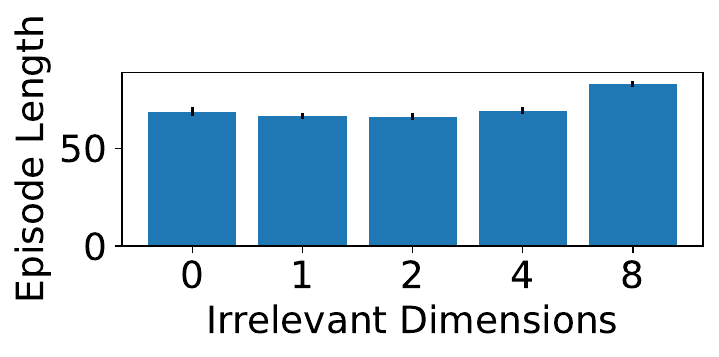}
            \caption[TD3]%
            {{\small TD3}}
            \label{fig:td3_irr_dims_len}
        \end{subfigure}
        \begin{subfigure}[]{0.325\textwidth}   
            \centering 
            \includegraphics[width=\textwidth]{figures/sac_move_to_a_point_irr_dims_train_final_reward_state_space_dim_episode_len_mean_1d.pdf}
            \caption[SAC]%
            {{\small SAC}}
            \label{fig:append_sac_irr_dims_len}
        \end{subfigure}

        \caption[ AUC of episodic reward and lengths at the end of training ]
        {AUC of episodic reward (above) and lengths (below) for DDPG, TD3, SAC at the end of training when varying \textbf{irrelevant dimensions} on the discrete continuous environment. Error bars represent bootstrapped confidence intervals with Bonferroni corrections for a significance level of 0.05.}
        \label{fig:irr_dims_toy_continuous}
\end{figure}

\FloatBarrier
\subsection{Algorithms for Automatically Generated Toy MDPs with MDP Playground}\label{append_sec:algorithms}


\begin{algorithm}[ht]
\caption{Automatically Generated Discrete Toy MDPs with MDP Playground}
\label{algorithm:generation_disc} 
\begin{algorithmic}[1]
\State \textbf{Input:} 
\State number of actions $\|A\|$,
\State \texttt{diameter},
\State reward density $rd$,
\State \texttt{terminal\_state\_density}, 
\State reward delay $d$,
\State sequence length $n$,
\State transition noise $t\_n$,
\State reward noise ${\sigma}_{r\_n}$,
\State \texttt{reward\_scale},
\State \texttt{reward\_shift},
\State \texttt{term\_state\_reward},
\State \texttt{irrelevant\_features},
\State \texttt{image\_representations},
\State transforms for image representations \texttt{transforms}
\State
\Function{init\_transition\_function}{$ $}: 
    \State Set $\|NS\| = \|A\| * diameter$
    \State Divide $NS$ into independent sets $NS_i$ with $\|A\|$ elements in each with $i = 1, 2, ..., diameter$
    \For {each independent set $NS_i$}
        \For {each state $ns$ in $NS_i$} 
            \State Set possible successor states: $NS' = NS_{i+1}$
            \For {each action $a$}
                \State Set $P_{rel}(ns, a) = ns'$ sampled uniformly from $NS'$ and remove $ns'$ from $NS'$
            \EndFor
        \EndFor
    \EndFor
    \If{irrelevant\_features} 
        \State Generate dynamics $P_{irr}$ of irrelevant part of $N$-state space as was done for $P_{rel}$
    \EndIf
\EndFunction
\State
\Function{init\_reward\_function}{$ $}:
    \State Randomly sample $rd * \frac{(\|NS\| - \|T\|)!}{(\|NS\| - \|T\| - n)!}$ and store in \texttt{rewardable\_sequences}
    \State \Comment{The actual formula is more complicated because of the diameter and independent sets. The formula shown here is valid for a diameter of 1.}
    \State \Comment{Only those sequences are sampled which are admissible according to $P_{rel}$}
\EndFunction
\State

\algstore{alg1}
\end{algorithmic}
\end{algorithm}

\begin{algorithm}                     
\begin{algorithmic} 
\algrestore{alg1}

\Function{transition\_function}{$ns, a$}:
    \State $ns' = P_{rel}(ns, a)$
    \If {$\mathcal{U}(0, 1)$ < $t\_n$}
       \State $ns' = $ a random state in $NS \setminus \{P_{rel}(ns, a)\}$ \Comment{Inject noise}
    \EndIf
    \State Observation $o = ns'$
    \If{\texttt{irrelevant\_features}} 
        \State Execute dynamics $P_{irr}$ of irrelevant part of state space and append to $ns'$ to get observation $o$
    \EndIf
    \If{\texttt{image\_representations}}
        \State Set $o =$ to image of corresponding polygon(s) with applied selected \texttt{transforms}
    \EndIf
    \State \Return $o$
\EndFunction
\State
\Function{reward\_function}{$ns, a$}:
    \State $r = 0$
    \If{\texttt{irrelevant\_features}}
        \State $ns = ns[0]$ \Comment{Select first dimension of the $N$-state which is relevant to the reward}
    \EndIf
    \If {not \texttt{make\_denser}}
        \If {state sequence $nss$ of $n$ $N$-states ending $d$ steps in the past is in \texttt{rewardable\_sequences}}
            \State $r = \,$ 1
        \EndIf
    \Else
        \For {$i$ in range($n$)}
            \If {sequence of $i$ states ending $d$ steps in the past
            is a prefix sub-sequence of a sequence in \texttt{rewardable\_sequences}}
            \State $r += \,$ i/$n$
            \EndIf
        \EndFor
    \EndIf
    \If{\textit{reward every n steps} and timesteps \% n != 0}
        \State $r = 0$
    \EndIf
    \State $r\,\, += \mathcal{N}(0,\,\, {\sigma^2}_{r\_n})$
    \State $r\,\, *= \texttt{reward\_scale}$
    \State $r\,\, += \texttt{reward\_shift}$
    \If{reached terminal state}
        \State $r\,\, += \texttt{term\_state\_reward} * \texttt{reward\_scale}$
    \EndIf
    \State \Return $r$
\EndFunction
\State

\Function{main}{$ $}:
\State \textproc{init\_terminal\_states()} \Comment{Set $T$ according to \texttt{terminal\_state\_density}}
\State \textproc{init\_init\_state\_dist()} \Comment{Set $\rho_o$ to uniform distribution over non-terminal states}
\State \textproc{init\_transition\_function()}
\State \textproc{init\_reward\_function()}
\EndFunction
\end{algorithmic}
\end{algorithm}

\begin{algorithm}
\caption{Automatically Generated Continuous Toy MDPs with MDP Playground}
\label{algorithm:generation_cont}
\begin{algorithmic}[1]                   
\State \textbf{Input:} 
\State reward delay $d$,
\State transition noise ${\sigma}_{t\_n}$,
\State reward noise ${\sigma}_{r\_n}$,
\State \texttt{reward\_scale},
\State \texttt{reward\_shift},
\State \texttt{term\_state\_reward},
\State \texttt{make\_denser},
\State \texttt{relevant\_dimensions},

\State \texttt{target\_point}, 
\State \texttt{target\_radius},
\State \texttt{transition\_dynamics\_order},
\State \texttt{time\_unit},
\State \texttt{inertia}
\State
\Function{init\_transition\_function}{$ $}: 
    \State \Comment{Do nothing as continuous environments have a fixed parameterisation} 
\EndFunction

\Function{init\_reward\_function}{$n$}:
    \State \Comment{Do nothing as continuous environments have a fixed parameterisation} 
\EndFunction
\State
\Function{transition\_function}{$ns, a$}:
    \State Set $n = \texttt{transition\_dynamics\_order}$ 
    \State Set $a^n = a$ \Comment{Superscript $n$ represents $n^{th}$ derivative}
    \State Set $ns^n = a^n / inertia$ \Comment{Each state dimension is controlled by each action dimension}
    \For {i in reversed(range(n))}
        \State Set $ns^i_{t+1} = \sum\limits^{n-i}_{j = 0} ns^{i+j}_{t} \cdot \frac{1}{j!} \cdot time\_unit^{j}$ \Comment{$t$ is current time step.} 
    \EndFor
    \State $ns_{t+1}\,\, += \mathcal{N}(0, {\sigma^2}_{t\_n})$
    \State $o = ns_{t+1}$
    \If{\texttt{image\_representations}}
        \State Set left part of $o =$ to image of point mass, terminal regions and target point according to the first 2 dimensions of $o$
        \If{irrelevant features}
            \State Set right part of $o =$ to image of point mass according to the last 2 dimensions of $o$
        \EndIf
    \EndIf
    \State \Return $o$
\EndFunction

\algstore{alg2}
\end{algorithmic}
\end{algorithm}

\begin{algorithm}                     
\begin{algorithmic} 
\algrestore{alg2}
\Function{reward\_function}{$ns, a$}:
    \State $r = 0$
    \If{relevant\_dimensions}
        \State $ns = ns[\texttt{relevant\_dimensions}]$ \Comment{Select the part of state space relevant to reward}
    \EndIf
    \State $r = \textrm{Distance moved towards the \texttt{target\_point}}$
    \State $r\,\, += \mathcal{N}(0,\,\, {\sigma^2}_{r\_n})$
    \State $r\,\, *= \texttt{reward\_scale}$
    \State $r\,\, += \texttt{reward\_shift}$
    \If{reached terminal state}
        \State $r\,\, += $ Sum of delayed rewards that were not handed out so far
        \State $r\,\, += \texttt{term\_state\_reward} * \texttt{reward\_scale}$
    \EndIf
    \State \Return $r$
\EndFunction
\State

\Function{main}{$ $}:
\State \textproc{init\_terminal\_states()} \Comment{Set $T$ to be states within \textit{target\_radius} of \textit{target\_point}}
\State \textproc{init\_init\_state\_dist()} \Comment{Set $\rho_o$ to uniform distribution over non-terminal states}
\State \textproc{init\_transition\_function()}
\State \textproc{init\_reward\_function()}
\EndFunction

\end{algorithmic}
\end{algorithm}

\FloatBarrier
\section{Effect of Dimensions on More Complex Environments}\label{sec:complex_env_transfer_expts}
We include here the remaining plots and tables referenced in the discussion in the main paper that provide further detail on the experiments.

\begin{table}[ht!]
 \caption{Spearman Rank Correlations for performance on toy and complex environments across different amounts of the dimension injected: \textbf{reward delay}}
 \label{tab:delay_rank_corr}
 \begin{tabular}{||c | c | c | c||} 
 \hline
 Environment/Agent & DQN & Rainbow & A3C \\ [0.5ex] 
 \hline\hline
 \textit{beam\_rider} & r=0.8, pvalue=0.104 & r=0.4, pvalue=0.504 & r=0.6, pvalue=0.284 \\ 
 \hline
 \textit{breakout} & r=0.0, pvalue=1.0 & r=0.3, pvalue=0.623 & r=0.8, pvalue=0.104 \\
 \hline
 \textit{qbert} & r=0.6, pvalue=0.284 & r=0.4, pvalue=0.504 & r=0.9, pvalue=0.037 \\
 \hline
 \textit{space\_invaders} & r=0.9, pvalue=0.037 & r=0.9, pvalue=0.037 & r=0.8, pvalue=0.104 \\
 \hline
\end{tabular}
\end{table}

\begin{table}[ht!]
 \caption{Spearman Rank Correlations for performance on toy and complex environments across different amounts of the dimension injected: \textbf{transition noise}}
 \label{tab:p_noise_rank_corr}
 \begin{tabular}{||c | c | c | c||} 
 \hline
 Environment/Agent & DQN & Rainbow & A3C \\ [0.5ex] 
 \hline\hline
 \textit{beam\_rider} & r=0.9, pvalue=0.037 & r=0.7, pvalue=0.188 & r=0.9, pvalue=0.037 \\ 
 \hline
 \textit{breakout} & r=1.0, pvalue=1e-24 & r=1.0, pvalue=1e-24 & r=0.9, pvalue=0.037 \\
 \hline
 \textit{qbert} & r=0.9, pvalue=0.037 & r=0.9, pvalue=0.037 & r=0.7, pvalue=0.188 \\
 \hline
 \textit{space\_invaders} & r=0.9, pvalue=0.037 & r=0.7, pvalue=0.188 & r=0.9, pvalue=0.037 \\
 \hline
\end{tabular}
\end{table}

\begin{table}[ht!]
 \caption{Spearman Rank Correlations for performance on toy and complex environments across different amounts of the dimension injected: \textbf{reward noise}}
 \label{tab:r_noise_rank_corr}
 \begin{tabular}{||c | c | c | c||} 
 \hline
 Environment/Agent & DQN & Rainbow & A3C \\ [0.5ex] 
 \hline\hline
 \textit{beam\_rider} & r=0.9, pvalue=0.037 & r=0.9, pvalue=0.037 & r=0.7, pvalue=0.188 \\ 
 \hline
 \textit{breakout} & r=0.8, pvalue=0.104 & r=0.9, pvalue=0.037 & r=0.9, pvalue=0.037 \\
 \hline
 \textit{qbert} & r=0.4, pvalue=0.504 & r=0.5, pvalue=0.391 & r=0.5, pvalue=0.391 \\
 \hline
 \textit{space\_invaders} & r=0.9, pvalue=0.037 & r=0.7, pvalue=0.188 & r=0.7, pvalue=0.188 \\
 \hline
\end{tabular}
\end{table}

\begin{figure}[ht]
        \vskip 0.5cm
        \centering
        \begin{subfigure}[]{0.325\textwidth}
            \centering
            \includegraphics[width=\textwidth]{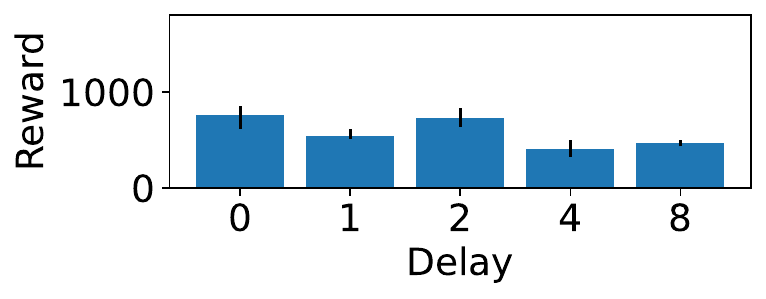}
            \caption[]%
            {{\small DQN }}    
            \label{fig:dqn_beam_rider_del}
        \end{subfigure}
        \begin{subfigure}[]{0.325\textwidth}
            \centering
            \includegraphics[width=\textwidth]{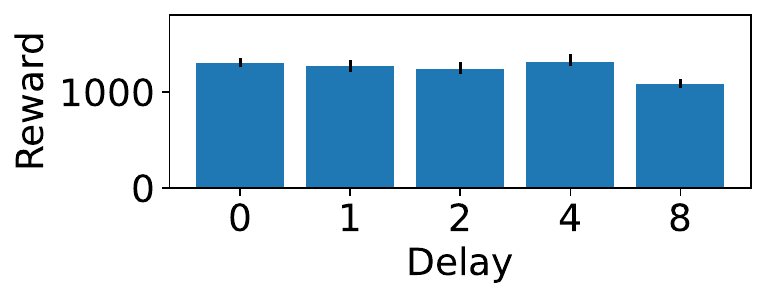}
            \caption[]%
            {{\footnotesize Rainbow }}    
            \label{fig:rainbow_beam_rider_del}
        \end{subfigure}        
        \begin{subfigure}[]{0.325\textwidth}
            \centering
            \includegraphics[width=\textwidth]{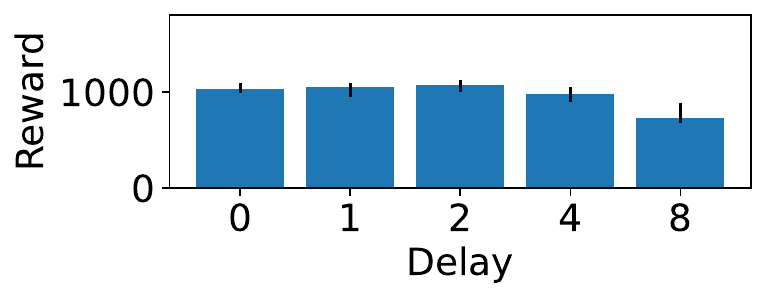}
            \caption[]%
            {{\small A3C }}    
            \label{fig:a3c_beam_rider_del}
        \end{subfigure}
        
        \begin{subfigure}[]{0.325\textwidth}
            \centering
            \includegraphics[width=\textwidth]{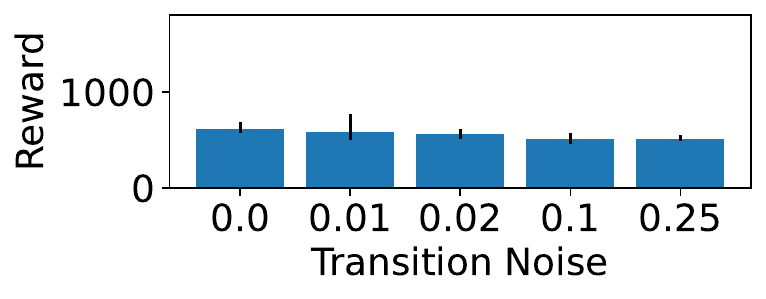}
            \caption[]%
            {{\small DQN }}    
            \label{fig:dqn_beam_rider_p_noise}
        \end{subfigure}
        \begin{subfigure}[]{0.325\textwidth}
            \centering
            \includegraphics[width=\textwidth]{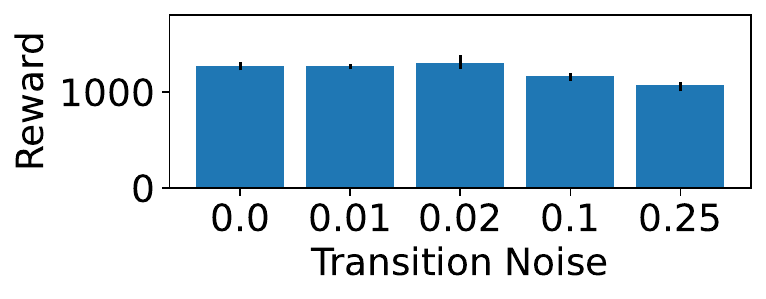}
            \caption[]%
            {{\small Rainbow }}    
            \label{fig:rainbow_beam_rider_p_noise}
        \end{subfigure}
        \begin{subfigure}[]{0.325\textwidth}
            \centering
            \includegraphics[width=\textwidth]{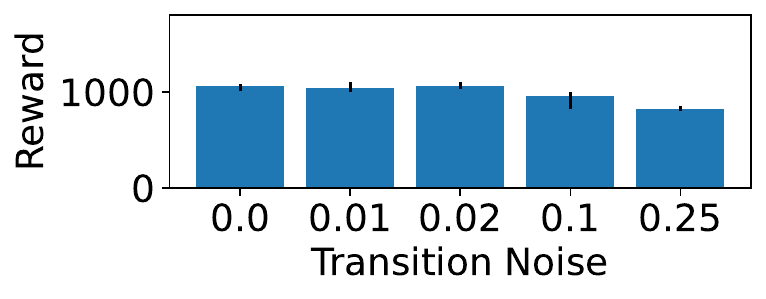}
            \caption[]%
            {{\small A3C }}    
            \label{fig:a3c_beam_rider_p_noise}
        \end{subfigure}

        \begin{subfigure}[]{0.325\textwidth}
            \centering
            \includegraphics[width=\textwidth]{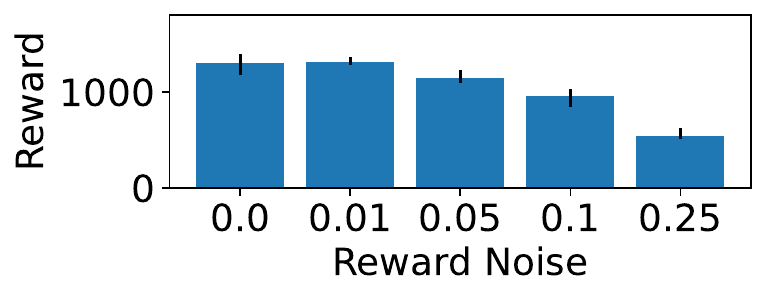}
            \caption[]%
            {{\small DQN }}    
            \label{fig:dqn_beam_rider_r_noise}
        \end{subfigure}
        \begin{subfigure}[]{0.325\textwidth}
            \centering
            \includegraphics[width=\textwidth]{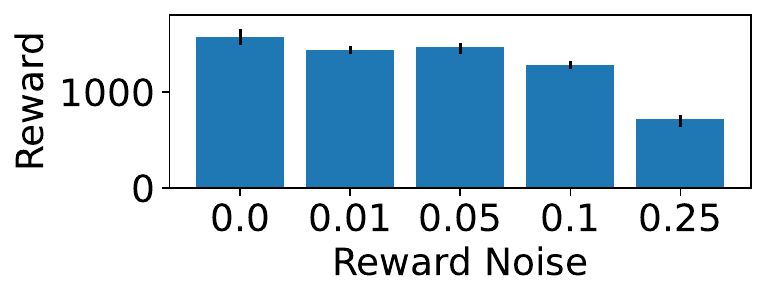}
            \caption[]%
            {{\small Rainbow }}    
            \label{fig:rainbow_beam_rider_r_noise}
        \end{subfigure}
        \begin{subfigure}[]{0.325\textwidth}
            \centering
            \includegraphics[width=\textwidth]{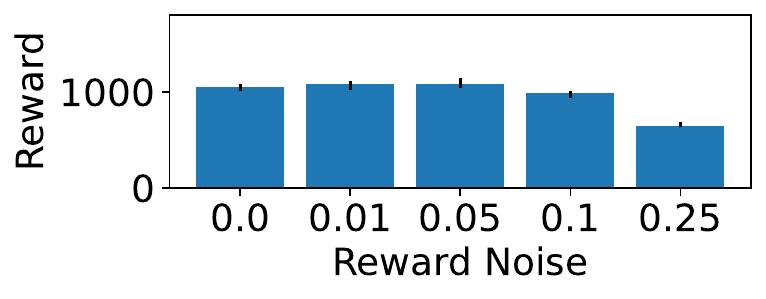}
            \caption[]%
            {{\small A3C }}    
            \label{fig:a3c_beam_rider_r_noise}
        \end{subfigure}

        \caption[]
        {AUC of episodic reward for DQN, Rainbow, A3C at the end of training on \textit{Beam Rider} for various dimensions of hardness. Error bars represent bootstrapped confidence intervals with Bonferroni corrections for a significance level of 0.05.}
        \label{fig:beam_rider_perfs}
\end{figure}

\begin{figure}[ht]
        \centering
        \begin{subfigure}[]{0.325\textwidth}
            \centering
            \includegraphics[width=\textwidth]{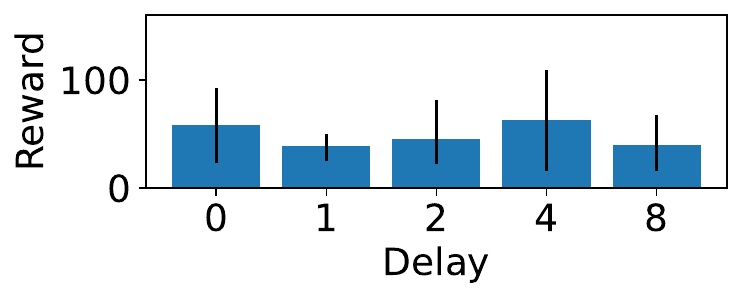}
            \caption[]%
            {{\small DQN}}    
            \label{fig:dqn_breakout_del}
        \end{subfigure}
        \begin{subfigure}[]{0.325\textwidth}
            \centering
            \includegraphics[width=\textwidth]{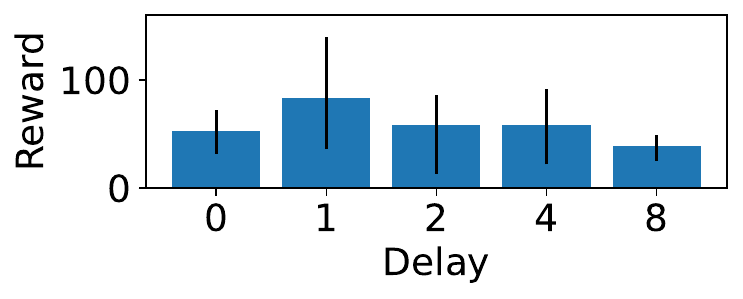}
            \caption[]%
            {{\small Rainbow}}    
            \label{fig:rainbow_breakout_del}
        \end{subfigure}
        \begin{subfigure}[]{0.325\textwidth}
            \centering
            \includegraphics[width=\textwidth]{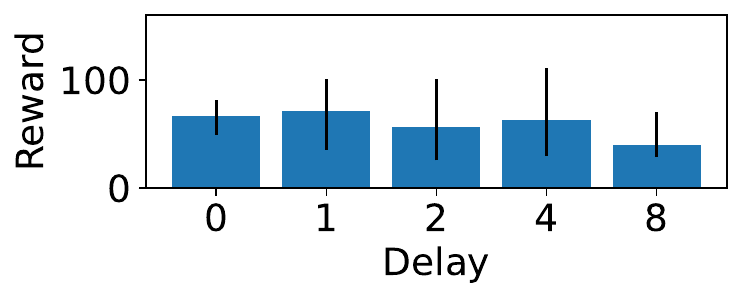}
            \caption[]%
            {{\small A3C}}    
            \label{fig:a3c_breakout_del}
        \end{subfigure}
        
        \begin{subfigure}[]{0.325\textwidth}
            \centering
            \includegraphics[width=\textwidth]{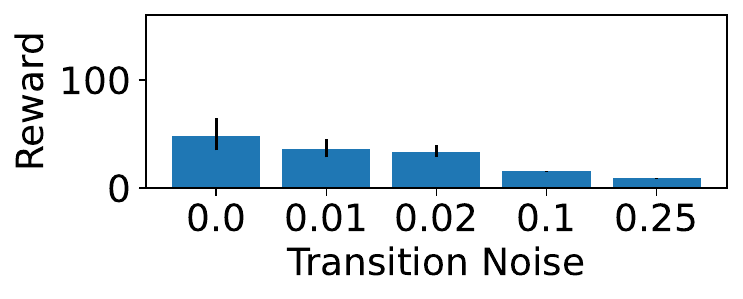}
            \caption[]%
            {{\small DQN}}
            \label{fig:append_dqn_breakout_p_noise}
        \end{subfigure}
        \begin{subfigure}[]{0.325\textwidth}
            \centering
            \includegraphics[width=\textwidth]{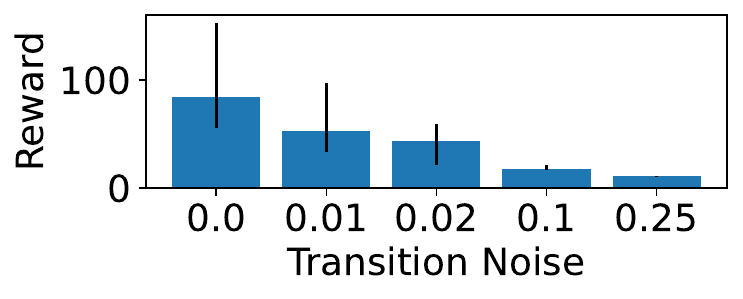}
            \caption[]%
            {{\small Rainbow}}    
            \label{fig:rainbow_breakout_p_noise append_fig}
        \end{subfigure}
        \begin{subfigure}[]{0.325\textwidth}
            \centering
            \includegraphics[width=\textwidth]{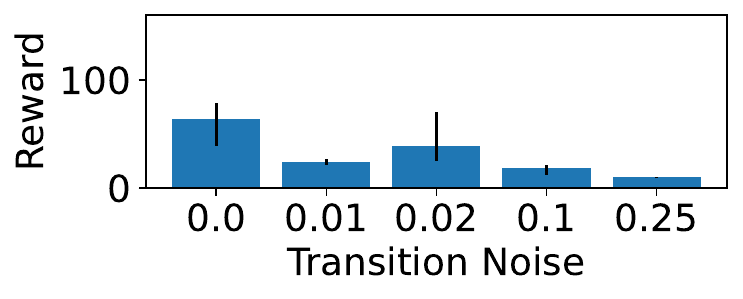}
            \caption[]%
            {{\small A3C}}    
            \label{fig:a3c_breakout_p_noise}
        \end{subfigure}
        
        \begin{subfigure}[]{0.325\textwidth}
            \centering
            \includegraphics[width=\textwidth]{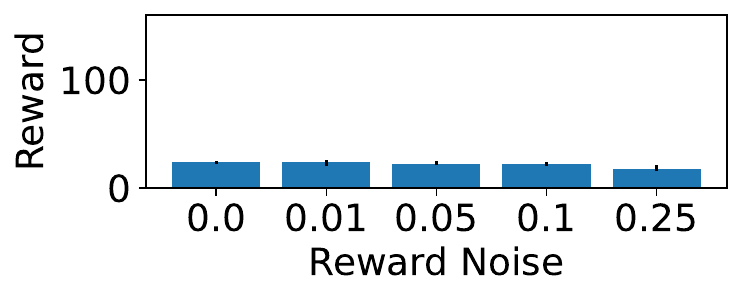}
            \caption[]%
            {{\small DQN}}
            \label{fig:append_dqn_breakout_r_noise}
        \end{subfigure}
        \begin{subfigure}[]{0.325\textwidth}
            \centering
            \includegraphics[width=\textwidth]{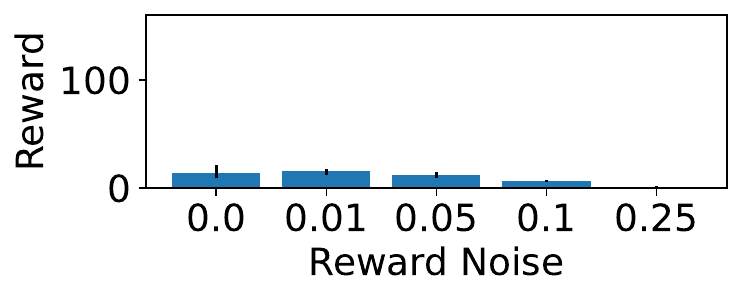}
            \caption[]%
            {{\small Rainbow}}    
            \label{fig:rainbow_breakout_r_noise}
        \end{subfigure}
        \begin{subfigure}[]{0.325\textwidth}
            \centering
            \includegraphics[width=\textwidth]{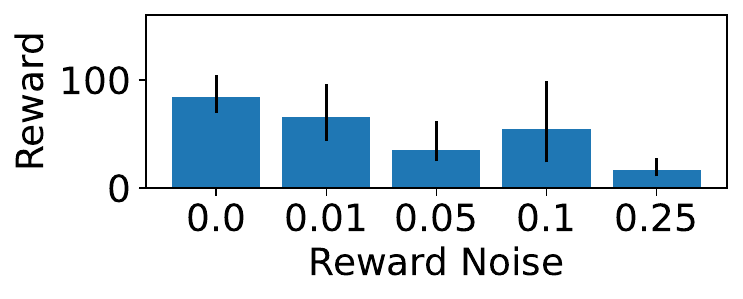}
            \caption[]%
            {{\small A3C}}    
            \label{fig:a3c_breakout_r_noise}
        \end{subfigure}

        \caption[]
        {AUC of episodic reward for DQN, Rainbow, A3C at the end of training on \textit{Breakout} for various dimensions of hardness. Error bars represent bootstrapped confidence intervals with Bonferroni corrections for a significance level of 0.05.}
        \label{fig:breakout_perfs}
\end{figure}

\begin{figure}[ht]
        \centering
        
        \begin{subfigure}[]{0.325\textwidth}
            \centering
            \includegraphics[width=\textwidth]{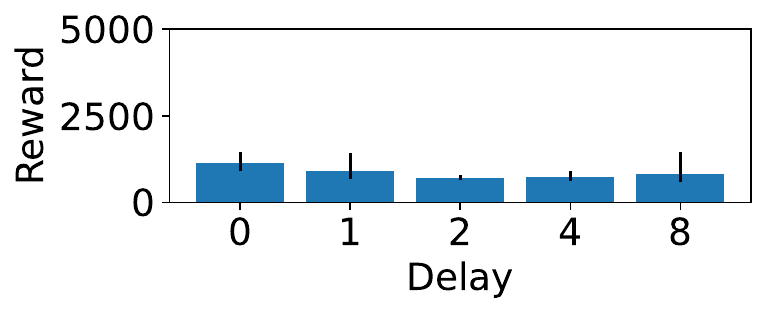}
            \caption[]%
            {{\small DQN}}
            \label{fig:dqn_qbert_del append_fig}
        \end{subfigure}
        \begin{subfigure}[]{0.325\textwidth}
            \centering
            \includegraphics[width=\textwidth]{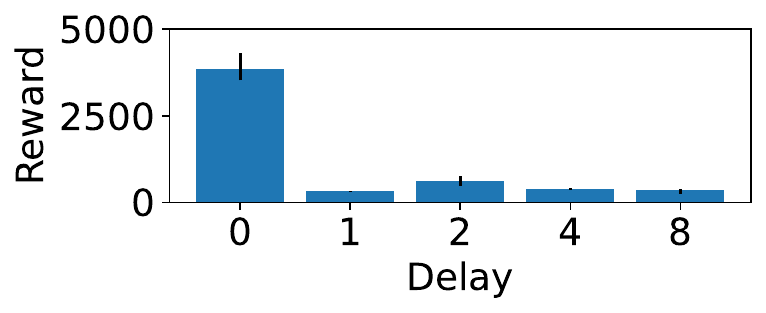}
            \caption[]%
            {{\small Rainbow}}
            \label{fig:rainbow_qbert_del}
        \end{subfigure}
        \begin{subfigure}[]{0.325\textwidth}
            \centering
            \includegraphics[width=\textwidth]{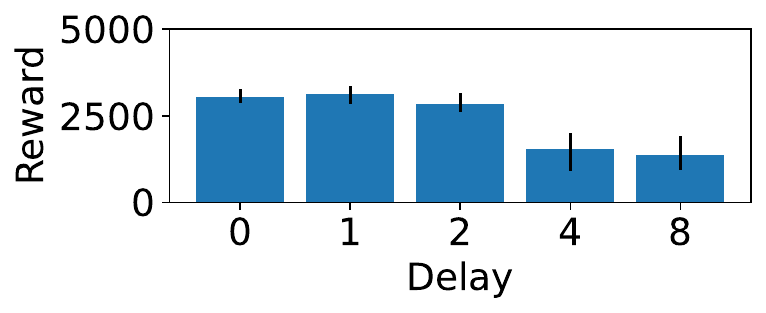}
            \caption[]%
            {{\small A3C}}
            \label{fig:a3c_qbert_del}
        \end{subfigure}
        
        \begin{subfigure}[]{0.325\textwidth}
            \centering
            \includegraphics[width=\textwidth]{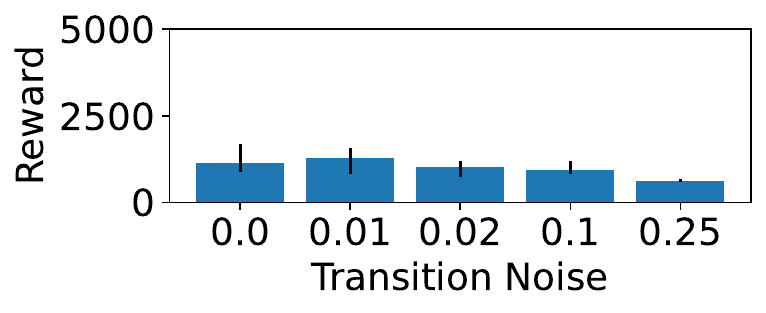}
            \caption[]%
            {{\footnotesize DQN}}
            \label{fig:dqn_qbert_p_noise}
        \end{subfigure}
        \begin{subfigure}[]{0.325\textwidth}
            \centering
            \includegraphics[width=\textwidth]{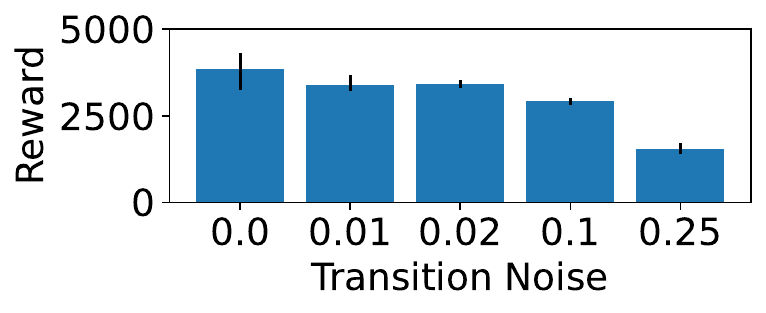}
            \caption[]%
            {{\small Rainbow}}
            \label{fig:rainbow_qbert_p_noise}
        \end{subfigure}
        \begin{subfigure}[]{0.325\textwidth}
            \centering
            \includegraphics[width=\textwidth]{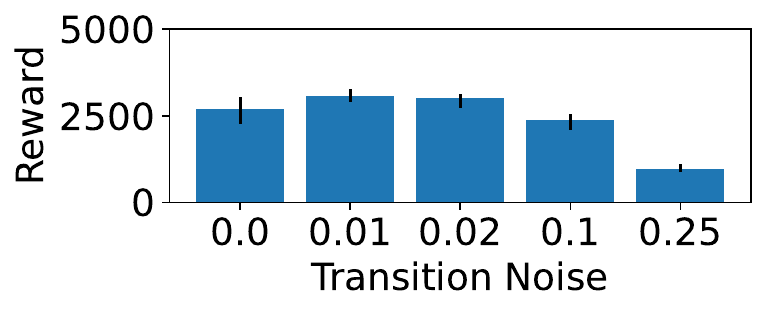}
            \caption[]%
            {{\small A3C}}
            \label{fig:a3c_qbert_p_noise}
        \end{subfigure}

        \begin{subfigure}[]{0.325\textwidth}
            \centering
            \includegraphics[width=\textwidth]{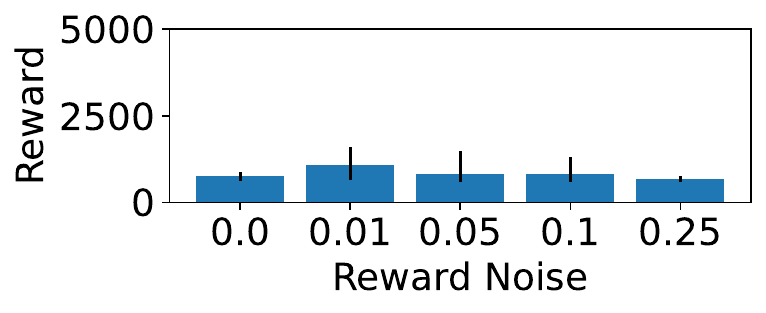}
            \caption[]%
            {{\small DQN}}
            \label{fig:dqn_qbert_r_noise}
        \end{subfigure}
        \begin{subfigure}[]{0.325\textwidth}
            \centering
            \includegraphics[width=\textwidth]{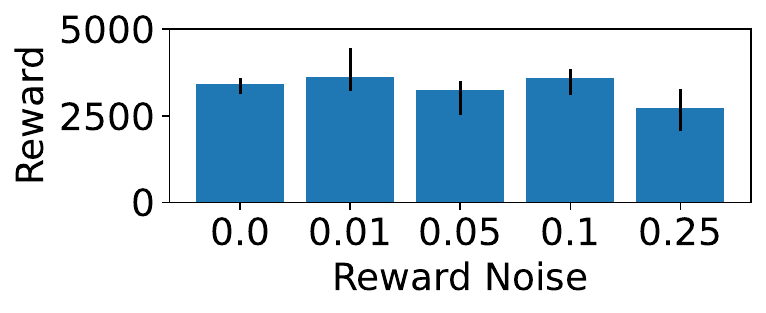}
            \caption[]%
            {{\small Rainbow}}
            \label{fig:rainbow_qbert_r_noise}
        \end{subfigure}
        \begin{subfigure}[]{0.325\textwidth}
            \centering
            \includegraphics[width=\textwidth]{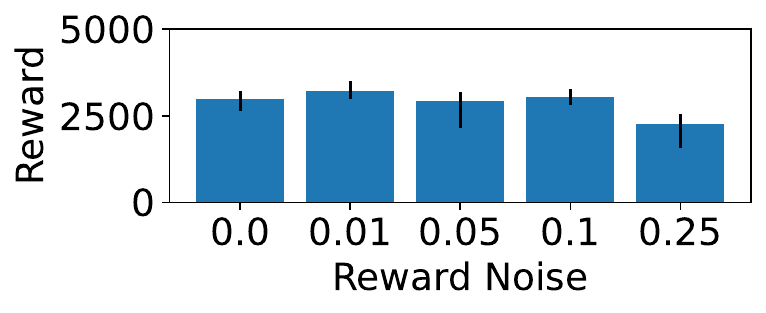}
            \caption[]%
            {{\small A3C}}
            \label{fig:a3c_qbert_r_noise}
        \end{subfigure}        

        \caption[]
        {AUC of episodic reward for DQN, Rainbow, A3C at the end of training on \textit{Qbert} for various dimensions of hardness. Error bars represent bootstrapped confidence intervals with Bonferroni corrections for a significance level of 0.05.}
        \label{fig:qbert_perfs}
\end{figure}

\begin{figure}[ht]
        \centering

        \begin{subfigure}[]{0.325\textwidth}
            \centering
            \includegraphics[width=\textwidth]{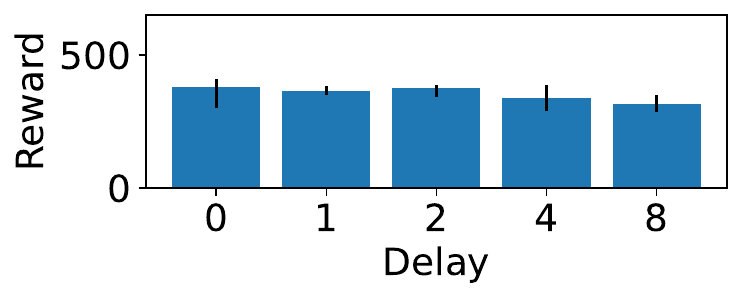}
            \caption[]%
            {{\small DQN}}    
            \label{fig:dqn_space_invaders_del}
        \end{subfigure}
        \begin{subfigure}[]{0.325\textwidth}
            \centering
            \includegraphics[width=\textwidth]{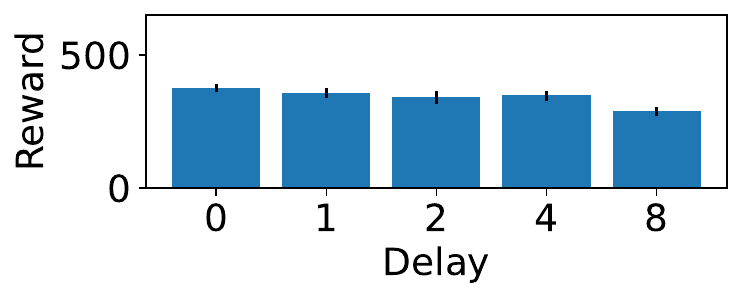}
            \caption[]%
            {{\footnotesize Rainbow}}    
            \label{fig:rainbow_space_invaders_del}
        \end{subfigure}
        \begin{subfigure}[]{0.325\textwidth}
            \centering
            \includegraphics[width=\textwidth]{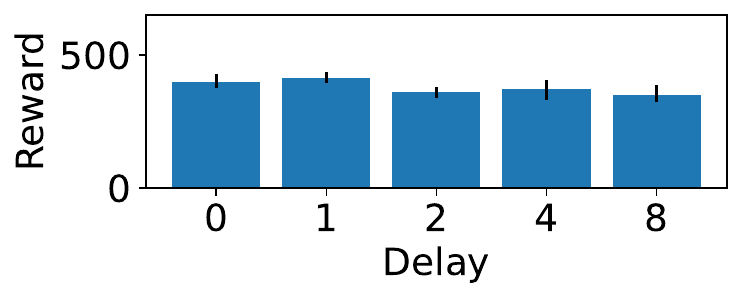}
            \caption[]%
            {{\small A3C}}    
            \label{fig:a3c_space_invaders_del}
        \end{subfigure}

        \begin{subfigure}[]{0.325\textwidth}
            \centering
            \includegraphics[width=\textwidth]{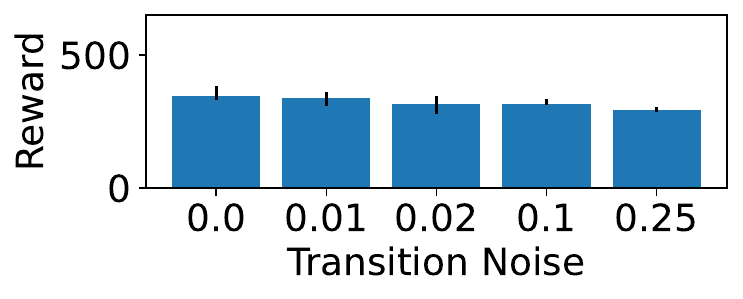}
            \caption[]%
            {{\footnotesize DQN}}    
            \label{fig:dqn_space_invaders_p_noise}
        \end{subfigure}
        \begin{subfigure}[]{0.325\textwidth}
            \centering
            \includegraphics[width=\textwidth]{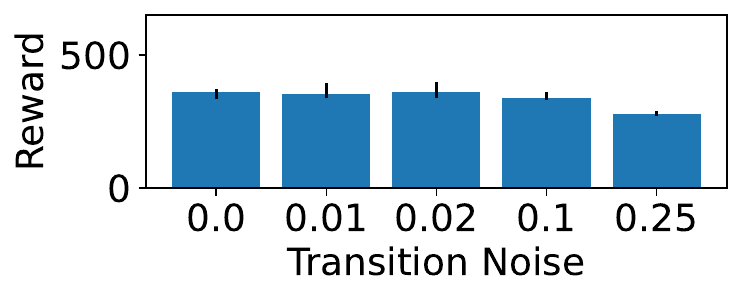}
            \caption[]%
            {{\footnotesize Rainbow}}    
            \label{fig:rainbow_space_invaders_p_noise}
        \end{subfigure}
        \begin{subfigure}[]{0.325\textwidth}
            \centering
            \includegraphics[width=\textwidth]{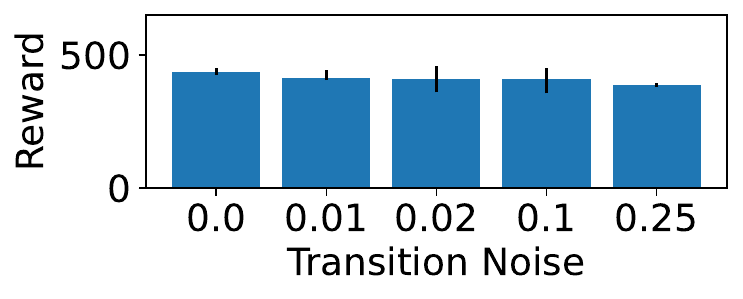}
            \caption[]%
            {{\small A3C}}    
            \label{fig:a3c_space_invaders_p_noise}
        \end{subfigure}

        \begin{subfigure}[]{0.325\textwidth}
            \centering
            \includegraphics[width=\textwidth]{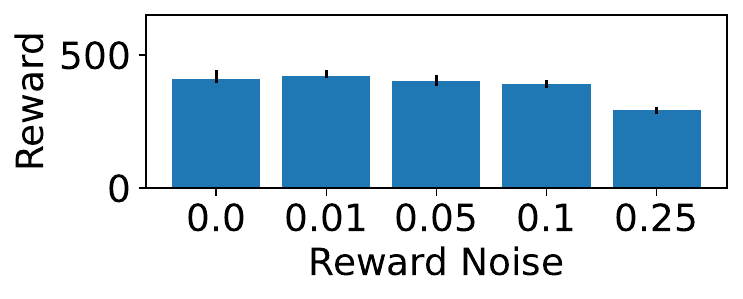}
            \caption[]%
            {{\small DQN}}    
            \label{fig:dqn_space_invaders_r_noise}
        \end{subfigure}
        \begin{subfigure}[]{0.325\textwidth}
            \centering
            \includegraphics[width=\textwidth]{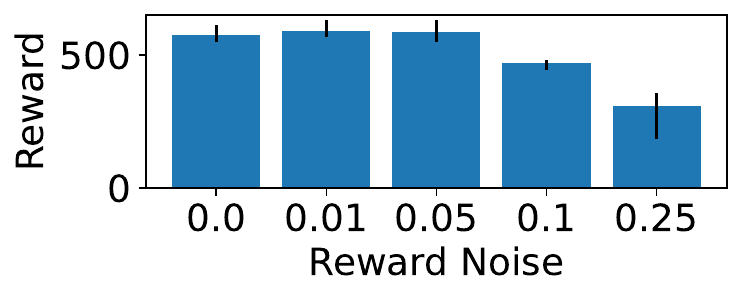}
            \caption[]%
            {{\footnotesize Rainbow}}    
            \label{fig:rainbow_space_invaders_r_noise}
        \end{subfigure}
        \begin{subfigure}[]{0.325\textwidth}
            \centering
            \includegraphics[width=\textwidth]{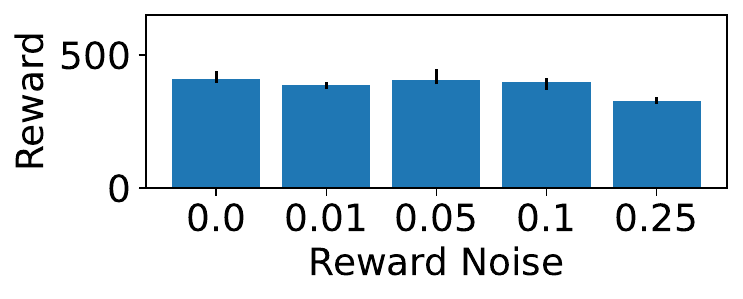}
            \caption[]%
            {{\small A3C}}    
            \label{fig:a3c_space_invaders_r_noise}
        \end{subfigure}

        \caption[]
        {AUC of episodic reward for DQN, Rainbow, A3C at the end of training on \textit{Space Invaders} for various dimensions of hardness. Error bars represent bootstrapped confidence intervals with Bonferroni corrections for a significance level of 0.05.}
        \label{fig:space_invaders_perfs}
\end{figure}

\begin{figure}[ht]
        \centering
        \begin{subfigure}[]{0.325\textwidth}
            \centering
            \includegraphics[width=\textwidth]{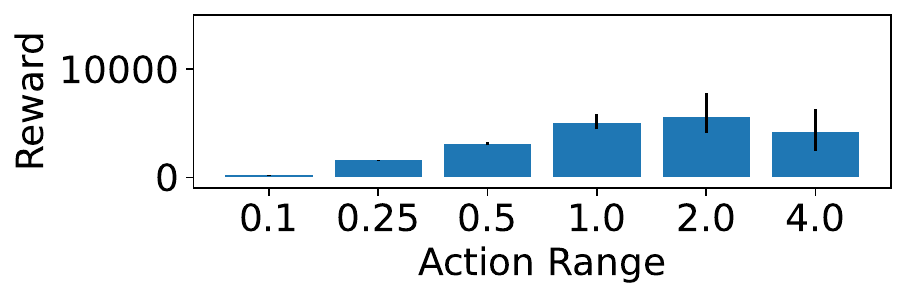}
            \caption[DDPG]%
            {{\small DDPG action range}}    
            \label{fig:append_ddpg_act_max_halfcheetah}
        \end{subfigure}
        \begin{subfigure}[]{0.325\textwidth}
            \centering
            \includegraphics[width=\textwidth]{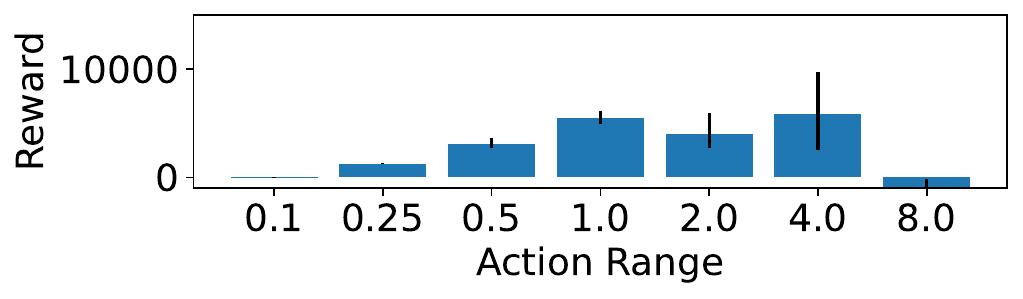}
            \caption[]%
            {{\small TD3 action range}}    
            \label{fig:td3_act_max_halfcheetah}
        \end{subfigure}
        \begin{subfigure}[]{0.325\textwidth}
            \centering
            \includegraphics[width=\textwidth]{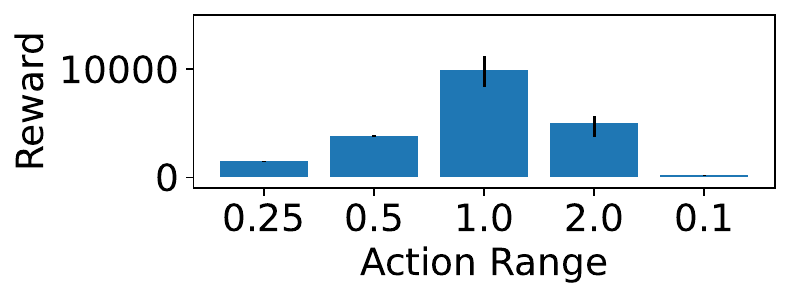}
            \caption[]%
            {{\small SAC action range}}    
            \label{fig:sac_act_max_halfcheetah}
        \end{subfigure}
        
        \begin{subfigure}[]{0.325\textwidth}   
            \centering 
            \includegraphics[width=\textwidth]{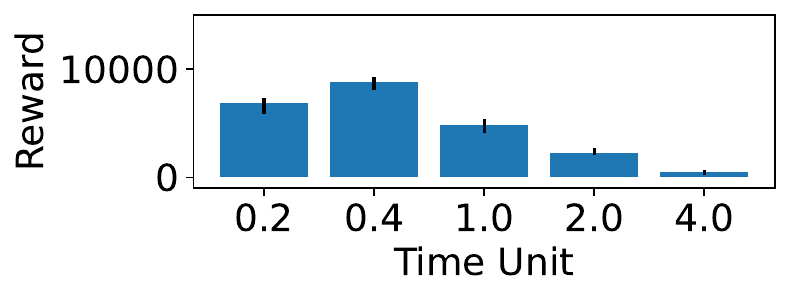} 
            \caption[DDPG]%
            {{\small DDPG time unit}}
            \label{fig:append_ddpg_time_unit_halfcheetah}
        \end{subfigure}
        \begin{subfigure}[]{0.325\textwidth}   
            \centering 
            \includegraphics[width=\textwidth]{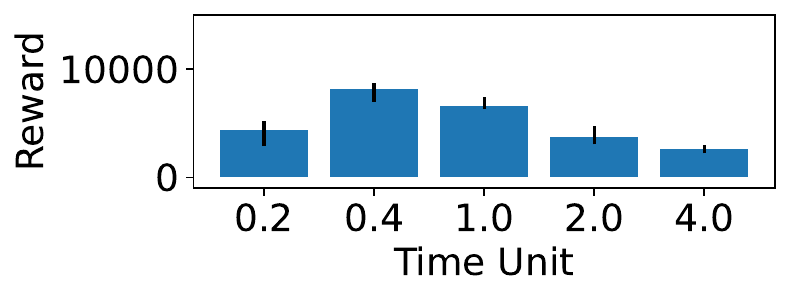}
            \caption[]%
            {{\small TD3 time unit}}    
            \label{fig:td3_time_unit_halfcheetah}
        \end{subfigure}
        \begin{subfigure}[]{0.325\textwidth}   
            \centering 
            \includegraphics[width=\textwidth]{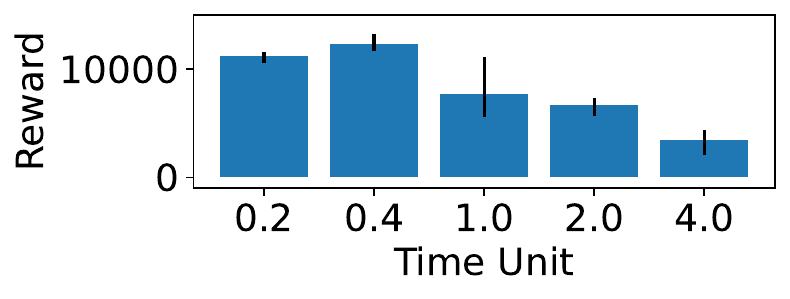}
            \caption[]%
            {{\small SAC time unit}}    
            \label{fig:sac_time_unit_halfcheetah}
        \end{subfigure}
    
        \caption[]
        {AUC of episodic reward for DDPG, TD3, SAC at the end of training on HalfCheetah when varying \textbf{action range} and \textbf{time unit}. Error bars represent bootstrapped confidence intervals with Bonferroni corrections for a significance level of 0.05. For SAC and DDPG, runs for \textit{action range} values $>=2$ and $>=4$ crashed and are absent from the plot.}
        \label{fig:algs_halfcheetah_perfs}
\end{figure}

\begin{figure}[ht]
        \centering
        \begin{subfigure}[]{0.325\textwidth}
            \centering
            \includegraphics[width=\textwidth]{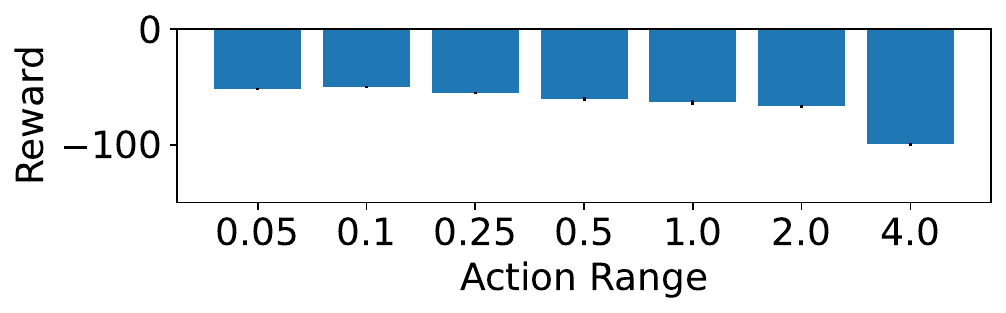}
            \caption[DDPG]%
            {{\small DDPG Pusher}}    
            \label{fig:ddpg_act_max_pusher}
        \end{subfigure}
        \begin{subfigure}[]{0.325\textwidth}
            \centering
            \includegraphics[width=\textwidth]{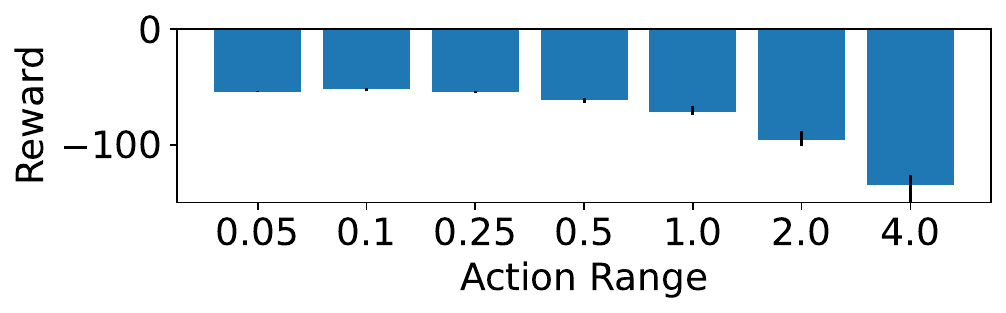}
            \caption[]%
            {{\small TD3 Pusher}}    
            \label{fig:td3_act_max_pusher}
        \end{subfigure}
        \begin{subfigure}[]{0.325\textwidth}
            \centering
            \includegraphics[width=\textwidth]{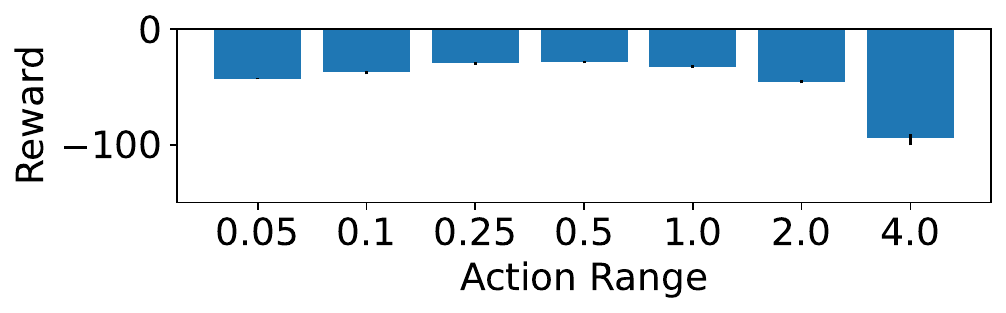}
            \caption[]%
            {{\small SAC Pusher}}    
            \label{fig:sac_act_max_pusher}
        \end{subfigure}

        \begin{subfigure}[]{0.325\textwidth}   
            \centering 
            \includegraphics[width=\textwidth]{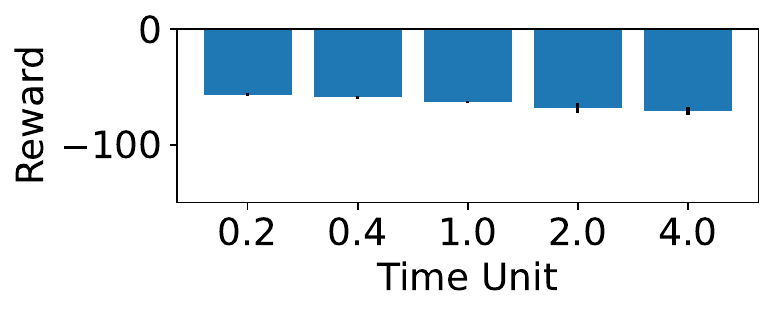}
            \caption[DDPG]%
            {{\small DDPG Pusher}}
            \label{fig:ddpg_time_unit_pusher}
        \end{subfigure}
        \begin{subfigure}[]{0.325\textwidth}   
            \centering 
            \includegraphics[width=\textwidth]{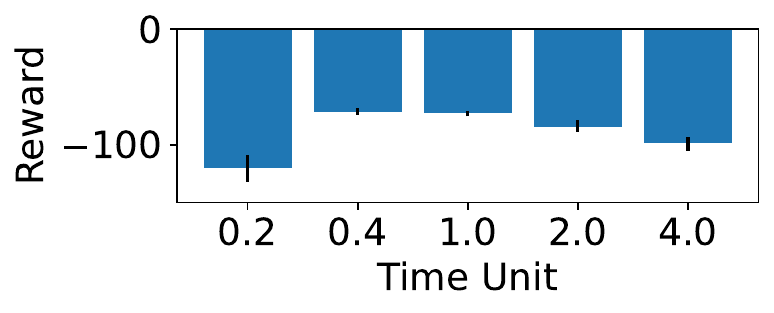}
            \caption[]%
            {{\small TD3 Pusher}}    
            \label{fig:td3_time_unit_pusher}
        \end{subfigure}
        \begin{subfigure}[]{0.325\textwidth}   
            \centering 
            \includegraphics[width=\textwidth]{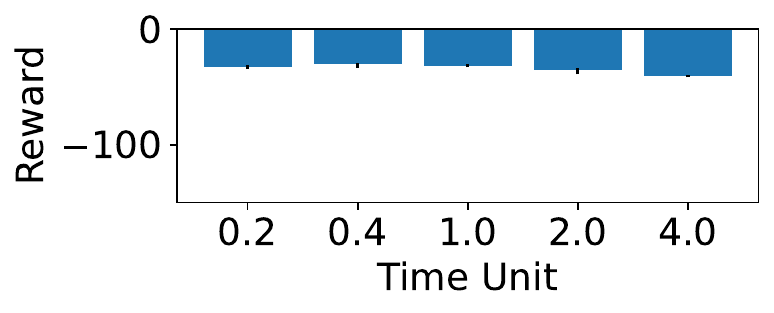}
            \caption[]%
            {{\small SAC Pusher}}    
            \label{fig:sac_time_unit_pusher}
        \end{subfigure}

        \caption[]
        {AUC of episodic reward for DDPG, TD3, SAC at the end of training on \textit{Pusher} when varying \textbf{action range} and \textbf{time unit}. Error bars represent bootstrapped confidence intervals with Bonferroni corrections for a significance level of 0.05.}
        \label{fig:algs_pusher_perfs}
\end{figure}

\begin{figure}
        \centering

        \begin{subfigure}[]{0.325\textwidth}
            \centering
            \includegraphics[width=\textwidth]{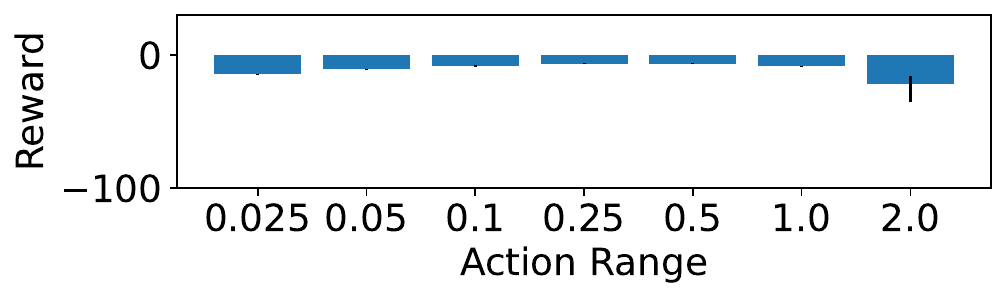}
            \caption[DDPG]%
            {{\small DDPG Reacher}}    
            \label{fig:ddpg_act_max_reacher}
        \end{subfigure}
        \begin{subfigure}[]{0.325\textwidth}
            \centering
            \includegraphics[width=\textwidth]{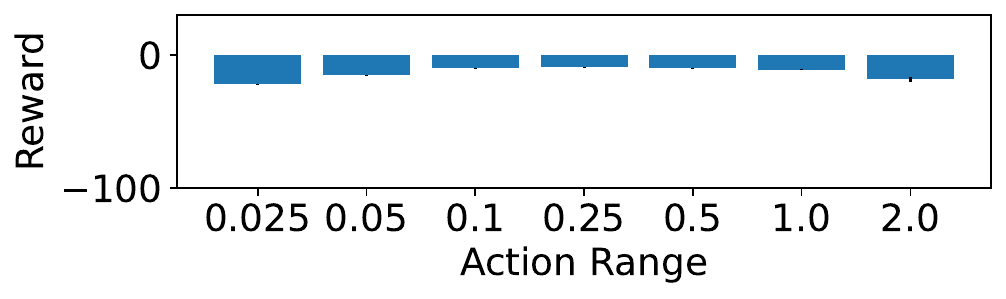}
            \caption[]%
            {{\small TD3 Reacher}}    
            \label{fig:td3_act_max_reacher}
        \end{subfigure}
        \begin{subfigure}[]{0.325\textwidth}
            \centering
            \includegraphics[width=\textwidth]{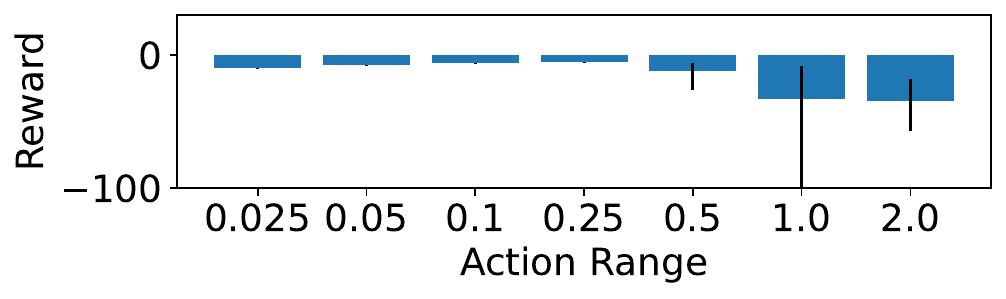}
            \caption[]%
            {{\small SAC Reacher}}    
            \label{fig:sac_act_max_reacher}
        \end{subfigure}

        \begin{subfigure}[]{0.325\textwidth}   
            \centering 
            \includegraphics[width=\textwidth]{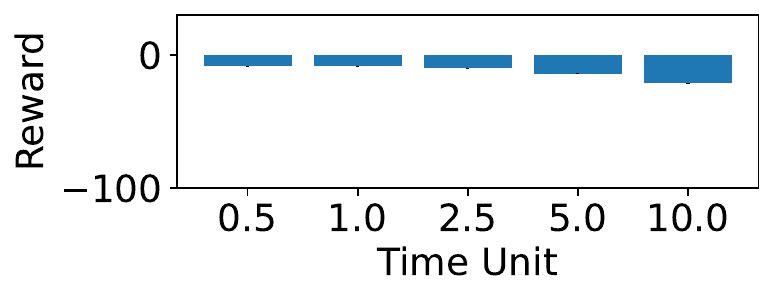}
            \caption[DDPG]%
            {{\small DDPG Reacher}}
            \label{fig:ddpg_time_unit_reacher}
        \end{subfigure}
        \begin{subfigure}[]{0.325\textwidth}   
            \centering 
            \includegraphics[width=\textwidth]{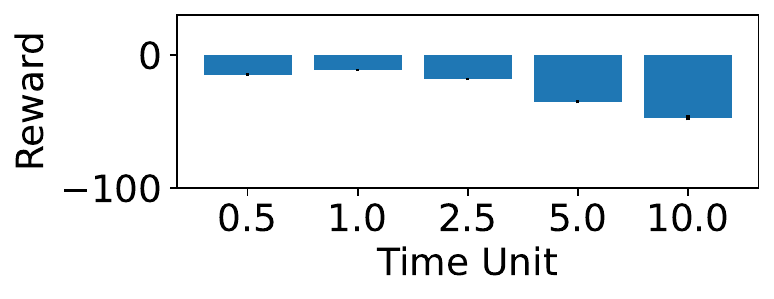}
            \caption[]%
            {{\small TD3 Reacher}}    
            \label{fig:td3_time_unit_reacher}
        \end{subfigure}
        \begin{subfigure}[]{0.325\textwidth}   
            \centering 
            \includegraphics[width=\textwidth]{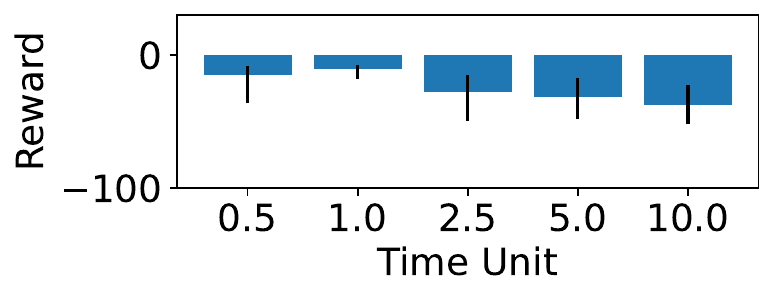}
            \caption[]%
            {{\small SAC Reacher}}    
            \label{fig:sac_time_unit_reacher}
        \end{subfigure}

        \caption[]
        {AUC of episodic reward for DDPG, TD3, SAC at the end of training on \textit{Reacher} when varying \textbf{action range} and \textbf{time unit}. Error bars represent bootstrapped confidence intervals with Bonferroni corrections for a significance level of 0.05.}
        \label{fig:algs_reacher_perfs}
\end{figure}

\FloatBarrier
\section{Plots for Tabular Baselines}\label{sec:tabular_baselines}
We present here plots of experiments with the tabular baselines on the discrete toy environments with image representations turned off. 

\begin{figure*}[ht]
        \centering
        \begin{subfigure}[]{0.32\textwidth}
            \centering
            \includegraphics[width=\textwidth]{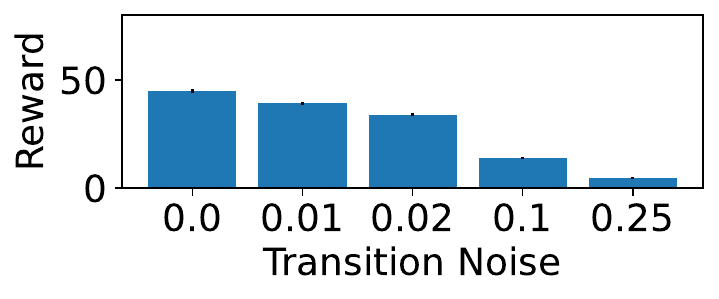}
            \caption[Q-Learning]%
            {{\small Q-Learning}}    
            \label{fig:q_learn_tabular_p_noise_train a}
        \end{subfigure}
        \begin{subfigure}[]{0.32\textwidth}   
            \centering 
            \includegraphics[width=\textwidth]{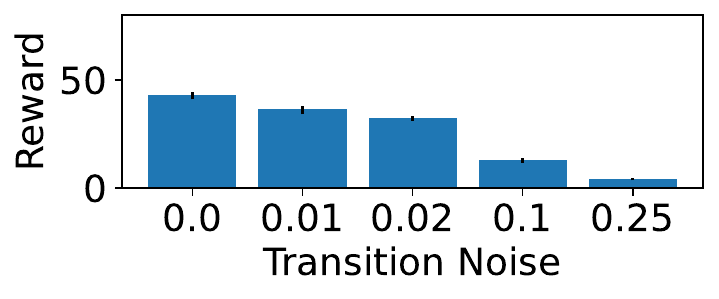}
            \caption[SARSA]%
            {{\small SARSA}}    
            \label{fig:sarsa_tabular_p_noise_train b}
        \end{subfigure}
        \begin{subfigure}[]{0.32\textwidth}
            \centering
            \includegraphics[width=\textwidth]{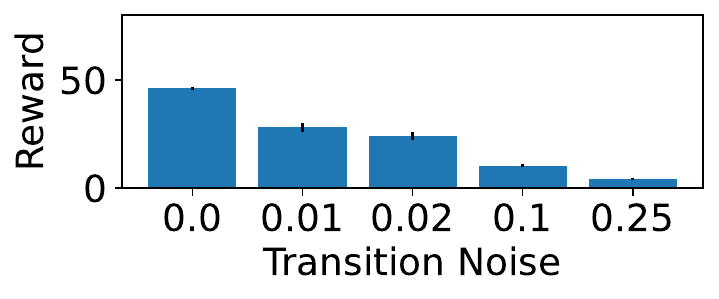}
            \caption[Double Q-Learning]%
            {{\small Double Q-Learning}}    
            \label{fig:double_q_learn_tabular_p_noise_train c}
        \end{subfigure}

        \caption[ AUC of episodic reward at the end of training ]
        {AUC of episodic reward at the end of training for three different tabular baseline algorithms when varying \textbf{transition noise}. Error bars represent bootstrapped confidence intervals with Bonferroni corrections for a significance level of 0.05.}
        \label{fig:tabular_p_noise}
\end{figure*}

\begin{figure*}[ht]
        \centering
        \begin{subfigure}[]{0.32\textwidth}
            \centering
            \includegraphics[width=\textwidth]{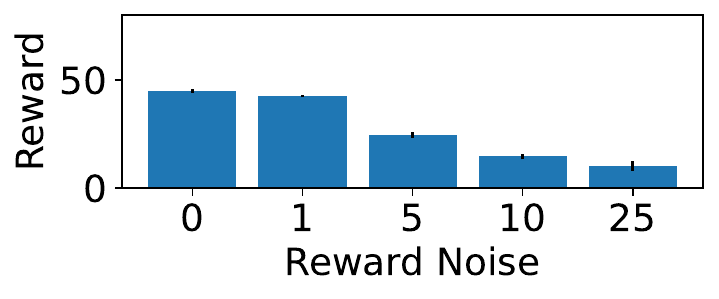}
            \caption[Q-Learning]%
            {{\small Q-Learning}}    
            \label{fig:q_learn_tabular_r_noise_train a}
        \end{subfigure}
        \begin{subfigure}[]{0.32\textwidth}   
            \centering 
            \includegraphics[width=\textwidth]{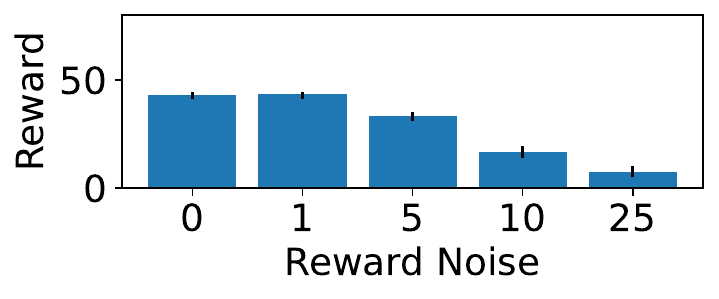}
            \caption[SARSA]%
            {{\small SARSA}}    
            \label{fig:sarsa_tabular_r_noise_train b}
        \end{subfigure}
        \begin{subfigure}[]{0.32\textwidth}
            \centering
            \includegraphics[width=\textwidth]{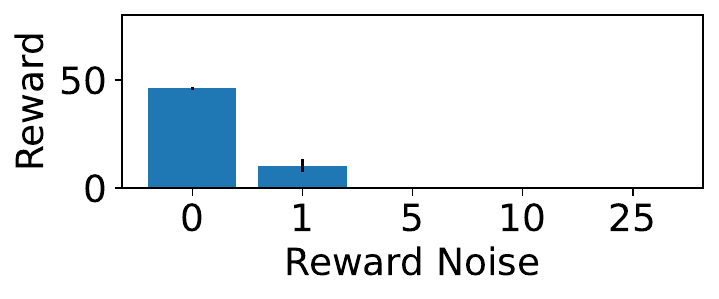}
            \caption[Double Q-Learning]%
            {{\small Double Q-Learning}}    
            \label{fig:double_q_learn_tabular_r_noise_train c}
        \end{subfigure}

        \caption[ AUC of episodic reward at the end of training ]
        {AUC of episodic reward at the end of training for three different tabular baseline algorithms when varying \textbf{reward noise}. Error bars represent bootstrapped confidence intervals with Bonferroni corrections for a significance level of 0.05.}
        \label{fig:tabular_rew_noise}
\end{figure*}

\begin{figure*}[ht]
        \centering
        \begin{subfigure}[]{0.32\textwidth}
            \centering
            \includegraphics[width=\textwidth]{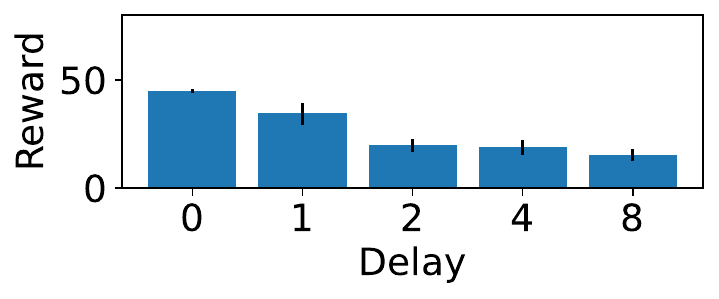}
            \caption[Q-Learning]%
            {{\small Q-Learning}}    
            \label{fig:q_learn_tabular_del_train a}
        \end{subfigure}
        \begin{subfigure}[]{0.32\textwidth}   
            \centering 
            \includegraphics[width=\textwidth]{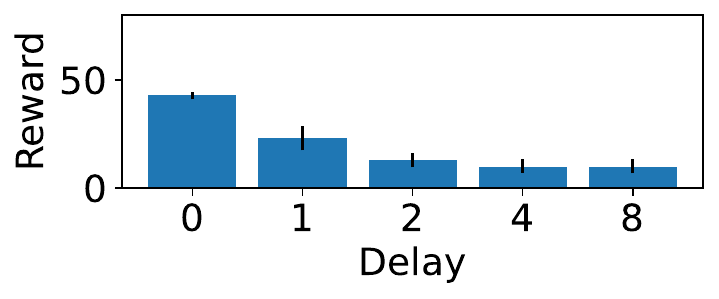}
            \caption[SARSA]%
            {{\small SARSA}}    
            \label{fig:sarsa_tabular_del_train b}
        \end{subfigure}
        \begin{subfigure}[]{0.32\textwidth}
            \centering
            \includegraphics[width=\textwidth]{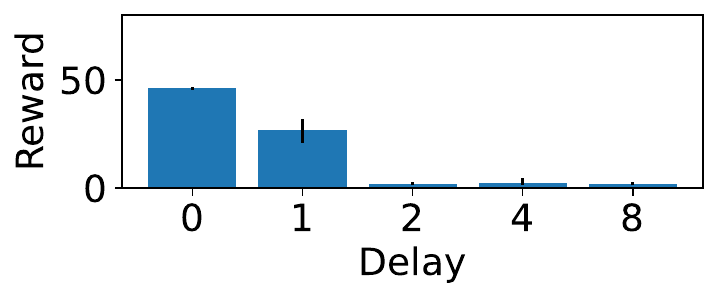}
            \caption[Double Q-Learning]%
            {{\small Double Q-Learning}}    
            \label{fig:double_q_learn_tabular_del_train c}
        \end{subfigure}

        \caption[ AUC of episodic reward at the end of training ]
        {AUC of episodic reward at the end of training for three different tabular baseline algorithms when varying \textbf{reward delay}. Error bars represent bootstrapped confidence intervals with Bonferroni corrections for a significance level of 0.05.}
        \label{fig:tabular_delay}
\end{figure*}

\begin{figure*}[ht]
        \centering
        \begin{subfigure}[]{0.32\textwidth}
            \centering
            \includegraphics[width=\textwidth]{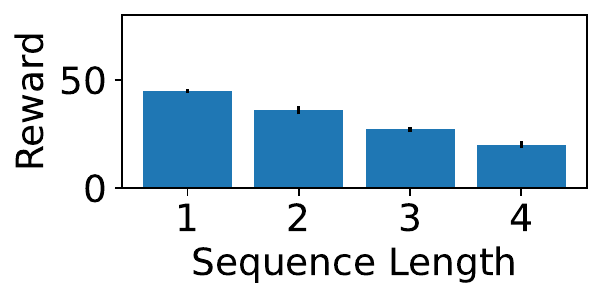}
            \caption[Q-Learning]%
            {{\small Q-Learning}}    
            \label{fig:q_learn_tabular_seq_len_train a}
        \end{subfigure}
        \begin{subfigure}[]{0.32\textwidth}   
            \centering 
            \includegraphics[width=\textwidth]{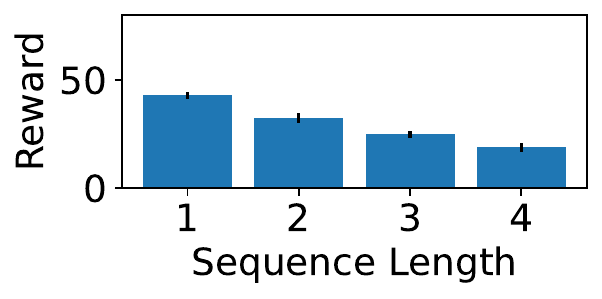}
            \caption[SARSA]%
            {{\small SARSA}}    
            \label{fig:sarsa_tabular_seq_len_train b}
        \end{subfigure}
        \begin{subfigure}[]{0.32\textwidth}
            \centering
            \includegraphics[width=\textwidth]{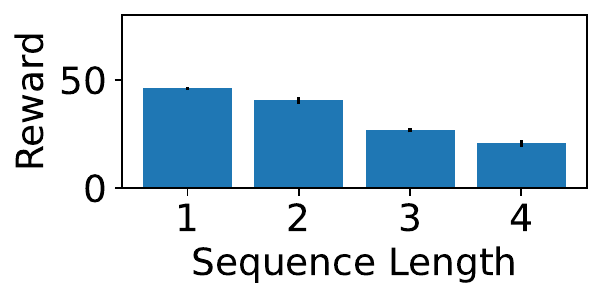}
            \caption[Double Q-Learning]%
            {{\small Double Q-Learning}}    
            \label{fig:double_q_learn_tabular_seq_len_train c}
        \end{subfigure}

        \caption[ AUC of episodic reward at the end of training ]
        {AUC of episodic reward at the end of training for three different tabular baseline algorithms when varying \textbf{sequence length}. Error bars represent bootstrapped confidence intervals with Bonferroni corrections for a significance level of 0.05.}
        \label{fig:tabular_seq_len}
\end{figure*}

\FloatBarrier
\section{Plots for Varying 2 Hardness Dimensions Together}\label{sec:append_2d_heatmaps}

We present here some more plots of experiments with 2 dimensions of hardness varied together. The experiments in this section are only for 10 seeds as opposed to 100 seeds for the main paper.

\begin{figure*}[ht!]
        \centering
        \begin{subfigure}[]{0.325\textwidth}
            \centering
            \includegraphics[width=\textwidth]{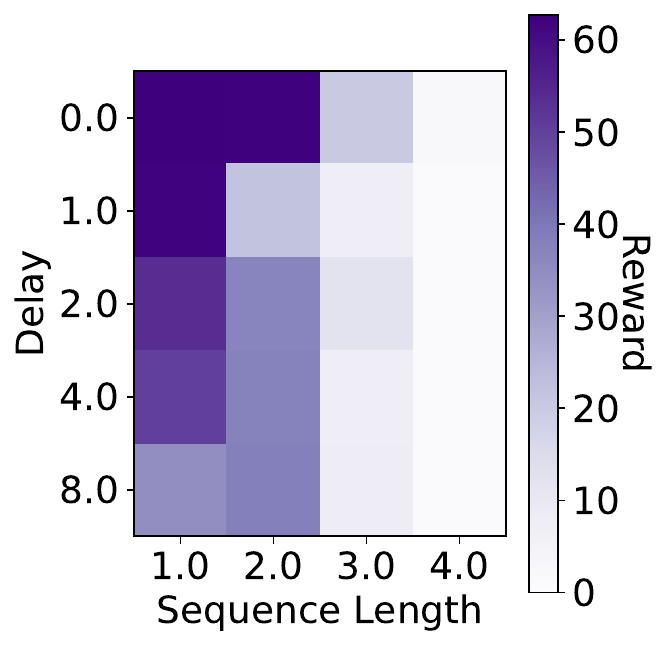}
            \caption[DQN]%
            {{\small DQN}}    
            \label{fig:DQN_seq_del a}
        \end{subfigure}
        \begin{subfigure}[]{0.325\textwidth}   
            \centering 
            \includegraphics[width=\textwidth]{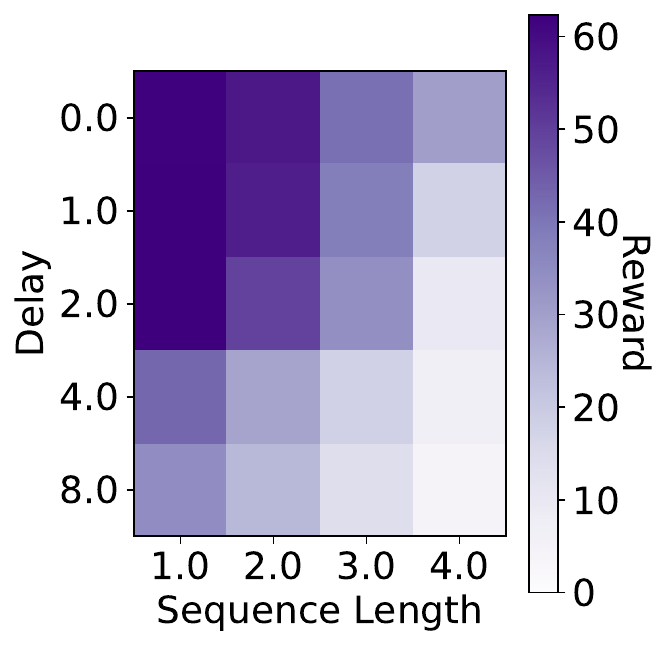}
            \caption[Rainbow]%
            {{\small Rainbow}}    
            \label{fig:rainbow del_seq c}
        \end{subfigure}
        \begin{subfigure}[]{0.325\textwidth}
            \centering
            \includegraphics[width=\textwidth]{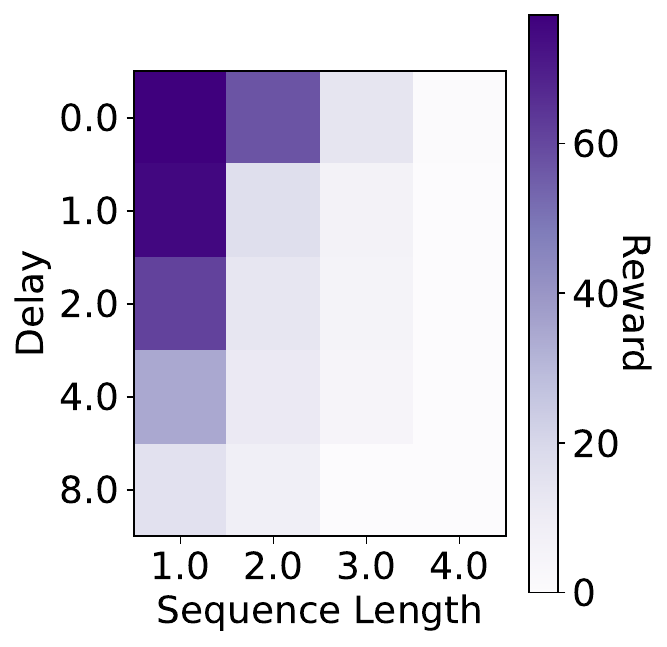}
            \caption[A3C]%
            {{\small A3C}}    
            \label{fig:a3c_del_seq a}
        \end{subfigure}

        \caption[ AUC of episodic reward at the end of training ]
        {AUC of episodic reward at the end of training for the different algorithms when varying \textbf{delay and sequence lengths}. Please note the different colour bar scales.} 
        \label{fig:algs_seq_del_train}
\end{figure*}

\begin{figure*}[ht]
        \centering
        \begin{subfigure}[]{0.325\textwidth}
            \centering
            \includegraphics[width=\textwidth]{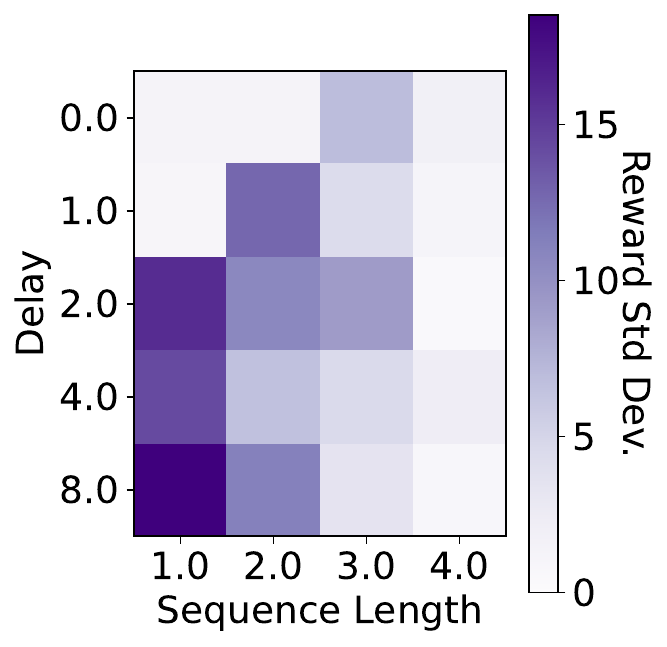}
            \caption[DQN]%
            {{\small DQN}}    
            \label{fig:DQN_seq_del_train_2d_std}
        \end{subfigure}
        \begin{subfigure}[]{0.325\textwidth}   
            \centering 
            \includegraphics[width=\textwidth]{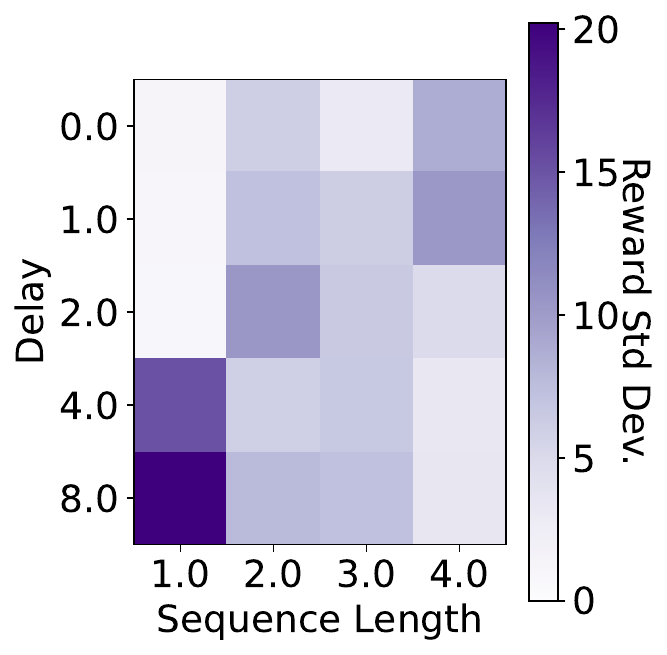}
            \caption[Rainbow]%
            {{\small Rainbow}}    
            \label{fig:rainbow_del_seq_train_2d_std}
        \end{subfigure}
        \begin{subfigure}[]{0.325\textwidth}
            \centering
            \includegraphics[width=\textwidth]{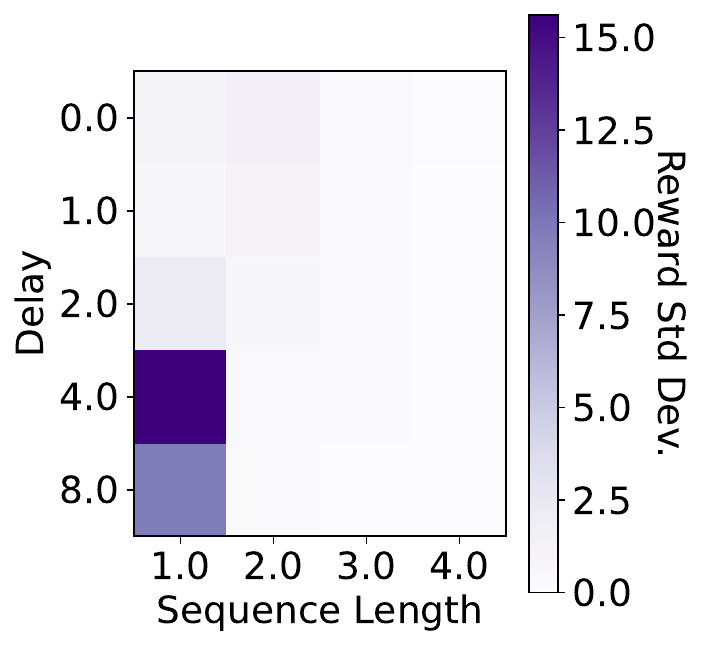}
            \caption[A3C]%
            {{\small A3C}}    
            \label{fig:a3c_del_seq_train_2d_std}
        \end{subfigure}

        \caption[ Standard deviation of mean episodic reward at the end of training ]
        {Standard deviation of AUC of mean episodic reward at the end of training for the different algorithms when varying \textbf{delay and sequence lengths}. Please note the different colour bar scales.}
        \label{fig:algs_seq_del_train_2d_std}
\end{figure*}

\begin{figure*}[ht]
        \centering
        \begin{subfigure}[]{0.325\textwidth}
            \centering
            \includegraphics[width=\textwidth]{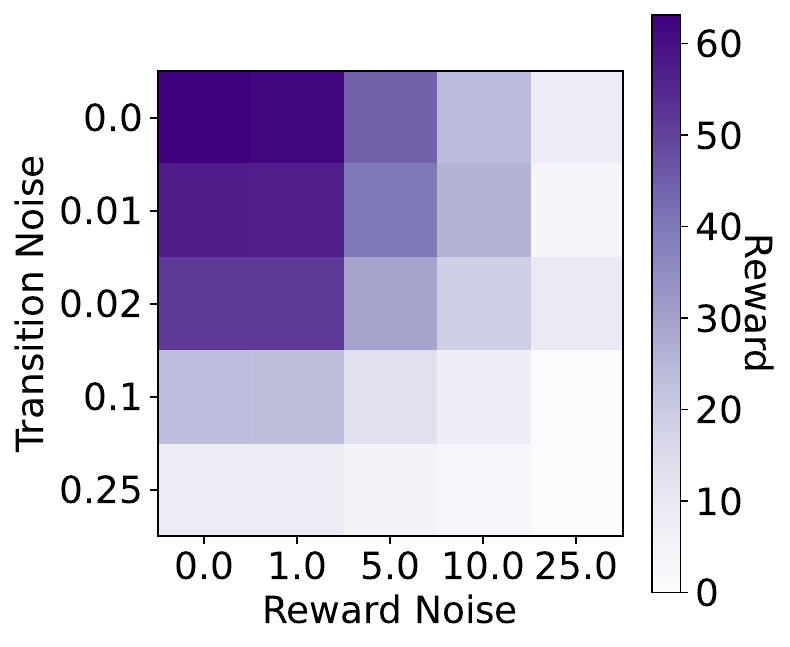}
            \caption[DQN]%
            {{\small DQN}}    
            \label{fig:dqn_noises}
        \end{subfigure}
        \begin{subfigure}[]{0.325\textwidth}   
            \centering 
            \includegraphics[width=\textwidth]{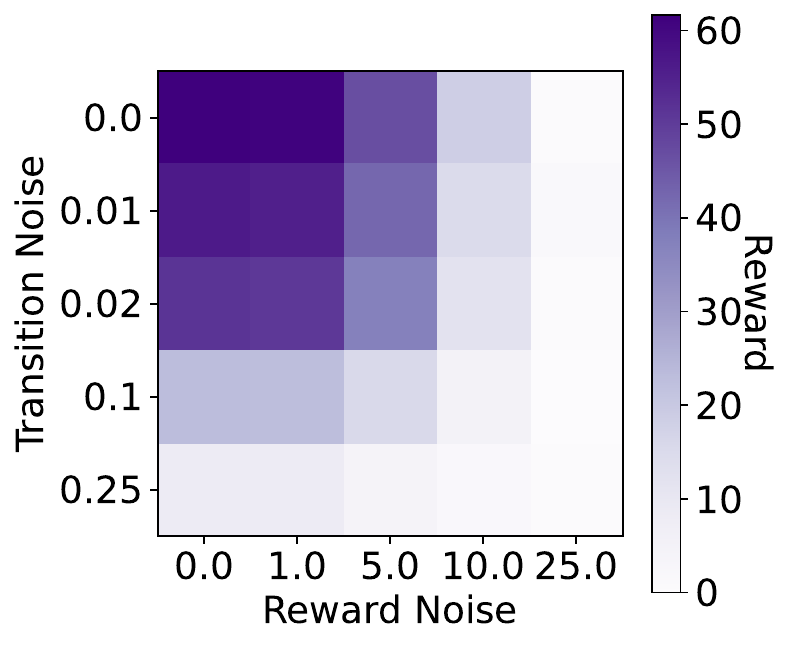}
            \caption[Rainbow]%
            {{\small Rainbow}}    
            \label{fig:rainbow noises}
        \end{subfigure}
        \begin{subfigure}[]{0.325\textwidth}
            \centering
            \includegraphics[width=\textwidth]{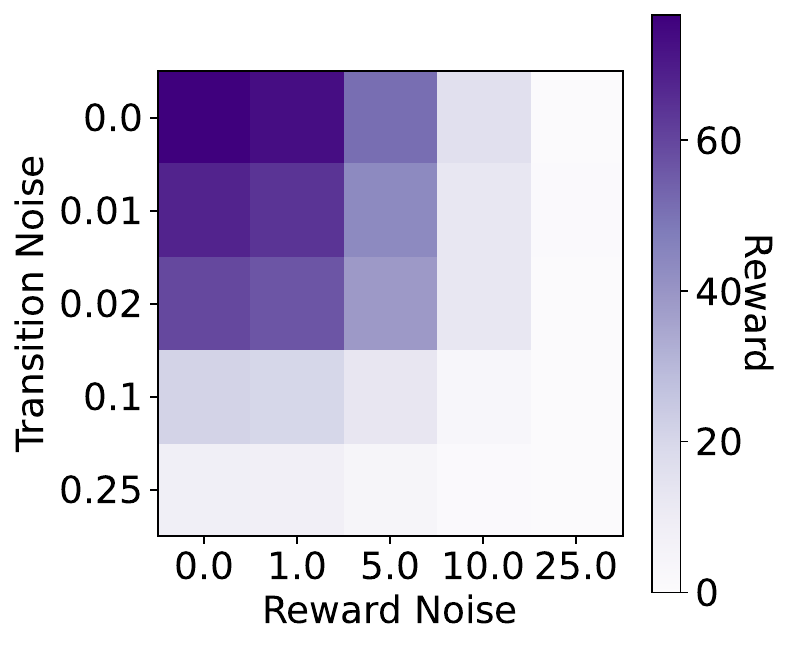}
            \caption[A3C]%
            {{\small A3C}}    
            \label{fig:a3c_noises}
        \end{subfigure}

        \caption[ AUC of episodic reward at the end of training ]
        {AUC of episodic reward at the end of training for the different algorithms when varying \textbf{transition noise and reward noise}. Please note the different colour bar scales.}
        \label{fig:algs_noises_train}
\end{figure*}

\begin{figure}[ht]
        \centering
        \begin{subfigure}[]{0.325\textwidth}
            \centering
            \includegraphics[width=\textwidth]{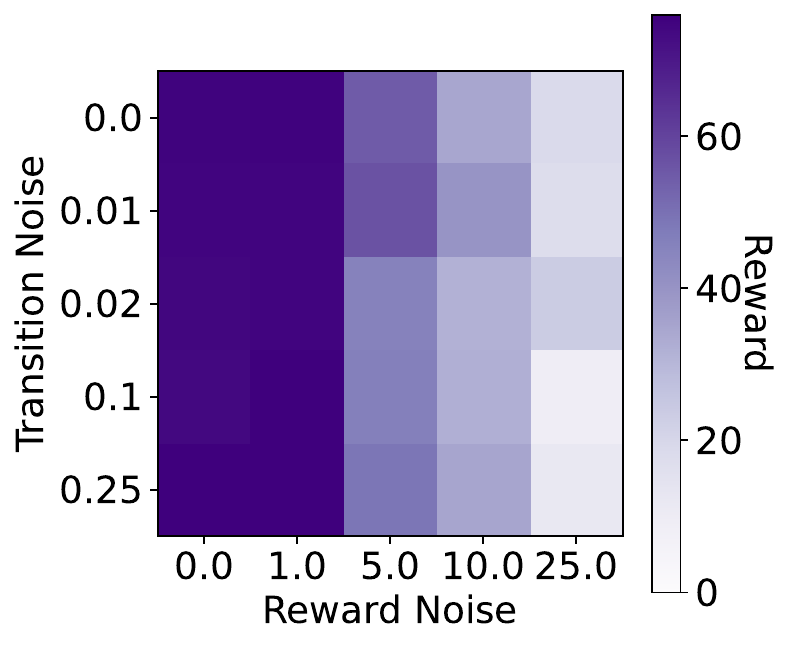}
            \caption[DQN]%
            {{\small DQN}}    
            \label{fig:dqn_noises eval}
        \end{subfigure}
        \begin{subfigure}[]{0.325\textwidth}   
            \centering 
            \includegraphics[width=\textwidth]{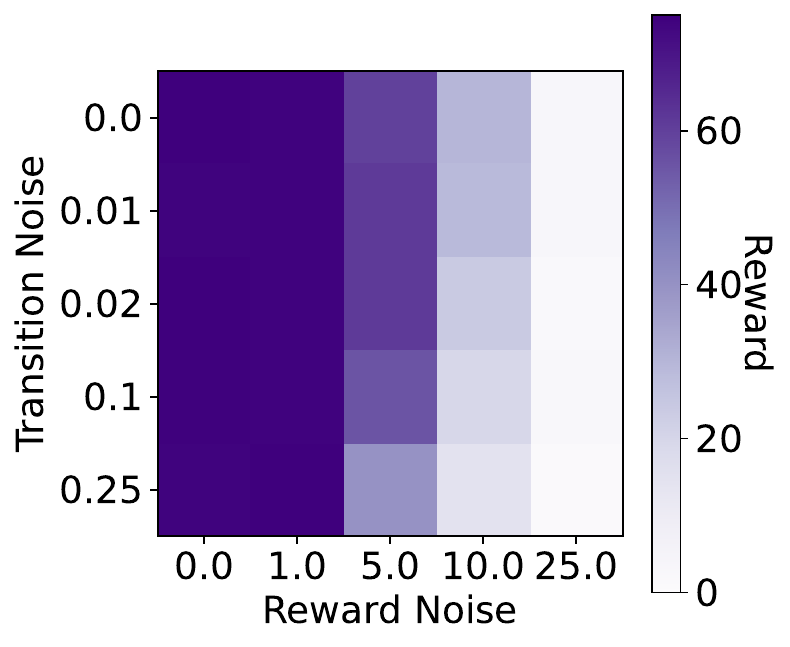}
            \caption[Rainbow]%
            {{\small Rainbow}}    
            \label{fig:rainbow noises eval}
        \end{subfigure}
        \begin{subfigure}[]{0.325\textwidth}
            \centering
            \includegraphics[width=\textwidth]{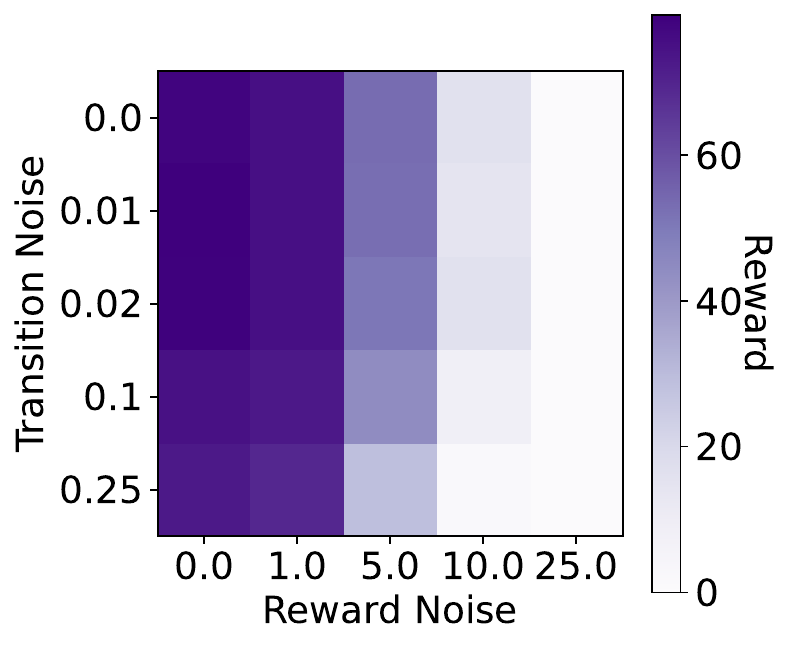}
            \caption[A3C]%
            {{\small A3C}}    
            \label{fig:a3c_noises eval}
        \end{subfigure}

        \caption[ Mean episodic evaluation rollout reward at the end of training ]
        {AUC of episodic reward when varying \textbf{transition and reward noise} when rolling out the policy that was learned on the noisy environment on a noise-free setting for the different algorithms. Please note the different colour bar scales.
        } 
        \label{fig:algs_noises_eval}
\end{figure}

\begin{figure*}[ht]
        \centering
        \begin{subfigure}[]{0.325\textwidth}
            \centering
            \includegraphics[width=\textwidth]{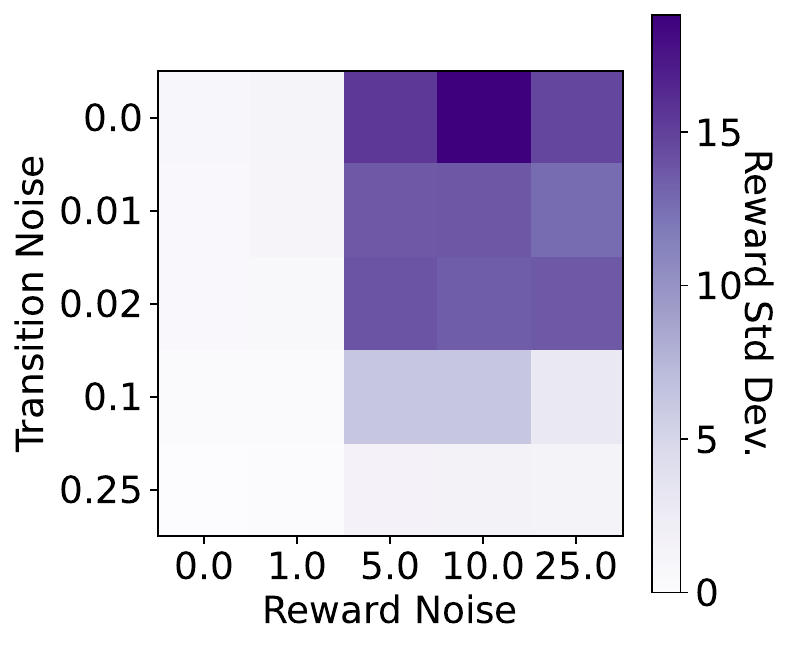}
            \caption[DQN]%
            {{\small DQN}}    
            \label{fig:dqn_noises_train_2d_std}
        \end{subfigure}
        \begin{subfigure}[]{0.325\textwidth}   
            \centering 
            \includegraphics[width=\textwidth]{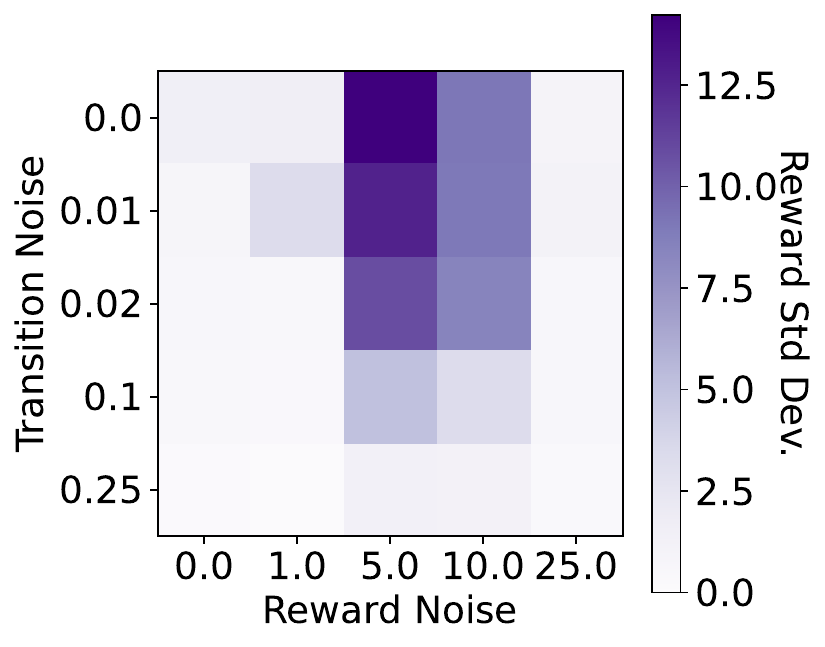}
            \caption[Rainbow]%
            {{\small Rainbow}}    
            \label{fig:rainbow_noises_train_2d_std}
        \end{subfigure}
        \begin{subfigure}[]{0.325\textwidth}
            \centering
            \includegraphics[width=\textwidth]{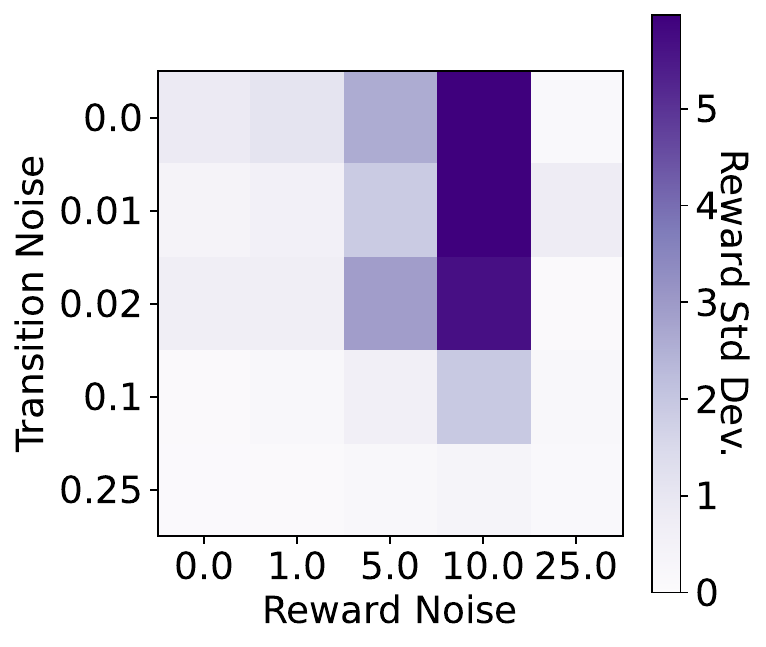}
            \caption[A3C]%
            {{\small A3C}}    
            \label{fig:a3c_noises_train_2d_std}
        \end{subfigure}

        \caption[ Standard deviation of mean episodic reward at the end of training ]
        {Standard deviation of AUC of mean episodic reward at the end of training for the different algorithms when varying \textbf{transition noise and reward noise}. Please note the different colour bar scales.}
        \label{fig:algs_noises_train_2d_std}
\end{figure*}

\begin{figure*}[ht]
        \centering
        \begin{subfigure}[]{0.325\textwidth}
            \centering
            \includegraphics[width=\textwidth]{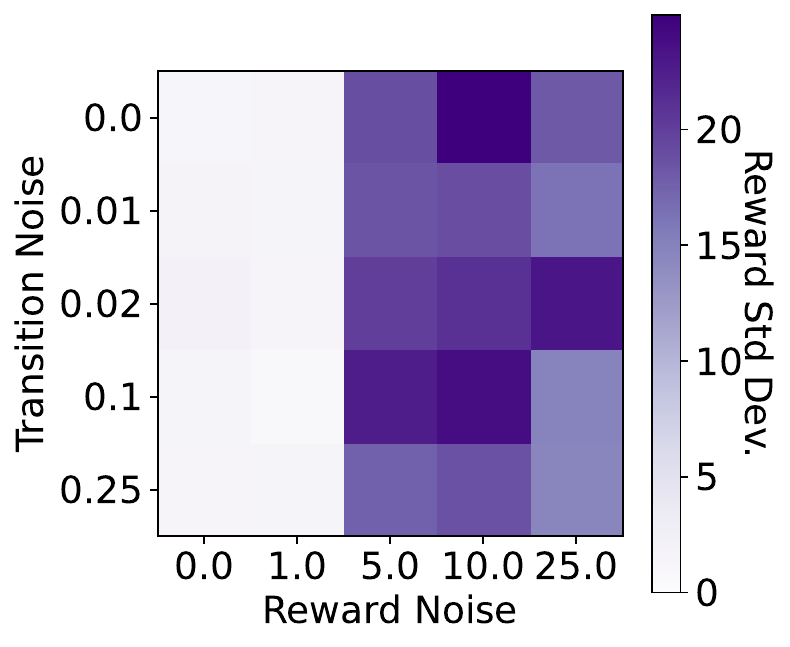}
            \caption[DQN]%
            {{\small DQN}}    
            \label{fig:dqn_noises_eval_2d_std}
        \end{subfigure}
        \begin{subfigure}[]{0.325\textwidth}   
            \centering 
            \includegraphics[width=\textwidth]{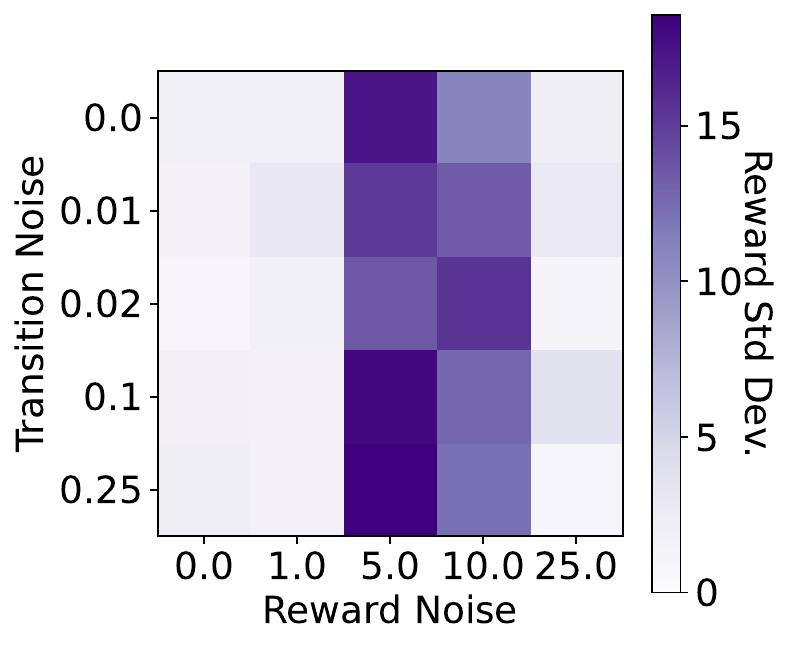}
            \caption[Rainbow]%
            {{\small Rainbow}}    
            \label{fig:rainbow_noises_eval_2d_std}
        \end{subfigure}
        \begin{subfigure}[]{0.325\textwidth}
            \centering
            \includegraphics[width=\textwidth]{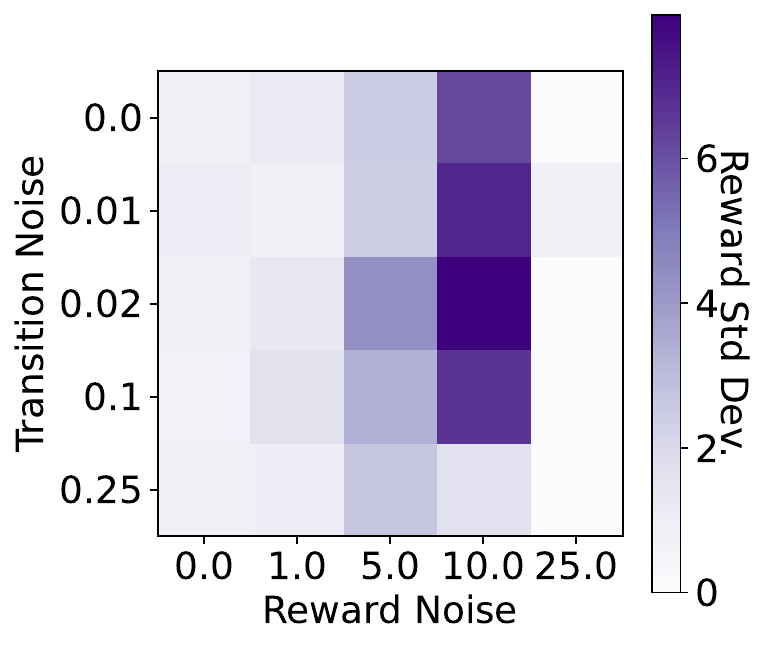}
            \caption[A3C]%
            {{\small A3C}}    
            \label{fig:a3c_noises_eval_2d_std}
        \end{subfigure}

        \caption[ Standard deviation of mean episodic reward at the end of training ]
        {Standard deviation of AUC of mean episodic reward when varying \textbf{transition and reward noise} when rolling out the policy that was learned on the noisy environment on a noise-free setting for the different algorithms. Please note the different colour bar scales.}
        \label{fig:algs_noises_eval_2d_std}
\end{figure*}

\begin{figure*}[ht!]
        \centering
        \begin{subfigure}[]{0.325\textwidth}
            \centering
            \includegraphics[width=\textwidth]{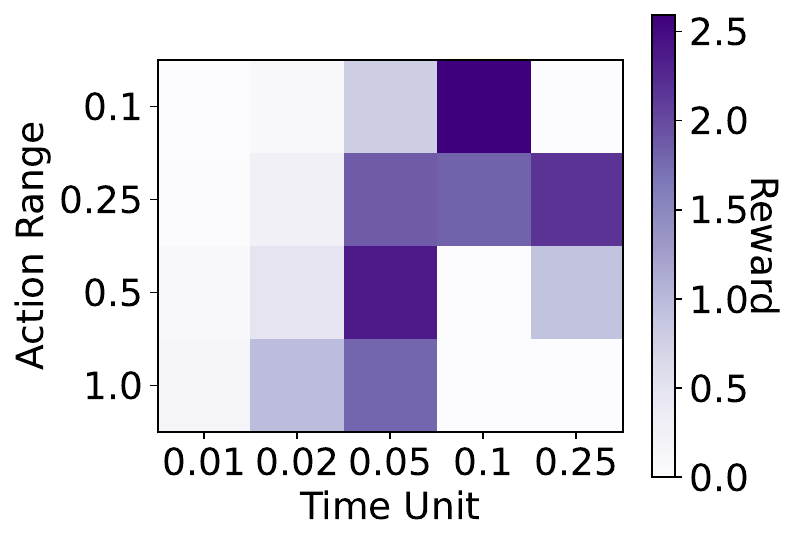}
            \caption[DDPG]%
            {{\small DDPG $P$ Order 2}}    
            \label{fig:ddpg_p_order_2_action_max_time_unit}
        \end{subfigure}
        \begin{subfigure}[]{0.325\textwidth}   
            \centering 
            \includegraphics[width=\textwidth]{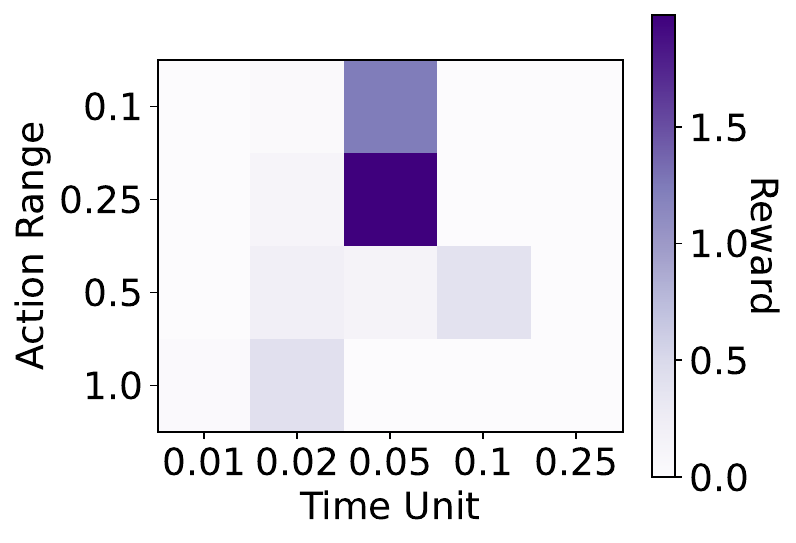}
            \caption[DDPG]%
            {{\small DDPG $P$ Order 3}}    
            \label{fig:ddpg_p_order_3_action_max_time_unit}
        \end{subfigure}
        \begin{subfigure}[]{0.325\textwidth}
            \centering
            \includegraphics[width=\textwidth]{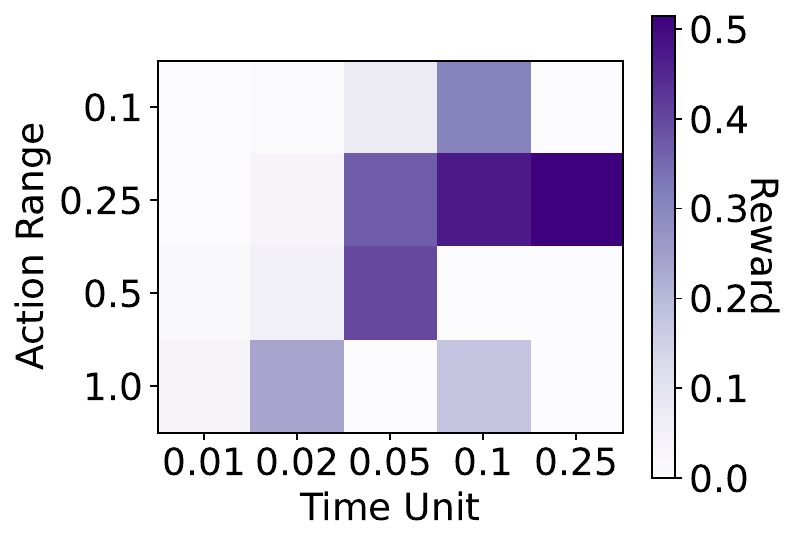}
            \caption[TD3]%
            {{\small TD3 $P$ Order 2}}    
            \label{fig:td3_p_order_2_action_max_time_unit}
        \end{subfigure}
        \begin{subfigure}[]{0.325\textwidth}   
            \centering 
            \includegraphics[width=\textwidth]{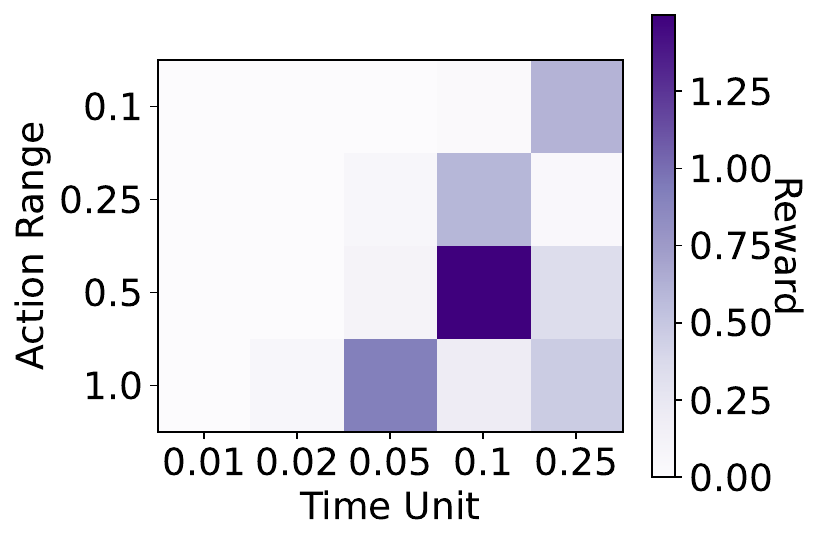}
            \caption[SAC]%
            {{\small SAC $P$ Order 2}}
            \label{fig:sac_p_order_2_action_max_time_unit}
        \end{subfigure}

        \caption[ AUC of episodic reward at the end of training ]
        {AUC of episodic reward at the end of training for the different algorithms when varying \textbf{action space max and time unit for a given $P$ order}. Please note the different colour bar scales.}
        \label{fig:cont_p_order_action_max_time_unit}
\end{figure*}

\begin{figure*}[ht!]
        \centering
        \begin{subfigure}[]{0.325\textwidth}
            \centering
            \includegraphics[width=\textwidth]{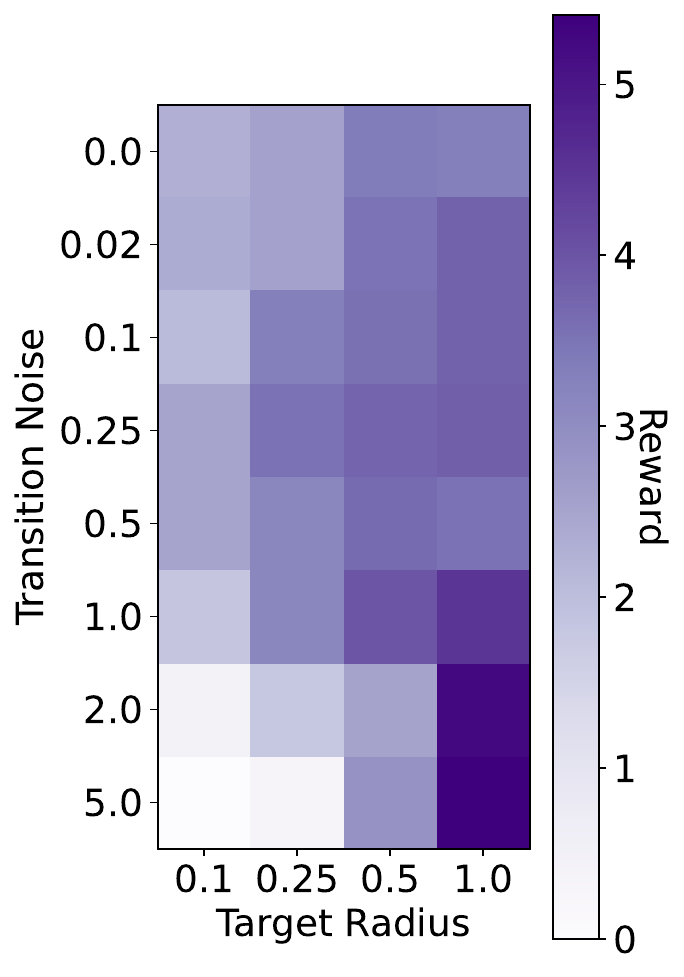}
            \caption[TD3]%
            {{\small TD3 episode rew.}}    
            \label{fig:td3_episode_rew_p_noise_target_radius}
        \end{subfigure}
        \begin{subfigure}[]{0.325\textwidth}   
            \centering 
            \includegraphics[width=\textwidth]{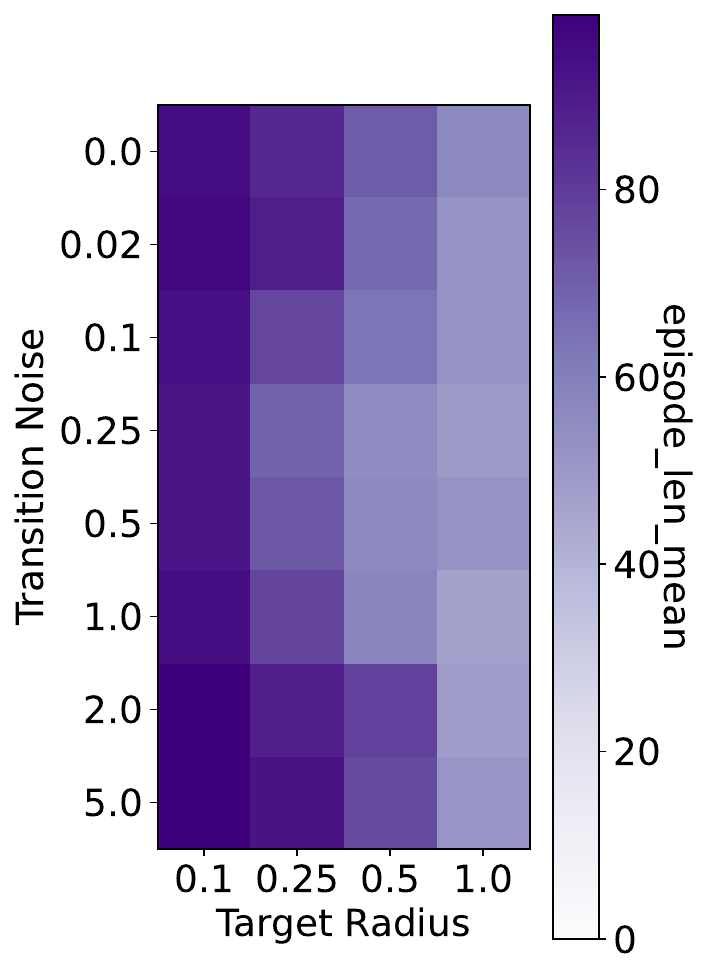}
            \caption[TD3]%
            {{\small TD3 episode len.}}    
            \label{fig:td3_episode_len_p_noise_target_radius}
        \end{subfigure}
        \begin{subfigure}[]{0.325\textwidth}
            \centering
            \includegraphics[width=\textwidth]{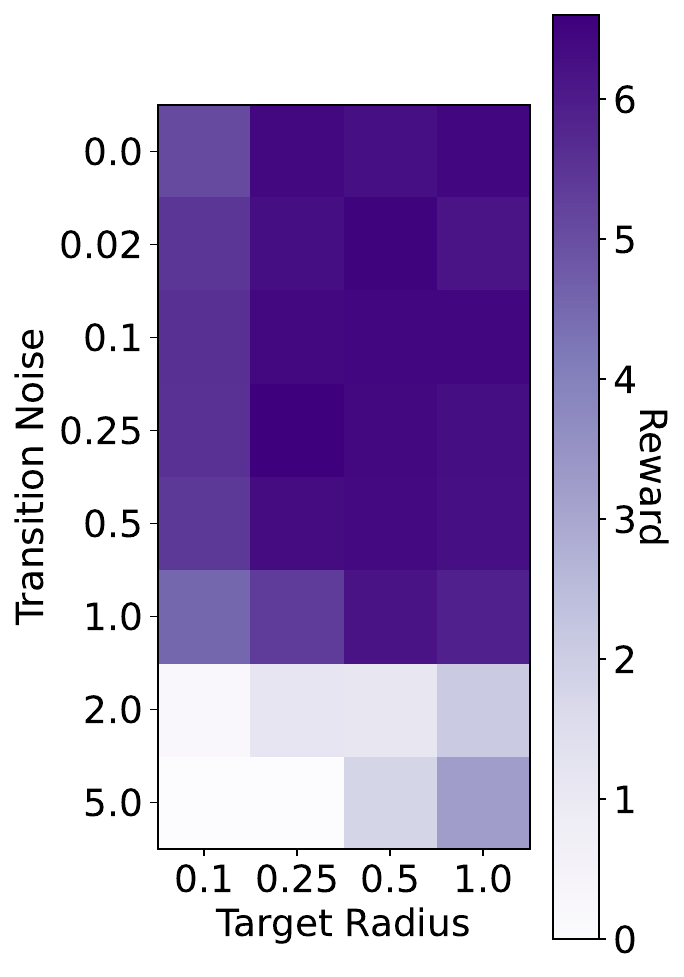}
            \caption[SAC]%
            {{\small SAC episode rew.}}    
            \label{fig:sac_episode_rew_p_noise_target_radius}
        \end{subfigure}
        \begin{subfigure}[]{0.325\textwidth}   
            \centering 
            \includegraphics[width=\textwidth]{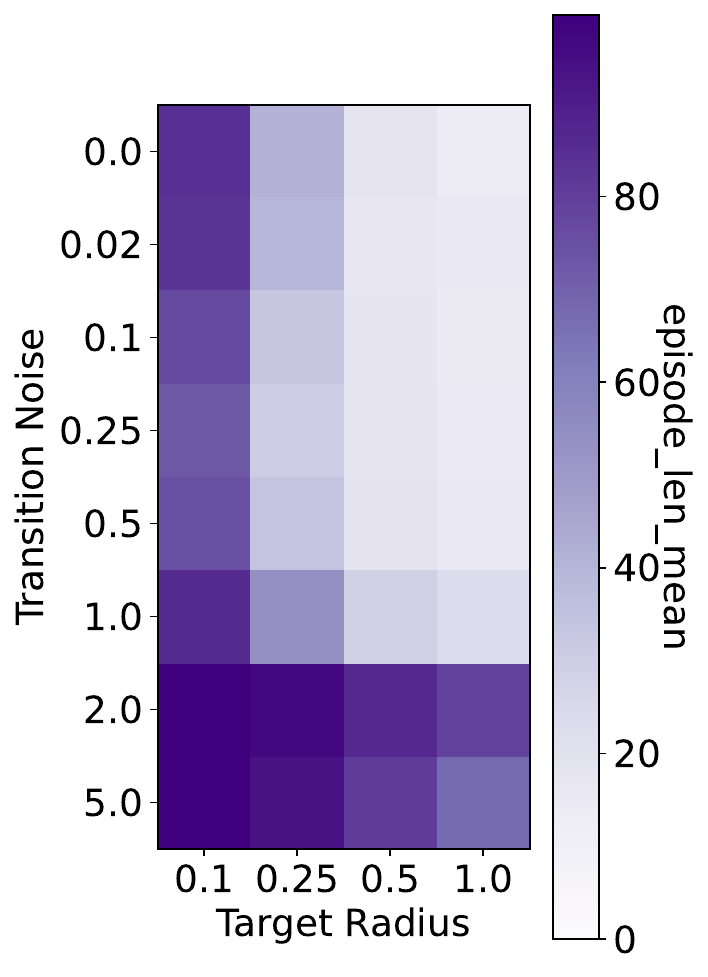}
            \caption[SAC]%
            {{\small SAC episode len.}}
            \label{fig:sac_episode_len_p_noise_target_radius}
        \end{subfigure}

        \caption[ AUC of episodic reward and lengths at the end of training ]
        {AUC of episodic reward and lengths at the end of training for the different algorithms when varying \textbf{$P$ noise and \textit{target radius}}. Please note the different colour bar scales.}
        \label{fig:cont_p_noise_target_radius}
\end{figure*}

\FloatBarrier
\section{Hyperparameter Tuning}\label{sec:hpo_append}
We gained some interesting insights into the significance of certain hyperparameters while tuning them for the different algorithms. Thus, our toy environments might in fact be good test beds for researching hyperparameters in RL, too. For instance, \textit{target network update frequency} turned out to be very significant for learning and sub-optimal values led to very noisy and unreliable training and unexpected results such as networks with greater capacity not performing well. Once we tuned it, however, training was much more reliable and, as expected, networks with greater capacity did well. 


Hyperparameters were tuned for the vanilla environment; we did so manually in order to obtain good intuition about them before applying automated tools. 
We tuned the hyperparameters in sets, loosely in order of their significance and did $3$ runs over each setting to get a more robust performance estimate. 
We describe an example of the tuning for DQN next. All hyperparameter settings for tuned agents can be found in Appendix \ref{sec:append_tuned_hyperparams}.

\begin{figure*}[!ht]
        \centering
        \begin{subfigure}[]{0.38\textwidth}
            \centering
            \includegraphics[width=\textwidth]{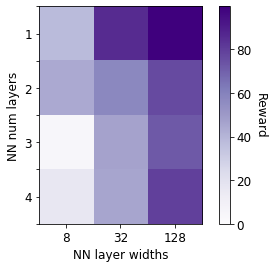}
            \caption[]%
            {{\small Reward}}    
            \label{fig:dqn_hyperparams nn size}
        \end{subfigure}
        \begin{subfigure}[]{0.38\textwidth}   
            \centering 
            \includegraphics[width=\textwidth]{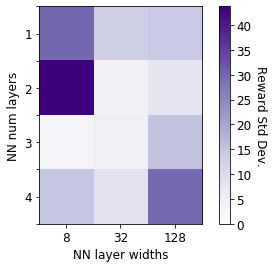}
            \caption[]%
            {{\small Std dev.}}
            \label{fig:dqn_hyperparams nn size std}
        \end{subfigure}
        \begin{subfigure}[]{0.38\textwidth}   
            \centering 
            \includegraphics[width=\textwidth]{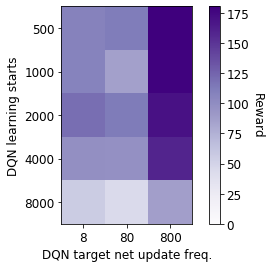}
            \caption[]%
            {{\small Reward}}    
            \label{fig:dqn params}
        \end{subfigure}
        \begin{subfigure}[]{0.38\textwidth}   
            \centering 
            \includegraphics[width=\textwidth]{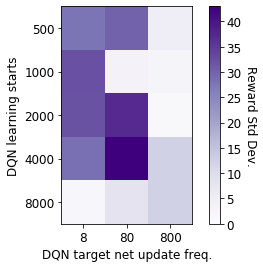}
            \caption[]%
            {{\small Std. Dev.}}    
            \label{fig:dqn params std}
        \end{subfigure}
        \begin{subfigure}[]{0.38\textwidth}
            \centering
            \includegraphics[width=\textwidth]{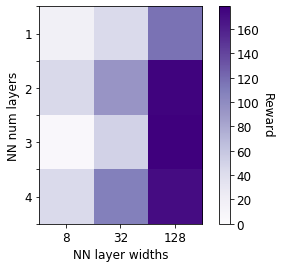}
            \caption[]%
            {{\small Reward}}    
            \label{fig:dqn_hyperparams nn size2}
        \end{subfigure}
        \begin{subfigure}[]{0.38\textwidth}   
            \centering 
            \includegraphics[width=\textwidth]{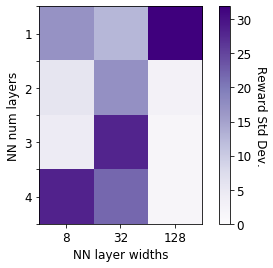}
            \caption[]%
            {{\small Std dev.}}
            \label{fig:dqn_hyperparams nn size2 std}
        \end{subfigure}
        \caption[ Results of tuning of hyperparams ]
        {AUC of episodic reward at the end of training for different hyperparameter sets for DQN. Please note the different colour bar scales.} 
        \label{fig:dqn hyperparam tuning}
\end{figure*}

We expected that quite small neural networks would already perform well for such toy environments and we initially grid searched over small network sizes (Figure \ref{fig:dqn_hyperparams nn size}). However, the variance in performance was quite high (Figure \ref{fig:dqn_hyperparams nn size std}). When we tried to tune DQN hyperparameters \textit{learning starts} and \textit{target network update frequency}, however, it became clear that the target network update frequency was very significant (Figure \ref{fig:dqn params} and \ref{fig:dqn params std}) and when we repeated the grid search over network sizes with a better value of $800$ for the target network update frequency (instead of the old $80$) this led to both better performance and lower variance (Figure \ref{fig:dqn_hyperparams nn size2} and \ref{fig:dqn_hyperparams nn size2 std}).

We then changed the network number of neurons grid to [$128$, $256$, $512$] and changed target network update frequency grid to [$80$, $800$, $8000$] and continued with further tuning using the grid values specified in Appendix \ref{sec:append_tuned_hyperparams}.

\section{Tuned Hyperparameters}
\label{sec:append_tuned_hyperparams}
The code for corresponding experiments for both discrete and continuous environments can be found in the accompanying code for the paper. The experiments with \textit{\_tune\_hps} in the names contain the grid of HPs that were tuned over. In some instances (where \textit{\_tune\_hps} experiments do not exist), in order to save costs, we used the default HPs in Ray. The README in the GitHub repo describes how to run the experiments using \textit{config} files and which \textit{config} files correspond to which experiments. Older experiments on the discrete toy environments were run with Ray 0.7.3, while for the newer continuous and complex environments, they were run with Ray 0.9.0. We had to use Ray 0.7.3 for the discrete toy environments and Ray 0.9.0 for the continuous toy ones because we had run the discrete cases for a previous version of the paper on 0.7.3. DDPG was not working and SAC was not implemented in Ray at that time. We tried to use Ray 0.9.0 also for the discrete version but found for the first few algorithms we tested that, for the same hyperparameters, the results did not transfer even across implementations of the same library. This further makes our point about using our platform to unit test algorithms. For the complex environments, since we had to tune the agents again anyway, we decided to use the newer Ray version. The names of the hyperparameters for the algorithms in the \textit{config} files will match those used in the respective Ray versions (i.e., 0.7.3 and 0.9.0).







\FloatBarrier
\section{CPU Specifications}\label{sec:cpu_info}
Cluster experiments were run on \textit{Intel(R) Xeon(R) CPU E5-2630 v4 @ 2.20GHz} cores for approximately 300000 CPU hours.

\subsection{CO2 Emission Related to Experiments}
Experiments were conducted using a private infrastructure, which has a carbon efficiency of 0.432 kgCO$_2$eq/kWh. A cumulative of 300000 hours of computation was performed on hardware of type Intel Xeon E5-2630v4 (TDP of 85W).

Total emissions are estimated to be 5140.8 kgCO$_2$eq of which 0 percents were directly offset.
    

Estimations were conducted using the \href{https://mlco2.github.io/impact#compute}{MachineLearning Impact calculator} presented in \shortciteA{co2_lacoste2019quantifying}.




    


The cluster CPU core specifications were:
\begin{lstlisting}[breaklines]
processor	: 0
vendor_id	: GenuineIntel
cpu family	: 6
model		: 79
model name	: Intel(R) Xeon(R) CPU E5-2630 v4 @ 2.20GHz
stepping	: 1
microcode	: 0xb000020
cpu MHz		: 1200.170
cache size	: 25600 KB
physical id	: 0
siblings	: 20
core id		: 0
cpu cores	: 10
apicid		: 0
initial apicid	: 0
fpu		: yes
fpu_exception	: yes
cpuid level	: 20
wp		: yes
flags		: fpu vme de pse tsc msr pae mce cx8 apic sep mtrr pge mca cmov pat pse36 clflush dts acpi mmx fxsr sse sse2 ss ht tm pbe syscall nx pdpe1gb rdtscp lm constant_tsc arch_perfmon pebs bts rep_good 
nopl xtopology nonstop_tsc aperfmperf eagerfpu pni pclmulqdq dtes64 monitor ds_cpl vmx smx est tm2 ssse3 sdbg fma cx16 xtpr pdcm pcid dca sse4_1 sse4_2 x2apic movbe popcnt tsc_deadline_timer aes xsave avx f16c r
drand lahf_lm abm 3dnowprefetch epb cat_l3 cdp_l3 invpcid_single intel_ppin intel_pt tpr_shadow vnmi flexpriority ept vpid fsgsbase tsc_adjust bmi1 hle avx2 smep bmi2 erms invpcid rtm cqm rdt_a rdseed adx smap x
saveopt cqm_llc cqm_occup_llc cqm_mbm_total cqm_mbm_local dtherm ida arat pln pts
bogomips	: 4389.48
clflush size	: 64
cache_alignment	: 64
address sizes	: 46 bits physical, 48 bits virtual
power management:
 
\end{lstlisting}

The laptop CPU core specifications were:
\begin{lstlisting}[breaklines]
processor	: 0
vendor_id	: GenuineIntel
cpu family	: 6
model		: 158
model name	: Intel(R) Core(TM) i7-8850H CPU @ 2.60GHz
stepping	: 10
microcode	: 0xb4
cpu MHz		: 900.055
cache size	: 9216 KB
physical id	: 0
siblings	: 12
core id		: 0
cpu cores	: 6
apicid		: 0
initial apicid	: 0
fpu		: yes
fpu_exception	: yes
cpuid level	: 22
wp		: yes
flags		: fpu vme de pse tsc msr pae mce cx8 apic sep mtrr pge mca cmov pat pse36 clflush dts acpi mmx fxsr sse sse2 ss ht tm pbe syscall nx pdpe1gb rdtscp lm constant_tsc art arch_perfmon pebs bts rep_good nopl xtopology nonstop_tsc cpuid aperfmperf tsc_known_freq pni pclmulqdq dtes64 monitor ds_cpl vmx smx est tm2 ssse3 sdbg fma cx16 xtpr pdcm pcid sse4_1 sse4_2 x2apic movbe popcnt tsc_deadline_timer aes xsave avx f16c rdrand lahf_lm abm 3dnowprefetch cpuid_fault epb invpcid_single pti ssbd ibrs ibpb stibp tpr_shadow vnmi flexpriority ept vpid ept_ad fsgsbase tsc_adjust bmi1 hle avx2 smep bmi2 erms invpcid rtm mpx rdseed adx smap clflushopt intel_pt xsaveopt xsavec xgetbv1 xsaves dtherm ida arat pln pts hwp hwp_notify hwp_act_window hwp_epp md_clear flush_l1d
bugs		: cpu_meltdown spectre_v1 spectre_v2 spec_store_bypass l1tf mds swapgs
bogomips	: 5184.00
clflush size	: 64
cache_alignment	: 64
address sizes	: 39 bits physical, 48 bits virtual
power management:
\end{lstlisting}

\section{Additional Learning Curves for Toy Environments}\label{sec:append_learn_curves}

Please find the learning curves for this section in the arxiv version of our paper: \url{https://arxiv.org/abs/1909.07750v3}. We present there plots of the learning curves of the experiments which provide more details into how learning progressed during training. The plots show how each seed performed. As a result, we plot them only for 10 seeds as opposed to 100 seeds for the main paper to maintain visual clarity.

\section{Additional Learning Curves for Complex environments}\label{sec:append_complex_envs_learn_curves}

Please find the learning curves for this section in the arxiv version of our paper: \url{https://arxiv.org/abs/1909.07750v3}. The curves contain a lot of data points and hence have large file sizes which leads to exceeding the file size limit for this submission.

\bibliography{biblio}
\bibliographystyle{theapa}


\end{document}